\documentclass[10pt,journal,tevc]{IEEEtran}
\usepackage{algorithm,algorithmic}
\usepackage{stackengine,multirow}
\usepackage{graphicx}
\usepackage[table]{xcolor}
\usepackage{epstopdf,rotating,array}
\usepackage{amsfonts,longtable,amssymb}
\usepackage{url}
\usepackage{cite}
\usepackage{color, balance,amsmath,subfigure,mathtools}
\usepackage{eucal}
\newcommand\MYhyperrefoptions{bookmarks=true,bookmarksnumbered=true,
pdfpagemode={UseOutlines},plainpages=false,pdfpagelabels=true,
colorlinks=true,linkcolor={black},citecolor={black},urlcolor={black},
pdftitle={An Efficient Approach for Solving Expensive Constrained Multiobjective Optimization Problems},
pdfsubject={Typesetting},
pdfauthor={},
pdfkeywords={Surrogate~Assisted~Optimization, Multi-objective algorithm, Infill strategy, Steady-state algorithm, constrained optimization}}
\usepackage[\MYhyperrefoptions,pdftex]{hyperref}
\newcolumntype{C}[1]{>{\centering\let\newline\\\arraybackslash\hspace{0pt}}m{#1}}

\begin{document}

\title{An Efficient Approach for Solving Expensive Constrained Multiobjective Optimization Problems}

\author{Kamrul Hasan Rahi,~\IEEEmembership{Graduate Student Member,~IEEE}
\IEEEcompsocitemizethanks{\IEEEcompsocthanksitem The author is with The University of New South Wales, Australia. Emails:
k.rahi@unsw.edu.au}}

\IEEEtitleabstractindextext{
\begin{abstract}  

To solve real-world \emph{expensive} constrained multi-objective optimization problems~(ECMOPs), surrogate/approximation models are commonly incorporated in evolutionary algorithms to pre-select promising candidate solutions for evaluation. However, the performance of existing approaches are highly dependent on the relative position of unconstrained and constrained Pareto fronts~(UPF and CPF, respectively). In addition, the uncertainty information of surrogate models is often ignored, which can misguide the search. To mitigate these key issues~(among others), an efficient probabilistic selection based constrained multi-objective EA is proposed, referred to as PSCMOEA. It comprises novel elements such as (a) an adaptive search bound identification scheme based on the feasibility and convergence status of evaluated solutions (b) a probabilistic selection method backed by theoretical formulations of model mean and uncertainties to conduct search in the predicted space to identify promising solutions (c) an efficient single infill sampling approach to balance feasibility, convergence and diversity across different stages of the search and (d) an adaptive switch to unconstrained search based on certain search conditions. Numerical experiments are conducted on an extensive range of challenging constrained problems using low evaluation budgets to simulate ECMOPs. The performance of PSCMOEA is benchmarked against five competitive state-of-the-art algorithms, to demonstrate its competitive and consistent performance. 
             
\end{abstract}

}

\maketitle
\IEEEdisplaynontitleabstractindextext
\IEEEpeerreviewmaketitle

\section{Introduction}
\label{sec:intro}
\IEEEPARstart{D}{ecision} making in real-world optimization problems often requires trade-off considerations among conflicting objectives subject to satisfying various practical constraint(s). Formally this constitutes a constrained multi-objective optimization problem~(CMOP)~\cite{liang2022survey}, represented as shown in Eq.~\ref{eqn:prob}.

\begin{equation}\footnotesize
\begin{array}{lr}
\underset{{\mathbf{x}}}{\text{Minimize:}}\hspace{2mm}\mathbf{F(x)} =  \{f_1(\mathbf{x}),f_2(\mathbf{x}), \ldots f_M(\mathbf{x})\}\\
\text{subject to} 
\begin{cases}
\mathbf{x} \in \Omega \\
g_j(\mathbf{x}) \leq 0, & j = 1,\ldots p \hspace{9mm}\\
\end{cases}
\end{array}
\label{eqn:prob}
\end{equation}

Here, $\mathbf{x} = (x_1, x_2, \ldots, x_n)$ is an $n$-dimensional decision vector defining the decision/search space $\Omega = {\left[x^L_i,x^U_i\right]}^n \subseteq \mathbf{R}^n$~for real-valued continuous variables. The $M$ objective functions are $f_1(\mathbf{x}), \ldots, f_M(\mathbf{x})$, subject to $p$ inequality constraints $g_{i=1:p}(\mathbf{x}) \leq 0$. Equality constraints, if present, are generally converted to inequality constraints by applying relaxation techniques~\cite{liang2022survey}, hence they are omitted for brevity. The constraint violation~(CV) of a solution is computed as $CV(\mathbf{x}) = \sum_{j=1}^{p} \text{max}(0,g_j(\mathbf{x}))$, which is 0 for any feasible solution~(which satisfies all constraints), and a positive quantity for an infeasible solution~(which violates at least one constraint).

Between two feasible solutions $\mathbf{x}_1$ and $\mathbf{x}_2$, solution $\mathbf{x}_1$ dominates $\mathbf{x}_2$ if $f_i(\mathbf{x}_1) \leq f_i(\mathbf{x}_2)$ for each $i \in \{1, \ldots, M\}$ and $f_j(\mathbf{x}_1) < f_j(\mathbf{x}_2)$ for at least one $j \in \{1, \ldots, M\}$, which is denoted as $\mathbf{x}_1 \prec \mathbf{x}_2$. A feasible solution $\mathbf{x}^*$ is said to be a Pareto optimal if no other feasible solution $\mathbf{x}$ dominates $\mathbf{x}^*$. The set of all feasible Pareto optimal solutions are represented by a trade-off set known as constrained Pareto set~(CPS), while its image in the objective space is termed as constrained Pareto front~(CPF). Likewise, the Pareto set/front obtained by disregarding the constraints is referred to as unconstrained PS/PF~(UPS/UPF)\footnote{Note that feasibility is a necessary condition for optimality, hence generally CPS/CPF are simply referred to as PS/PF. The use of the terms CPS/CPF here is to explicitly distinguish them from UPS/UPF.}. Solving CMOPs is typically acknowledged to be more challenging than solving unconstrained MOPs~(UMOPs) due to the need to account for feasibility in addition to convergence and diversity of the solutions. This challenge is further aggravated for \emph{expensive} problems~(ECMOPs), wherein evaluation of each candidate design requires significant expense, for example, time-consuming numerical simulations or physical experiments~\cite{regis2021two}. Evidently, there are stringent limitations on the number of calls to the true~(expensive) evaluation that can be practically afforded when solving real-world problems of this nature. As an example, consider the Ford crash simulation discussed in \cite{wang2006review}, requiring 36-160 hours to assess a single design. Extrapolating from this, conducting 100 design evaluations~(sequentially) would require approx 5 months to 1.8 years. 

Multi-objective Evolutionary Algorithms~(MOEAs) are a common choice to solve MOPs. This is primarily attributed to their ability to deal with highly non-linear/black-box functions and to evolve a set of solutions~(`population') that naturally yields itself to a discrete approximation of the PF. Even so, they require evaluation of a large number of candidate solutions prior to convergence, and hence not suitable for solving EMOPs in their native form. To circumvent this challenge, surrogate-assisted MOEAs~(SA-MOEAs) are often used, wherein approximations of the expensive functions are built to guide the search for the most part, with expensive evaluation called only sparingly for promising candidates. While there exist many different types of surrogate models, such as polynomial response surface modeling~(RSM), radial basis function~(RBF), support vector regression~(SVR), one particular type of model, i.e, Kriging~(Gaussian Process), has gained significant attention~\cite{martinez2016kriging}. This is mainly attributed to the fact that unlike other models, which typically only provide a mean estimates of the response function, Kriging can also provide uncertainty bounds, which can be used for a more informed selection of infill solutions~\cite{rahi2022steady,martinez2016kriging,mazumdar2022probabilistic}. While SA-MOEAs have demonstrated significant progress in addressing EMOPs, majority of the studies focus on unconstrained EMOPs, as evident from~(and highlighted in) a recent survey~\cite{chugh2019survey}. Only a limited number of investigations have delved into ECMOPs, although this important problem of practical significance is gaining traction in the research community recently~\cite{deb2018taxonomy,hussein2016generative,zhang2023multigranularity,song2023balancing,yang2023surrogate,yin2022fast}.
 
Constraint-handling is a key element of solution approaches for solving CMOPs. It refers to strategies incorporated in one or more evolutionary operations, such as ranking, recombination or environmental selection, in order to drive the population towards feasibility and eventually cover the PF. The constraints may introduce several topological features in the search landscape that make it difficult for the search methods to navigate. These include, but are not limited to, presence of very constricted feasible regions, multiple disconnected regions and feasible regions with significant bias~(e.g., some parts of a CPF are achieved well before others). These different constraint scenarios affect the relative positions of UPF and CPF and/or induce particular difficulties in achieving certain parts of the CPF, which in turn impacts the performance of the solution methods. A number of constraint handling techniques~(CHTs) have been proposed for solving general CMOPs, some of which are also adapted to work for ECMOPs. However, limited evaluation budget constraint induces significant challenge to the algorithms for solving ECMOPs~\cite{wu2024surrogate}. While prior approaches have demonstrated potential in addressing ECMOPs~\cite{liang2022survey,he2023review,mazumdar2022probabilistic}, several unresolved challenges and promising prospects require further attention which motivate this research. Firstly, the algorithm is required to be efficient in navigating potential feasible region(s) followed by identifying well converged and distributed solutions on the CPF within the allocated evaluation budget by ensuring proper balance among feasibility, convergence and diversity. Secondly, the algorithm should be capable of utilizing the uncertainty information of surrogate models to ensure reliable performance comparison of solutions in the predicted landscape in order to avoid any misguided search. Lastly, limited attention has been paid to the development of \emph{steady-state} methods, where only a single candidate solution can be evaluated at a time~\cite{rahi2022steady}. There exist scarce approaches that cater to such scenarios, and are generally extensions of efficient global optimization~(EGO) techniques, e.g. MultiObjectiveEGO~\cite{hussein2016generative}. This is likely because formulating a sampling criterion that can balance feasibility, convergence and diversity concurrently is challenging in its own right. Most existing approaches therefore operate on a generational model~(multiple expensive evaluations done in each generation). However, recent studies have shown that steady-state approaches may be more advantageous for expensive UMOPs under limited evaluation budgets~\cite{rahi2022steady}, especially where the evaluations are non-parallelizable, and it is worth investigating if similar benefits can be realized for ECMOPs.

Towards addressing the above research gaps, the following contributions are made in this study:

\begin{enumerate}
    \item First, illustrative examples are provided to highlight the influence of the relative positions of UPF and CPF on some recent representative search methods. This helps explain why some existing approaches might struggle, and provides a motivation to address these challenges in the algorithm design. 
    \item A new CMOEA based on \emph{probabilistic selection} is proposed, referred to as PSCMOEA, to deal with ECMOPs. PSCMOEA is a steady-state algorithm, unlike most of its peers. Its mainstay in efficient search is the careful consideration of model uncertainties in its operation along with the status of the solutions in the archive in terms of feasibility, convergence and diversity. These considerations feed into various components, including (a) ranking/environmental selection using a theoretically derived probabilistic constrained dominance ($PCD$); (b) infill criteria that adapts itself based on the status of the evaluated solutions in the archive, catering for scenarios where all solutions are feasible, all are infeasible, or a mix of both. 
    \item PSCMOEA also contains other additional improvisations, such as normalization of the solutions in the presence of a mix of feasible/infeasible solutions, and adaptive switch to unconstrained search to improve convergence rate, triggered based on correlation between search directions of UPF and CPF. 
    \item Numerical experiments are presented on an extensive range of problems with challenging features to demonstrate the competence of PSCMOEA relative to five other state-of-the-art methods. 
\end{enumerate}

Following this introduction, Section~\ref{sec:lit} provides an overview of related work, highlighting the limitations of existing methods on specific problems to justify the motivation for this study. The proposed approach is detailed in Section~\ref{sec:algorithm}, followed by the numerical experiments in Section~\ref{sec:exp}. Concluding remarks are presented in Section~\ref{sec:conclusion}.

\section{Related work and motivation}
\label{sec:lit}

\subsection{Conventional constraint handling techniques for CMOPs}
\label{subsec:cmoeas}

CMOEAs have received significant attention in the recent years~\cite{liang2022survey}. Existing CMOEAs can be roughly classified into four main categories based on their specific constraint handling techniques~(CHTs). The first and perhaps the most used category separates the solutions into feasible and infeasible blocks based on the CV information, followed by preferring feasible solutions over infeasible ones~\cite{zhou2022domination,yuan2021indicator}. This is referred to as parameterless technique, constraint domination principle or feasibility-first principle. While it is a simple and an intuitive way of handling constraints, it may drive the population towards an easily identifiable portion of the CPF and miss other parts~\cite{ray2009infeasibility}. The second category involves converting a CMOP into an UMOP by adapting constraints as additional objectives or penalty functions~\cite{takahama2010constrained,liu2022multiobjective,ray2009infeasibility}. However, ensuring the UMOP shares the same optimum as the CMOP requires extensive parameter tuning. Besides, addition of new objective(s) may induce further complexities, such as those encountered in \emph{many}-objective optimization. The third and fourth categories can be considered as the representative of the most recent techniques. The third category utilizes multiple populations by setting different goals to optimize objectives and constraints collaboratively~\cite{qiao2023self,liu2023constrained,zou2023multi}. This is achieved by evolving dedicated populations for each goal. Typically the primary goal is set to identify feasible region(s) by considering constraints while the secondary goal attempts to expedite convergence without considering constraints~(i.e., search for UPF). Subsequently, these populations adopt a co-evolutionary approach by interacting with each other to enhance the effectiveness of evolutionary search and selection. However, the performance primarily depends on the relationship between these two goals, and disparities in optimal solution sets~(for UPF vs CPF) may misguide the search. The fourth category divides the evolutionary process into stages favoring specific optimization tasks~(e.g., with or without considering constraints) at different stages~\cite{ming2021novel,zuo2023process,sun2022multi}. However, appropriate transition of stages and optimal distribution of computational resources are critical considerations that affect the performance of the approach.

\subsection{Constraint handling techniques for ECMOPs}
\label{subsec:ecmoeas}

Although some of the past approaches incorporated simple mechanisms to handle constraints for ECMOPs~\cite{shankar2016multi,habib2019multiple,hussein2016generative}, dedicated works to solve more challenging ECMOPs have only recently started emerging. The design of SA-CMOEAs typically involves three key elements: model management, optimization framework, and infill identification approach. Various recent SA-MOEAs~(all proposed within last 3 years), such as SAMO-COBRA~\cite{de2022constrained}, IC-SA-NSGA-II~\cite{blank2021constrained}, KTS~\cite{song2023balancing}, ASA-MOEA/D~\cite{yang2023surrogate}, and KMGSAEA~\cite{zhang2023multigranularity}, employ different surrogate models and optimization strategies for evolving candidate solutions. It is also seen that recent SA-CMOEAs adopt strategies inspired by parallel developments in the field of generic CMOEAs. Notably, KTS uses two inter-switchable surrogate-assisted EAs~(with and without considering constraints) based on the correlation between the convergence direction and CV minimization direction, motivated by the concept of co-evolutionary approaches as discussed above. ASA-MOEA/D adaptively applies three types of local search~(e.g., convergence driven, diversity driven and feasibility driven) to optimize the subpopulations along different reference vectors~(RVs) within a decomposition based framework, demonstrating effectiveness of approximating CPFs of different shapes. KMGSAEA introduces a multigranularity Kriging-assisted framework to mitigate the adverse effects of model errors. Furthermore, another category of SA-CMOEAs is evident in literature which involves constructing and optimizing effective acquisition functions based on the surrogate prediction of response functions and different performance indicators such as expected Hypervolume (HV)~\cite{shang2020survey} improvement and expected improvement (EI)~\cite{jones1998efficient}. To address constraints in these approaches, the probability of feasibility (PoF)~\cite{forrester2008engineering} of solutions is typically multiplied with these acquisition functions to balance objective optimization with constraint satisfaction. In \cite{singh2017constrained}, a constraint-oriented criterion is formulated by multiplying PoF with the probability of HV improvement followed by optimizing the criterion to identify the optimal solution in antenna design optimization, while in \cite{han2020Kriging}, an R2 indicator-based EI is multiplied with PoF. In~\cite{zhan2017expected}, three separate infill criteria is designed by multiplying PoF with computationally efficient EIM~(expected improvement matrix), wherein non-dominated objective values are replaced with corresponding EI values. Besides, decomposition-based SA-CMOEAs~\cite{han2019efficient,tran2022srmo} enable parallel selection of multiple infill points by decomposing ECMOPs into constrained single-objective sub-problems where PoF is multiplied with unary indicators, contributing to improved diversity. 

\subsection{Motivations of this study}
\label{subsec:motive}

\begin{figure}[!ht]
\centering    
\subfigure[]{\label{fig:PF1}\includegraphics[width=0.14\textwidth]{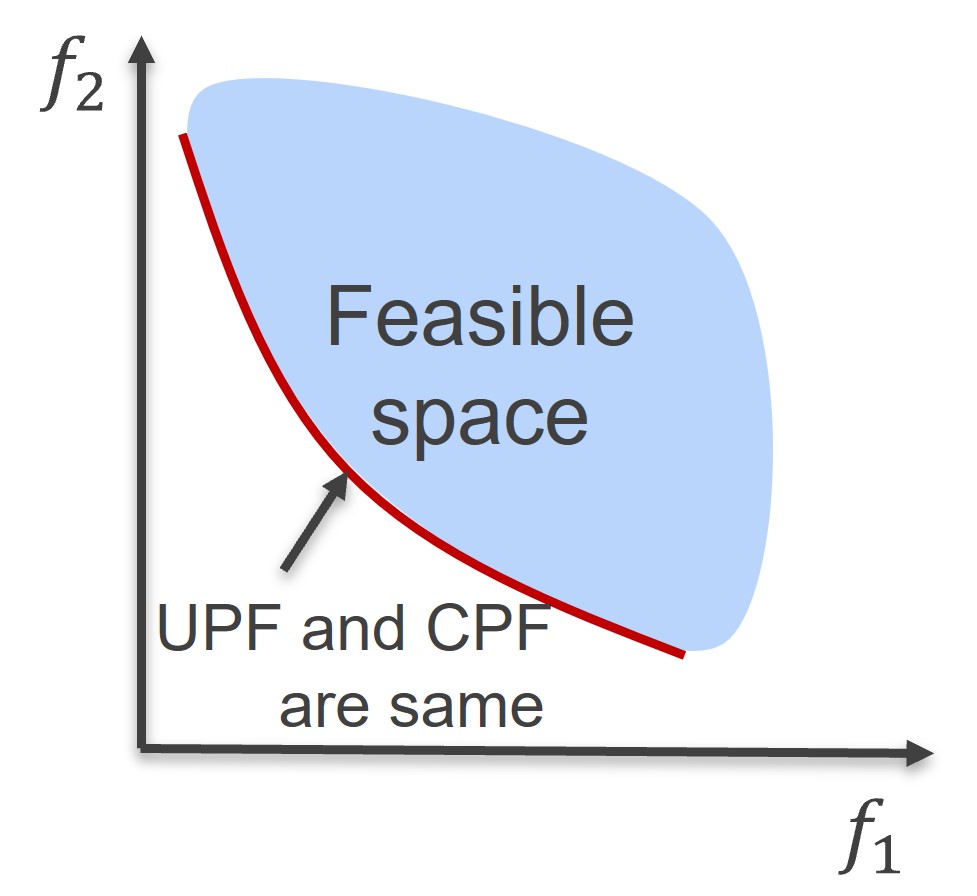}} \quad
\subfigure[]{\label{fig:PF2}\includegraphics[width=0.14\textwidth]{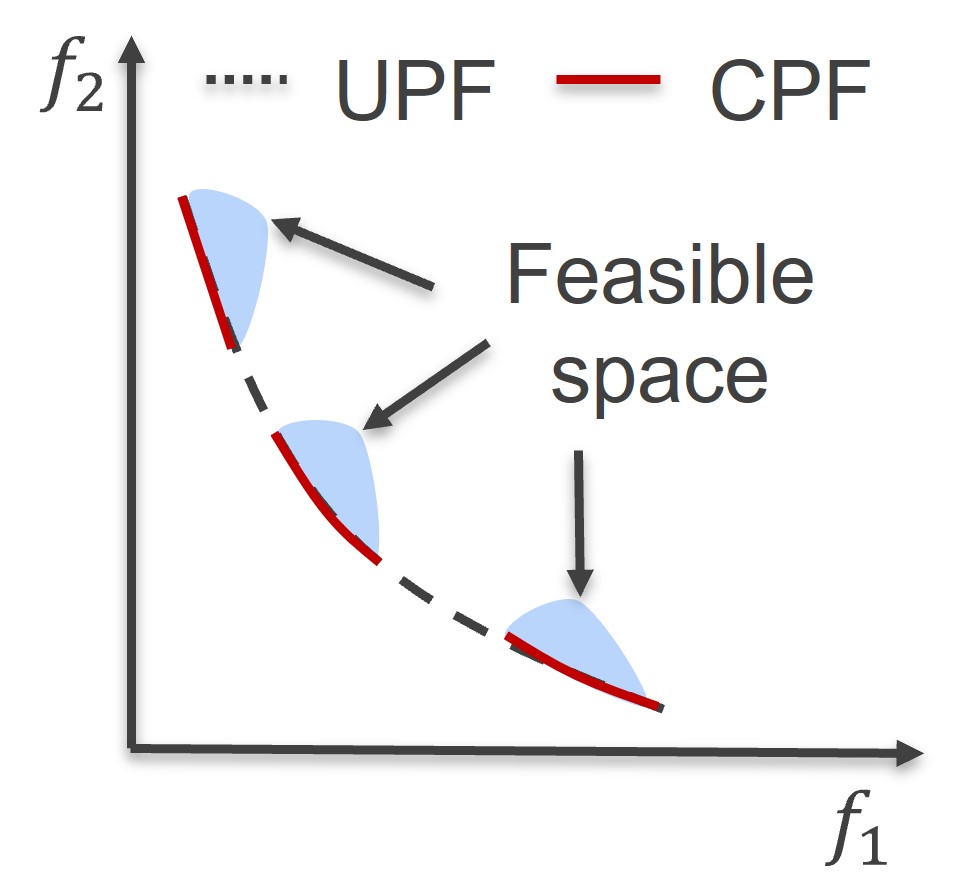}} \quad
\subfigure[]{\label{fig:PF3}\includegraphics[width=0.14\textwidth]{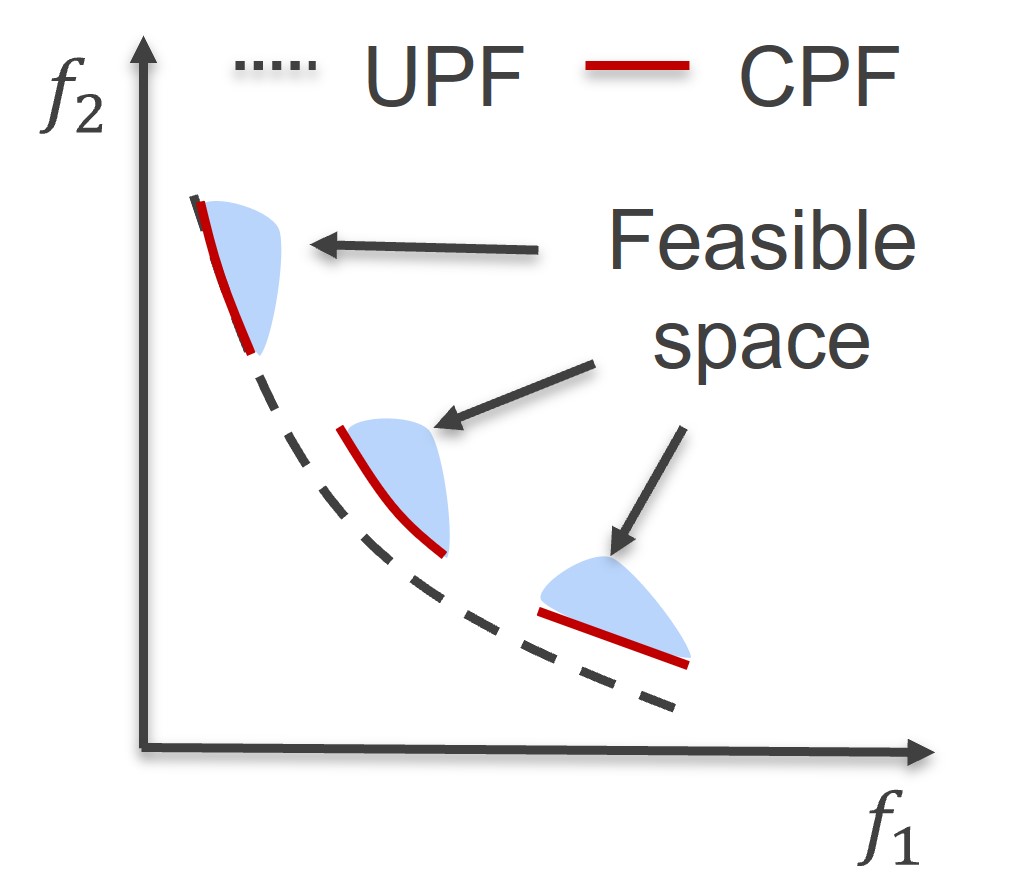}} \\
\vspace{-0.5em}
\subfigure[]{\label{fig:PF4}\includegraphics[width=0.14\textwidth]{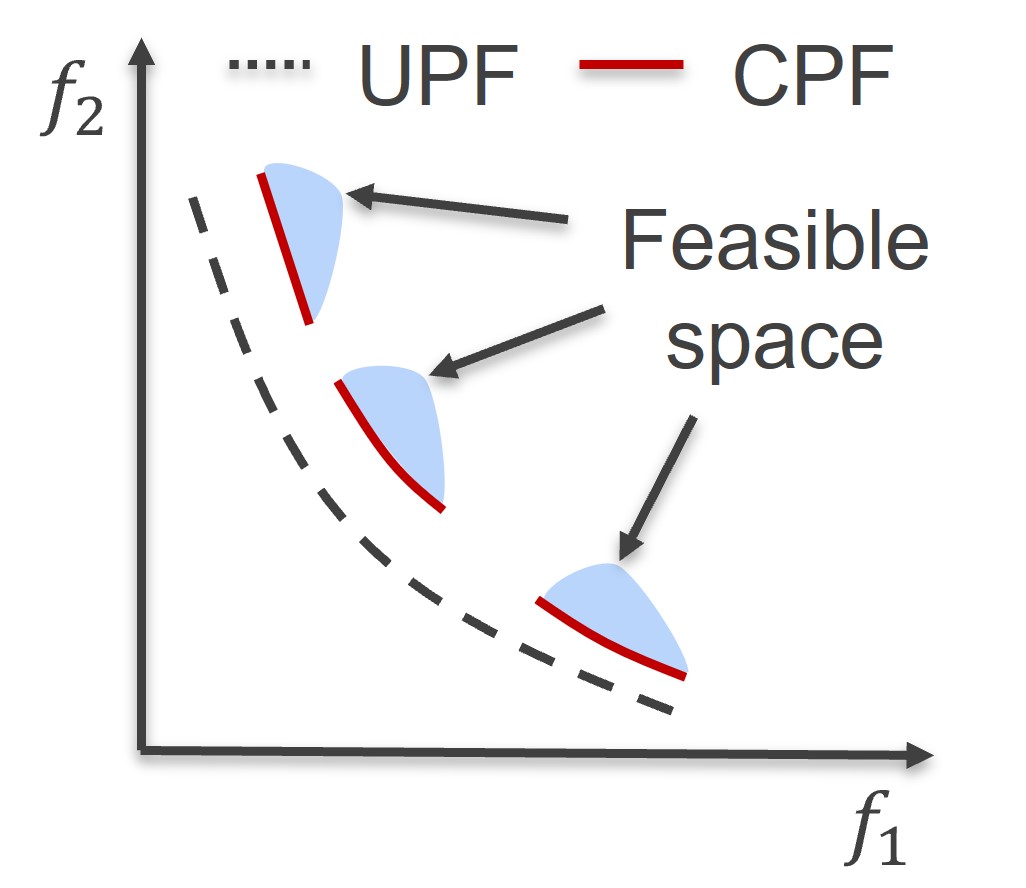}} \quad
\subfigure[]{\label{fig:PF5}\includegraphics[width=0.14\textwidth]{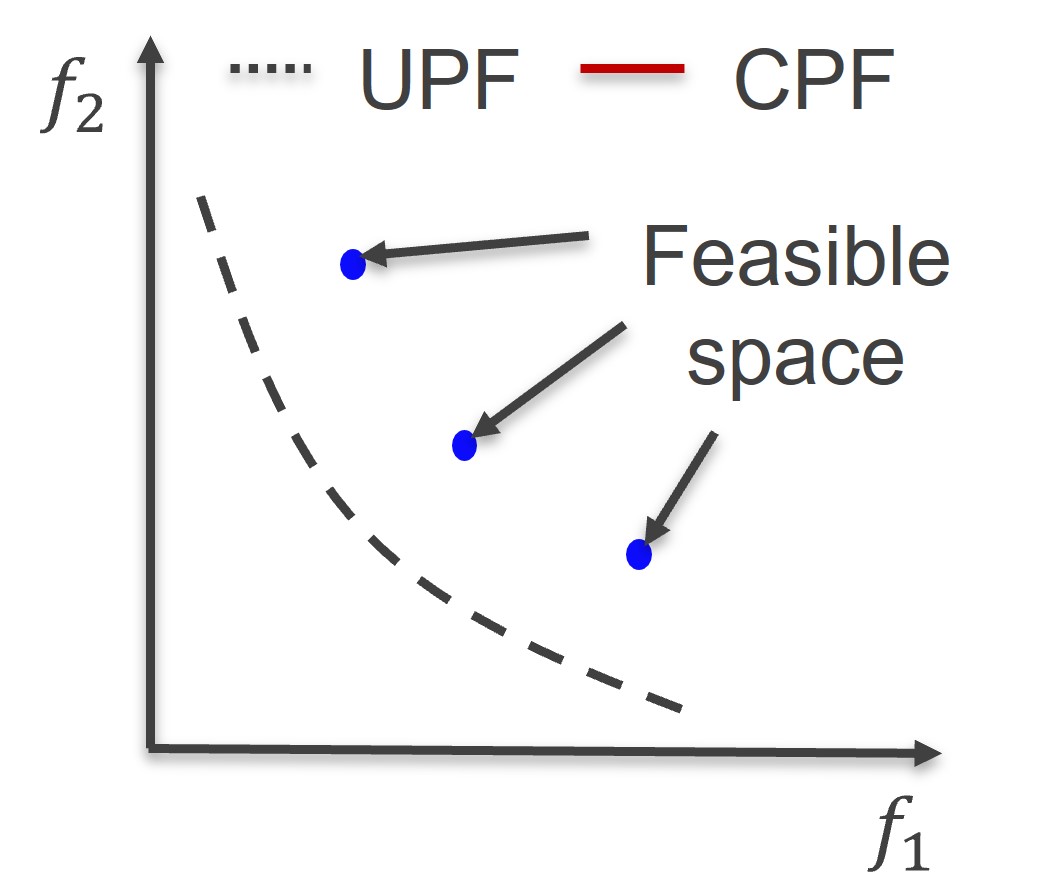}} \quad
\subfigure[]{\label{fig:PF6}\includegraphics[width=0.14\textwidth]{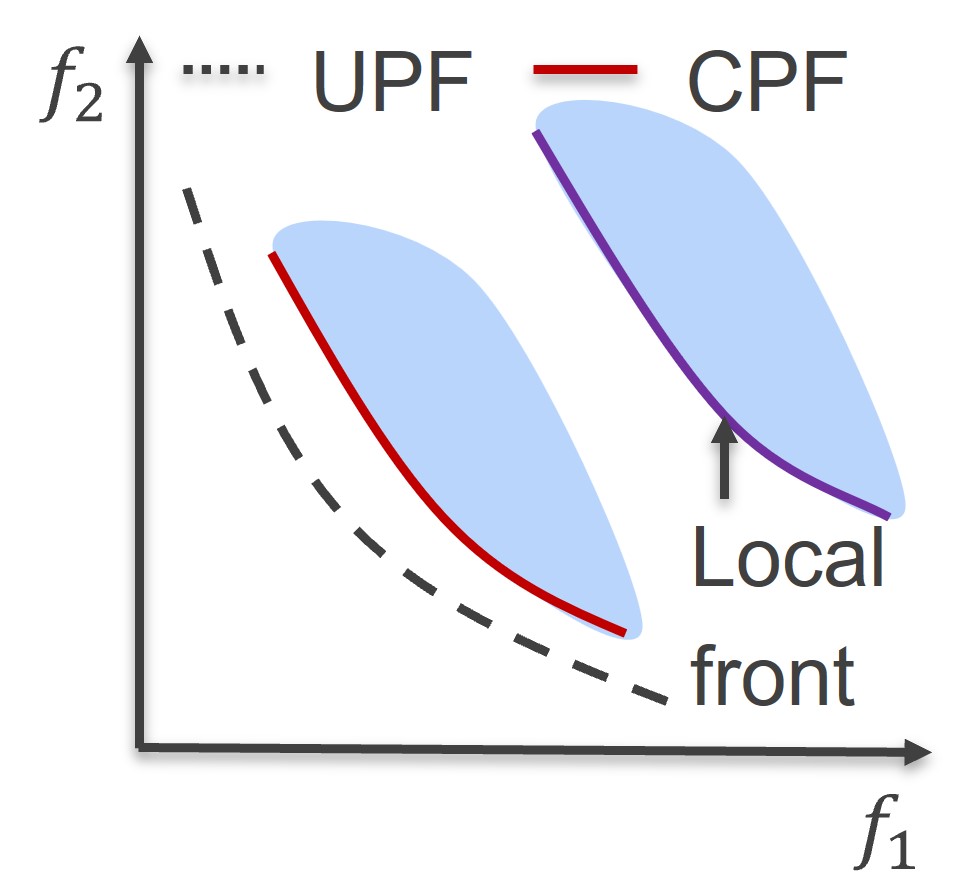}}
\caption{Different relative positions of UPF and CPF.}
\label{fig:PF}
\end{figure}

Some of the challenges in CMOPs can be understood by observing relative locations of UPF and CPF, and the geometry of CPF itself. A few common instances as represented in~\cite{liu2024learning} relevant to this study are schematically shown in Fig.~\ref{fig:PF}. Fig.~\ref{fig:PF1} represents a case where CPF and UPF are identical, making it easy for feasibility-first~(FF) and/or co-evolutionary based strategies given their higher focus on feasibility. However, for the instances in Fig.~\ref{fig:PF2} - \ref{fig:PF3}, enhanced diversity during evolutionary search is required since the UPF and CPF share disconnected segments. Emphasis on the FF or unconstrained search might bias the search in these scenarios while, strategies that preserve some infeasible solutions~\cite{ray2009infeasibility} might be more suitable to navigate the disconnected feasible regions, followed by intensification of search close to the CPF. In Fig.~\ref{fig:PF4}, the CPF and UPF are completely separated by infeasible region(s) between them where balancing convergence and feasibility is a considerable challenge. Therefore, co-evolutionary approaches are less effective since ignoring constraints may waste computing resources by driving the search towards UPF, especially when the gap between UPF and CPF is large. Fig.~\ref{fig:PF5} entails extremely relative small size of CPF and feasible regions, which is challenging for all CMOEAs because identification of the feasible region(s) is very difficult. In Fig.~\ref{fig:PF6}, convergence to the CPF can be restricted by the presence of disconnected feasible regions with local basins of attraction. Conventional CMOEAs may likely get stuck in such local fronts if the approach cannot navigate infeasible and local feasible regions effectively during evolution.

\begin{figure*}[!h]
\centering
\subfigure[MW11]{\label{fig:casemw11}\includegraphics[width=0.25\textwidth]{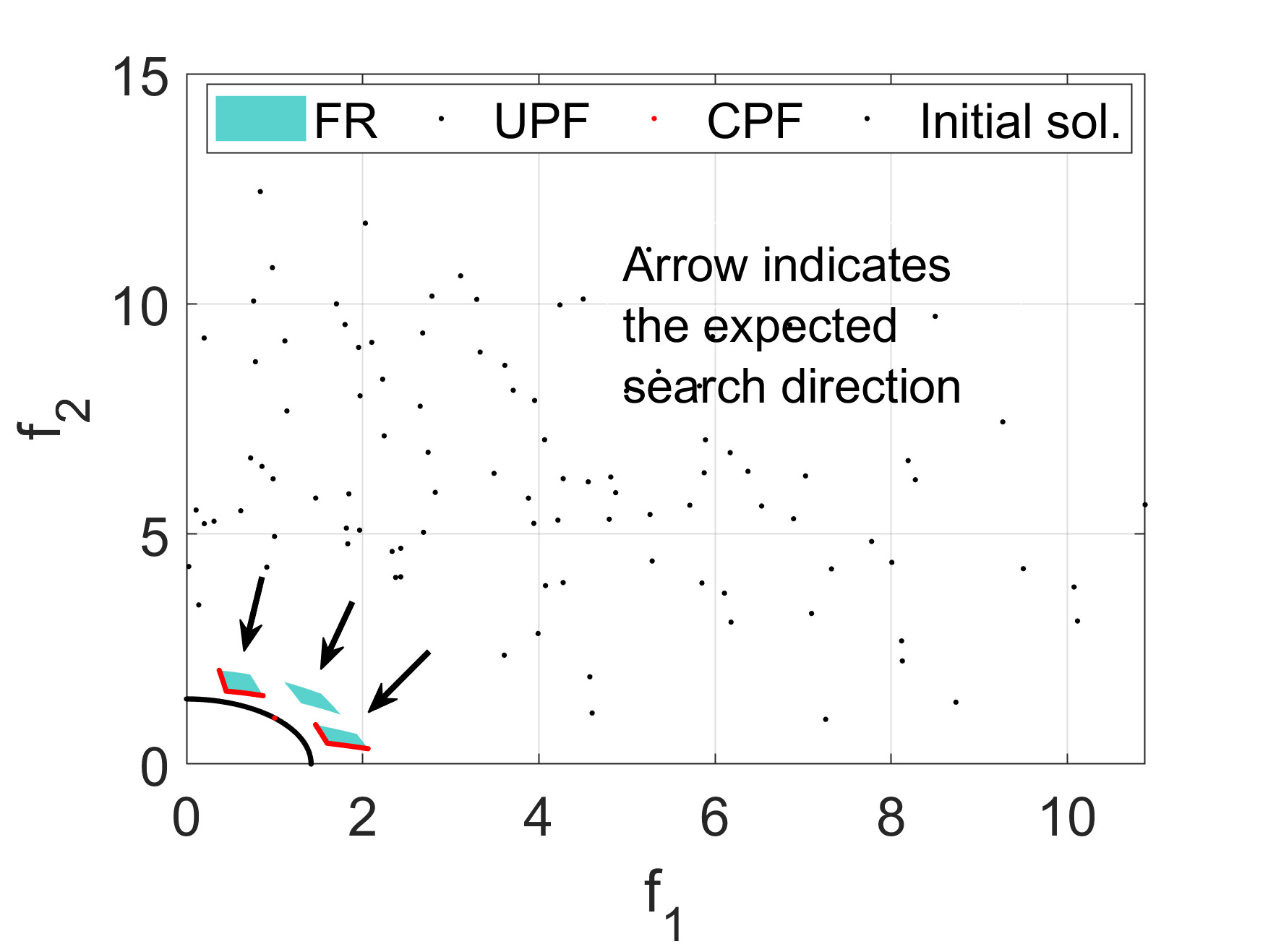}} \quad
\subfigure[KTS on MW11]{\label{fig:casemw11kts}\includegraphics[width=0.25\textwidth]{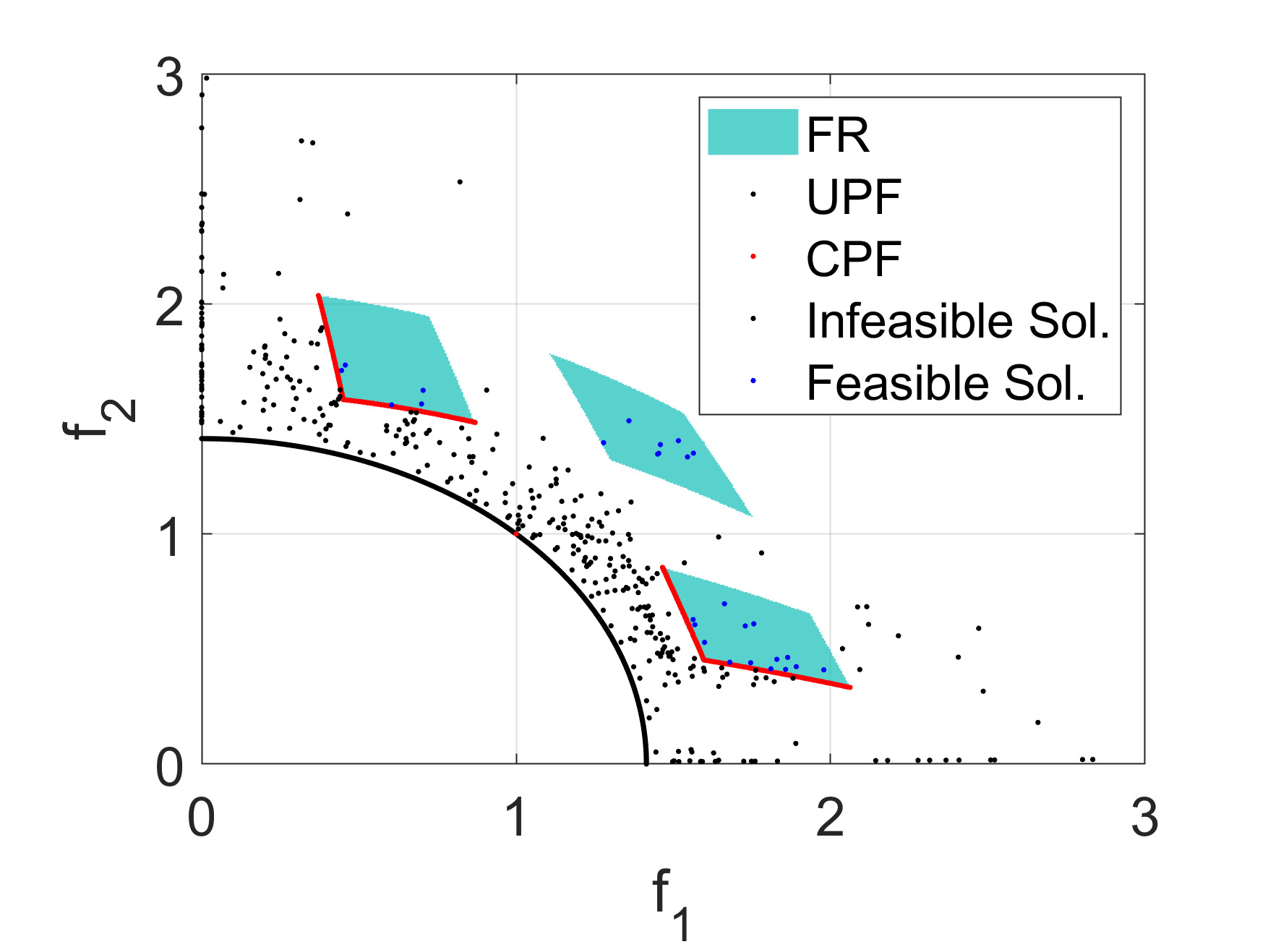}} \quad
\subfigure[KMGSAEA on MW11]{\label{fig:casemw11mg}\includegraphics[width=0.25\textwidth]{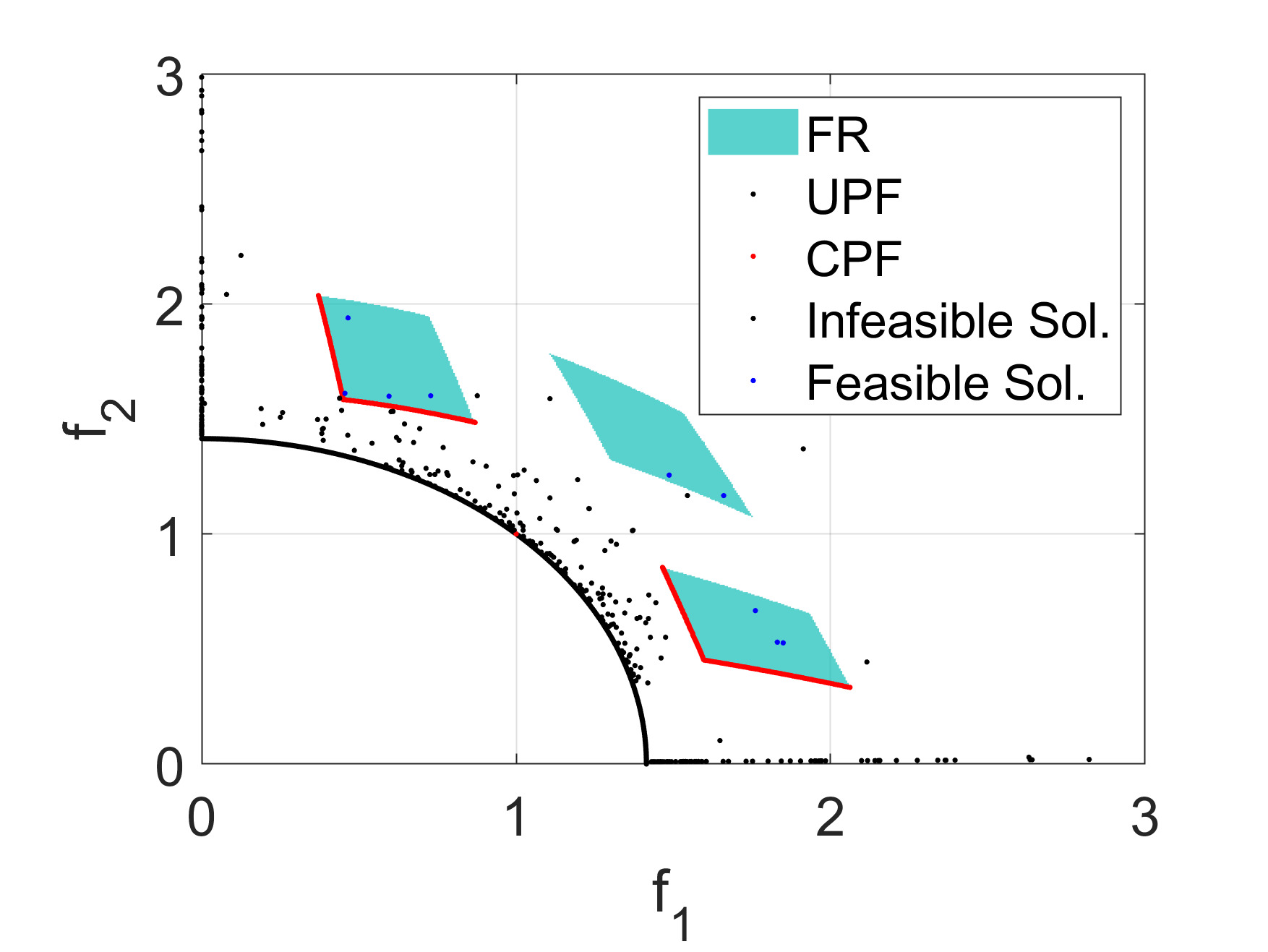}} \\
\vspace{-1em}
\subfigure[Test1]{\label{fig:casetest1}\includegraphics[width=0.25\textwidth]{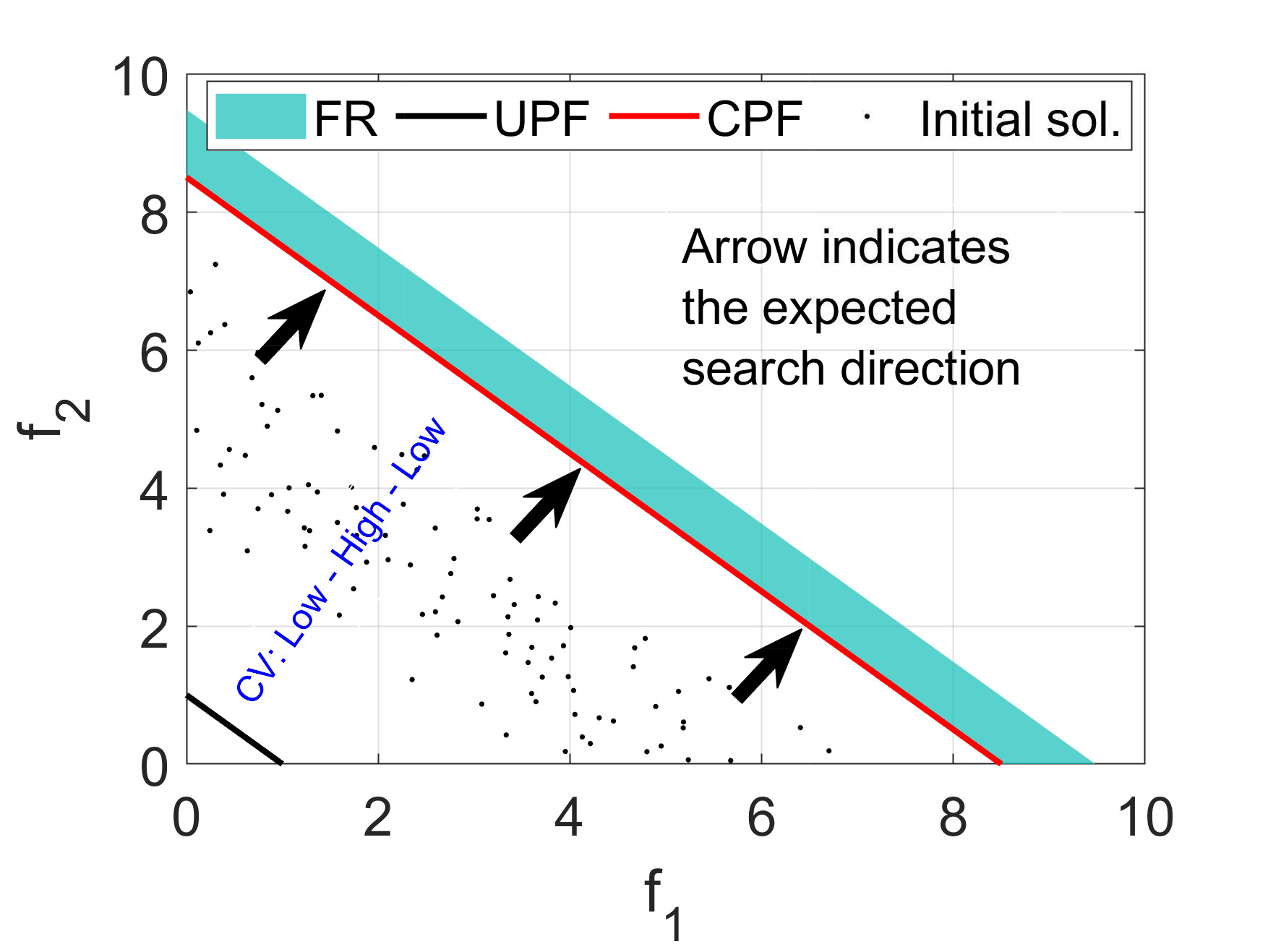}} \quad
\subfigure[KTS on Test1]{\label{fig:casetest1kts}\includegraphics[width=0.25\textwidth]{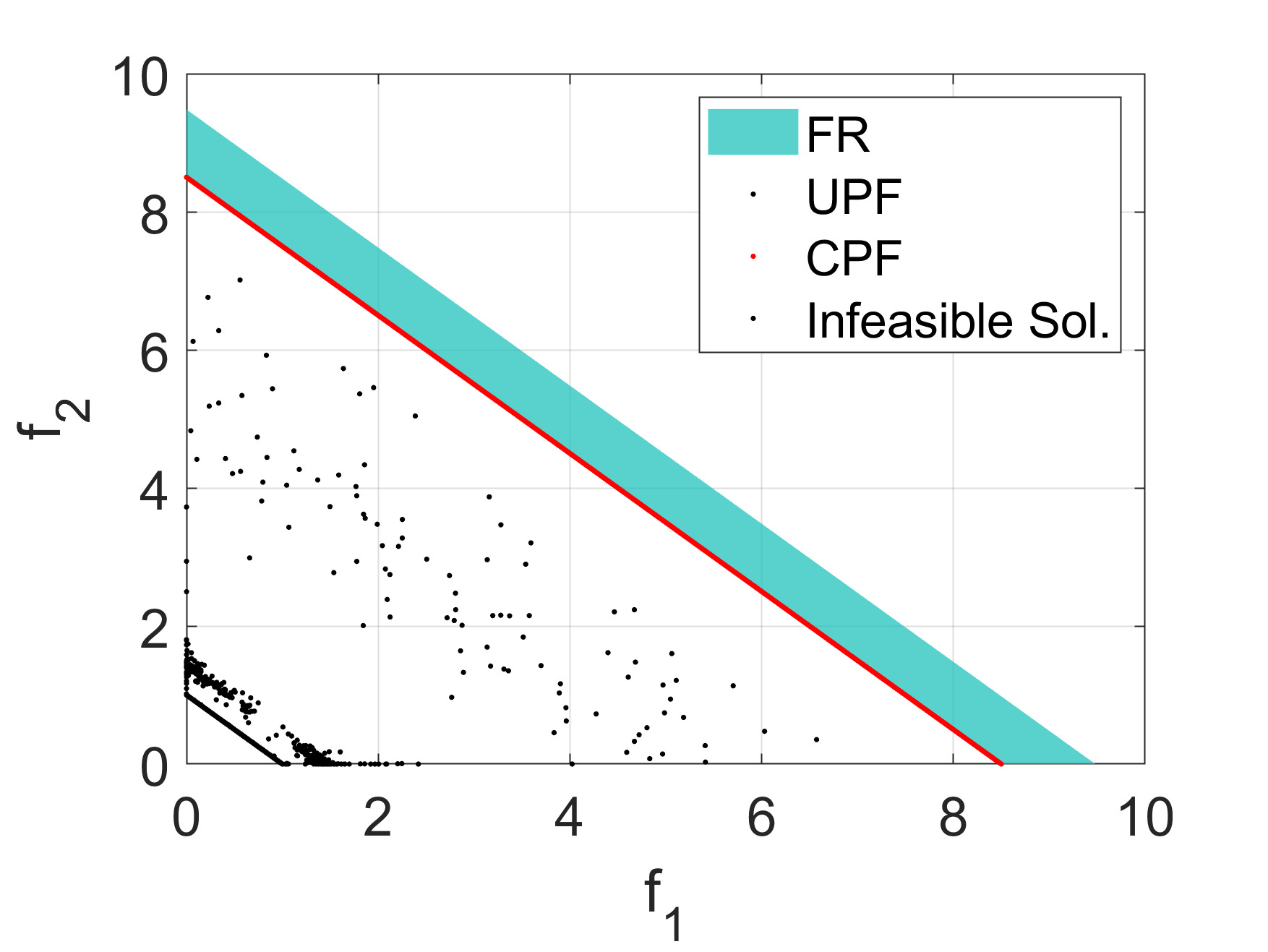}} \quad
\subfigure[KMGSAEA on Test1]{\label{fig:casetest1mg}\includegraphics[width=0.25\textwidth]{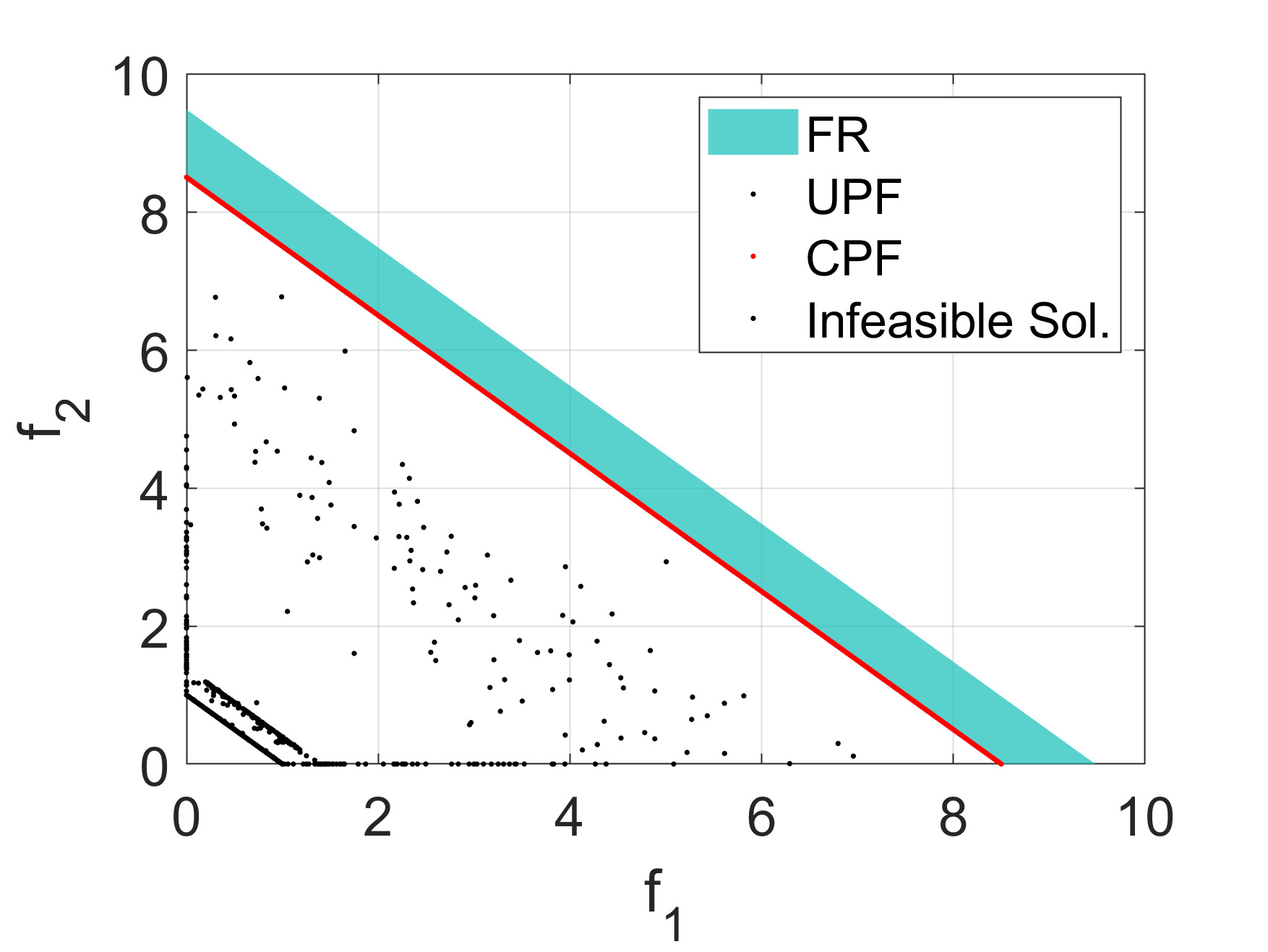}} \\
\vspace{-1em}
\subfigure[Test2]{\label{fig:casetest2}\includegraphics[width=0.25\textwidth]{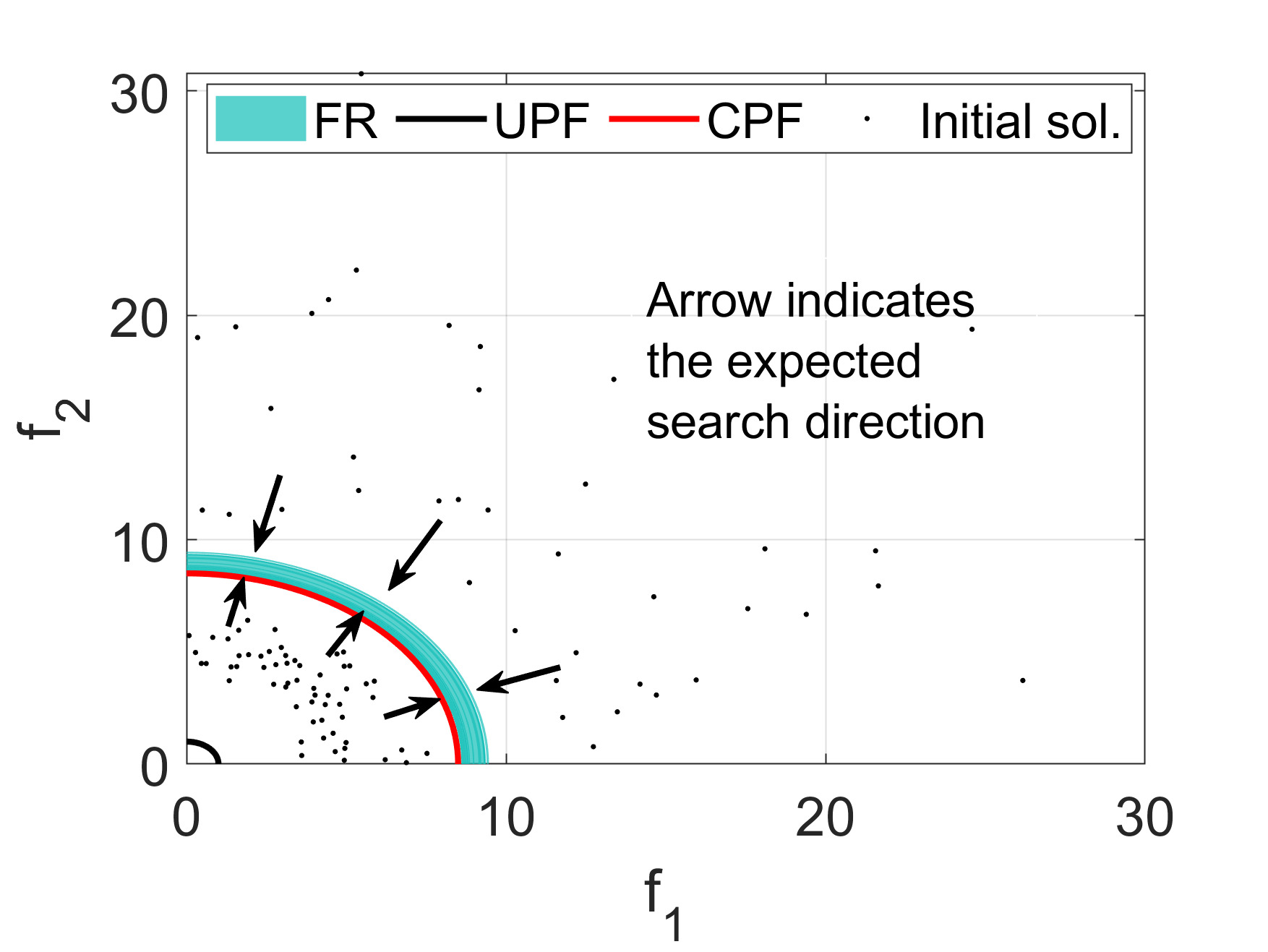}} \quad
\subfigure[KTS on Test2]{\label{fig:casetest2kts}\includegraphics[width=0.25\textwidth]{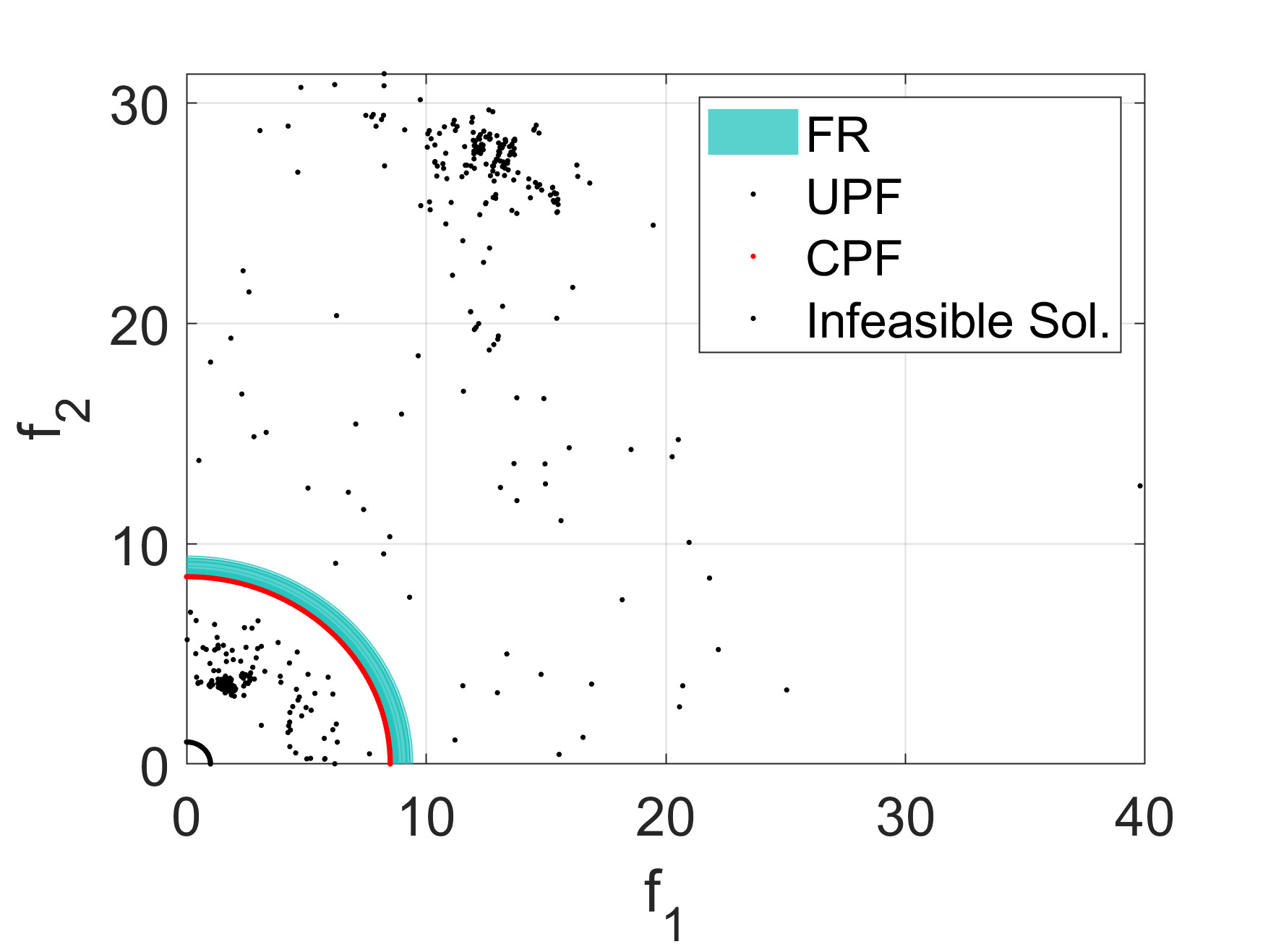}} \quad
\subfigure[KMGSAEA on Test2]{\label{fig:casetest2mg}\includegraphics[width=0.25\textwidth]{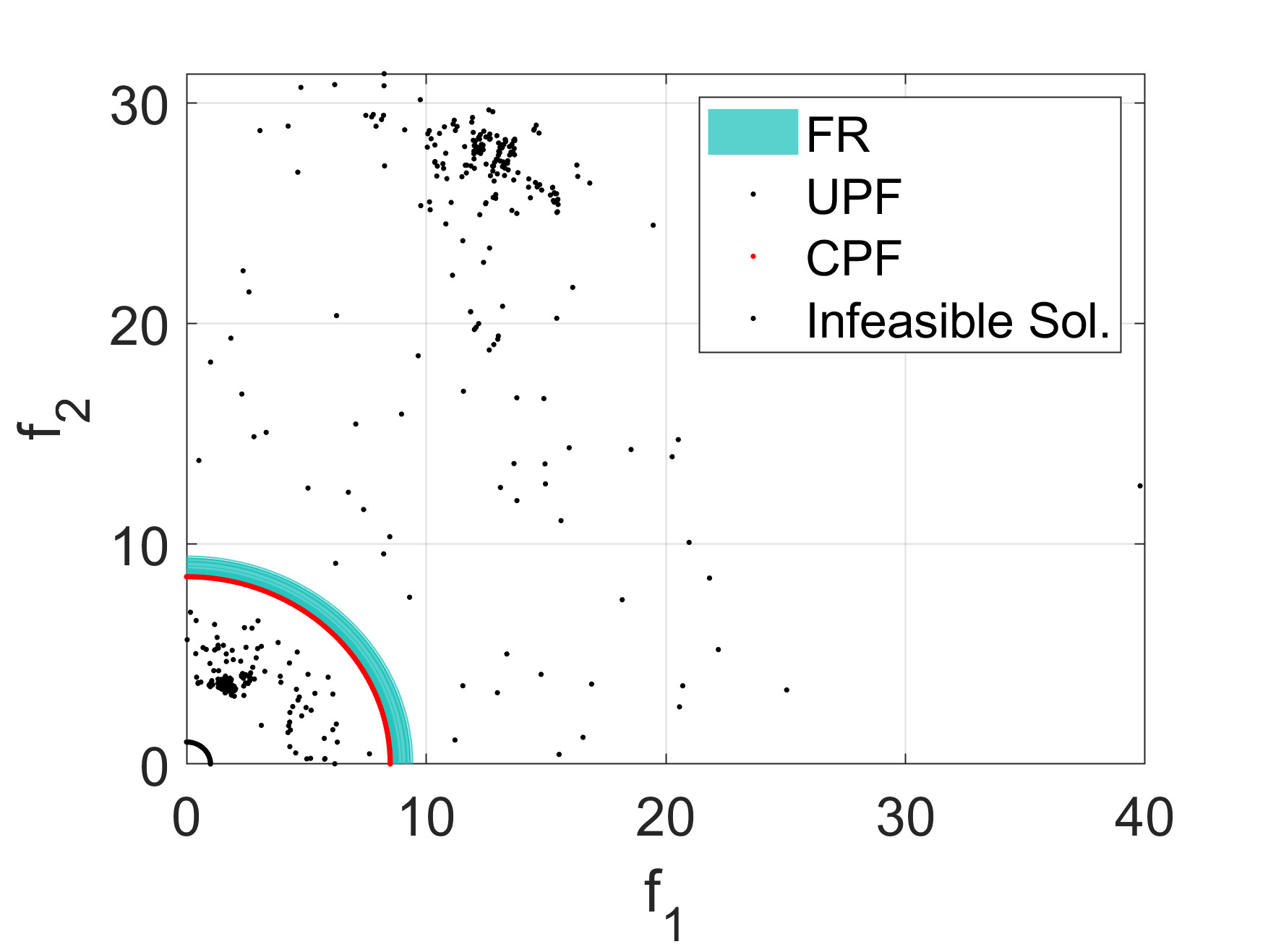}}
\caption{Test cases to illustrate the influence of the relative position of UPF and CPF on search performance of existing approaches.}
\label{fig:cases}
\end{figure*}

In order to solve CMOPs, it is critical to balance convergence, diversity, and feasibility of the solutions during the course of search. Although feasibility is important, discarding all infeasible individuals encountered during the search may significantly impair diversity, limiting exploration of promising regions. Conversely, retaining too many infeasible individuals risks guiding evolution away from the CPF. Besides these, ECMOPs have additional challenges due to limited the number of function evaluations. Although surrogates in SA-CMOEAs help to partially drive the solutions towards promising locations, the information they provide regarding feasibility, convergence and diversity of the candidate solutions is subject to their prediction uncertainties. If not carefully accounted for, the inaccurate information about the landscapes can adversely affect the search performance of the algorithms. The search difficulties encountered in some concrete instances of problems are demonstrated in Fig.~\ref{fig:cases}, taking two most recently proposed SA-CMOEAs, namely KTS~\cite{song2023balancing} and KMGSAEA~\cite{zhang2023multigranularity} as examples.    

Scenario 1 shows the case of MW11 test problem~\cite{ma2019evolutionary}, shown in Fig.~\ref{fig:casemw11}. The CPF comprises two disconnected feasible regions detached from the UPF by an infeasible region along with a solitary point on the surface of UPF. It also includes one disconnected feasible region yielding no CPF solution. A set of 100 solutions generated using Latin Hypercube Sampling~(LHS) is shown in Fig.~\ref{fig:casemw11}, which spans a large range in the infeasible region. The arrows indicate that both UPF and CPF lie in the same optimization direction~(i.e., correlation between ranks based on objectives and CV minimization is positive). This implies that co-evolutionary based approaches will swiftly identify feasible region on the way to the UPF via unconstrained search. Therefore, both KTS and KMGSAEA are able to identify a few feasible solutions at the end of computation budget~(500 evaluations are performed in this study) as shown in Fig.~\ref{fig:casemw11kts} - \ref{fig:casemw11mg}. However, it is also clearly noticeable that the density of converged solutions is much higher in UPF compared to the CPF for both cases. Due to the partial focus on unconstrained search, a number of evaluations are wasted even after identifying the feasible regions. This situation becomes even more pronounced in scenarios 2 and 3~(Fig.~\ref{fig:casetest1} and~\ref{fig:casetest2} respectively). These two test problems, referred to as Test1 and Test2, are constructed based on FCP series~\cite{yuan2021indicator} in which search is biased towards some local regions in the objective space. In Fig.~\ref{fig:casetest1}, all the initially generated 100 solutions are located in between UPF and CPF which means the expected search direction to achieve CPF is opposite to that of achieving the UPF. The CV is high in the middle region which gradually decreases towards both UPF and CPF. Now, an interesting thing can be observed from Fig.~\ref{fig:casetest1kts} - \ref{fig:casetest1mg} that the search is conducted totally opposite to the desired direction by both KTS and KMGSAEA. Similar situation can be observed for scenario 3 in Fig.~\ref{fig:casetest2}. In this problem, the initial solutions are generated on both sides of the CPF. Between UPF and CPF, there are a couple of infeasible regions with low CV values. As from the illustration in Fig.~\ref{fig:casetest2kts} - \ref{fig:casetest2mg}, the search is directed towards those infeasible regions by both KTS and KMGSAEA. All the scenarios demonstrate the significant impact that the relative locations of UPF and CPF have on the algorithm performance. Additionally, it is evident that discarding constraint information can sometimes lead to negative impact on the search while solving ECMOPs. 

Some of these shortcomings are aimed to overcome through PSCMOEA, which is intended to solve a diverse range of ECMOPs efficiently.

\section{Proposed approach}
\label{sec:algorithm}

\begin{figure}[!ht]
\centering    
\includegraphics[width=0.40\textwidth]{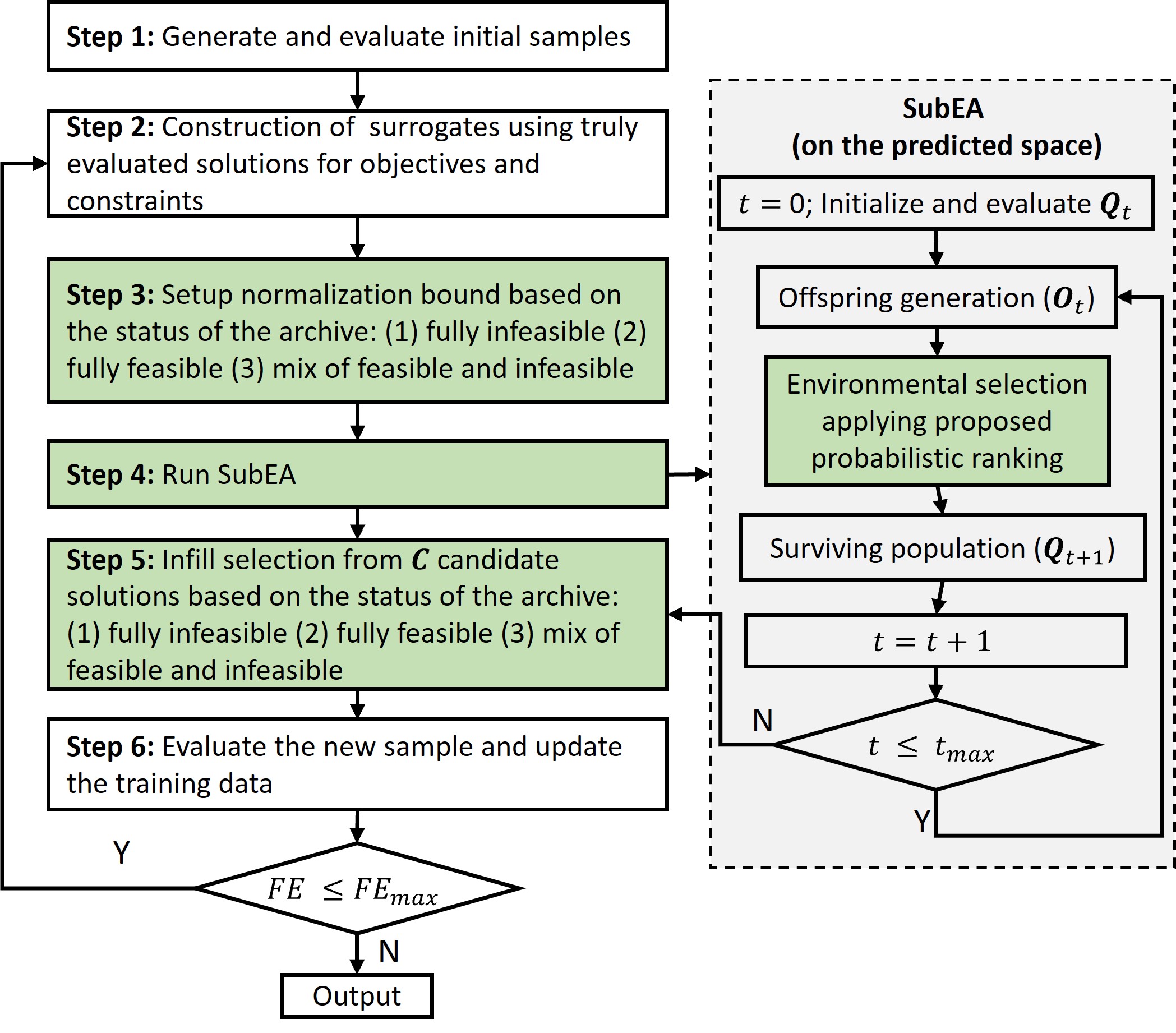} 
\caption{A general overview of the PSCMOEA framework. The green shaded boxes indicate the steps where this study introduces new contributions.}
\label{fig:flowchart}
\end{figure}

The general framework of PSCMOEA\footnote{For research purposes, Matlab implementation of PSCMOEA will be made available in its final form after manuscript review process is complete; to incorporate any suggested updates.} is shown in Fig.~\ref{fig:flowchart}, which, in brief, comprises the following steps. In \textbf{step 1}, the initial solutions are generated and evaluated, followed by training of Kriging surrogate models in \textbf{step 2}. In \textbf{step 3}, normalization bounds are computed based on the status of the existing archive of evaluated solutions, to conduct the decomposition-based search in \textbf{step 4}. The \textbf{step 4} executes a search on the surrogate space in order to identify promising candidate solutions. In \textbf{step 5}, the infill solution for true evaluation is selected from these promising candidates. The infill solution is evaluated in \textbf{step 6}, added to the archive and the process loops back to \textbf{step 2}, until the prescribed number of evaluations are exhausted. More details of these steps, along with rationale behind their design, are discussed in the following subsections. 

\begin{algorithm}[h]\footnotesize
\caption{PSCMOEA baseline framework} 
\begin{flushleft}
\algorithmicrequire \hspace{1mm} Maximum function evaluations $FE_{max}$, initial population size $N$, sparsity threshold~($\epsilon$), SubEA parameters: population size~($N_S$), number of generations~($G_S$), crossover and mutation probability~($P_c$ and $P_m$), crossover and mutation distribution index~(${\eta}_c$ and ${\eta}_m$), infeasible ratio for initial ranking of SubEA~($\alpha$), RVTagFlag: `ON' or `OFF', SearchFlag: `Constrained' or `Unconstrained'.\\
\textbf{Output:} Constrained Pareto front~(CPF) approximation.
\end{flushleft}
\begin{algorithmic}[1]
\STATE $\mathbf{X} \leftarrow$ \textbf{Initialize} $N$ solutions. \COMMENT{Following LHS strategy}
\STATE ${\left[\mathbf{F(x)}, \mathbf{G(x)}\right]}_{1:N} \leftarrow$ \textbf{Evaluate} ($\mathbf{X}$); $FE = |\mathbf{X}|$. 
\STATE $\mathbf{A} \leftarrow$ Store initially evaluated database: $\left[\mathbf{X},{\mathbf{F(x)}}_{1:N}, {\mathbf{G(x)}}_{1:N}\right]$.
\STATE Identify feasible~($\mathbf{A}_{\text{feas}}$) and infeasible~($\mathbf{A}_{\text{infeas}}$) solutions in $\mathbf{A}$.
\STATE Set RVTagFlag: `OFF' and SearchFlag: `Constrained'.
\IF{$\mathbf{A}_{\text{feas}} \neq \emptyset$} \label{line:gen1}
\STATE Initialize the shadow ND archive~($\mathbf{A^S}$). \COMMENT{Section~\ref{subsec:infill}}
\ELSE
\STATE $\mathbf{A^S} \longleftarrow \emptyset$. \COMMENT{Initial shadow ND archive is set empty}
\ENDIF \label{line:gen2}
\WHILE{$(FE \le FE_{max})$}
\STATE \textbf{Train} the Kriging model for each objective and constraint. \label{line:gen3} 
\STATE \textbf{Set} the normalization bounds $\mathbf{Z^I}$ and $\mathbf{Z^{N}}$. \COMMENT{Algo.~\ref{algo:norm}} \label{line:gen4}
\STATE $\mathbf{C} \leftarrow$ SubEA ($\mathbf{A}$, models, $\mathbf{Z^I}$, $\mathbf{Z^{N}}$, Flags). \label{line:gen5}
\STATE \textbf{New sample} $\leftarrow$ Infill Identification ($\mathbf{C}$, $\mathbf{A}$, $\mathbf{A^S}$, SearchFlag, RVTagFlag). \COMMENT{Algo.~\ref{algo:infill}} \label{line:gen6}
\STATE \textbf{Evaluate} the new sample with expensive functions.
\STATE Update archive~$\mathbf{A}$, shadow ND archive $\mathbf{A^S}$ and $FE = FE + 1$.
\STATE \textbf{Update} the RVTagFlag and SearchFlag 
\ENDWHILE

\end{algorithmic}
\label{algo:general}
\end{algorithm}

\subsection{Initialization and Kriging models}
\label{subsec:initialization}
The initialization process is quite similar to most canonical SA-CMOEAs. $N$ solutions are generated in the variable space based on LHS sampling and the corresponding objective and constraint functions are computed through true~(expensive) evaluations. This dataset is then used to build surrogate models. In this study, Kriging is used, also known as Gaussian Process~(GP) regression~\cite{martinez2016kriging}, as the underlying surrogate model for its ability to provide information about uncertainty along with mean predictions. Using the archive $\mathbf{A}$ of all evaluated solutions, a Kriging model is built for each individual objective and constraint function. To ensure model stability, input data is pre-screened by removing solutions closer than $1e^{-4}$ in the normalized variable space. The process of modeling and choice of hyperparameters is consistent with a number of recent studies~\cite{rahi2021partial,rahi2022steady} and the PlatEMO framework~\cite{tian2017platemo}, hence the details are not replicated here for brevity.

\subsection{Setting the normalization bounds}
\label{subsec:normalbound}

\begin{figure}[!ht]
\centering
\subfigure[]{\label{fig:bound1}\includegraphics[width=0.15\textwidth]{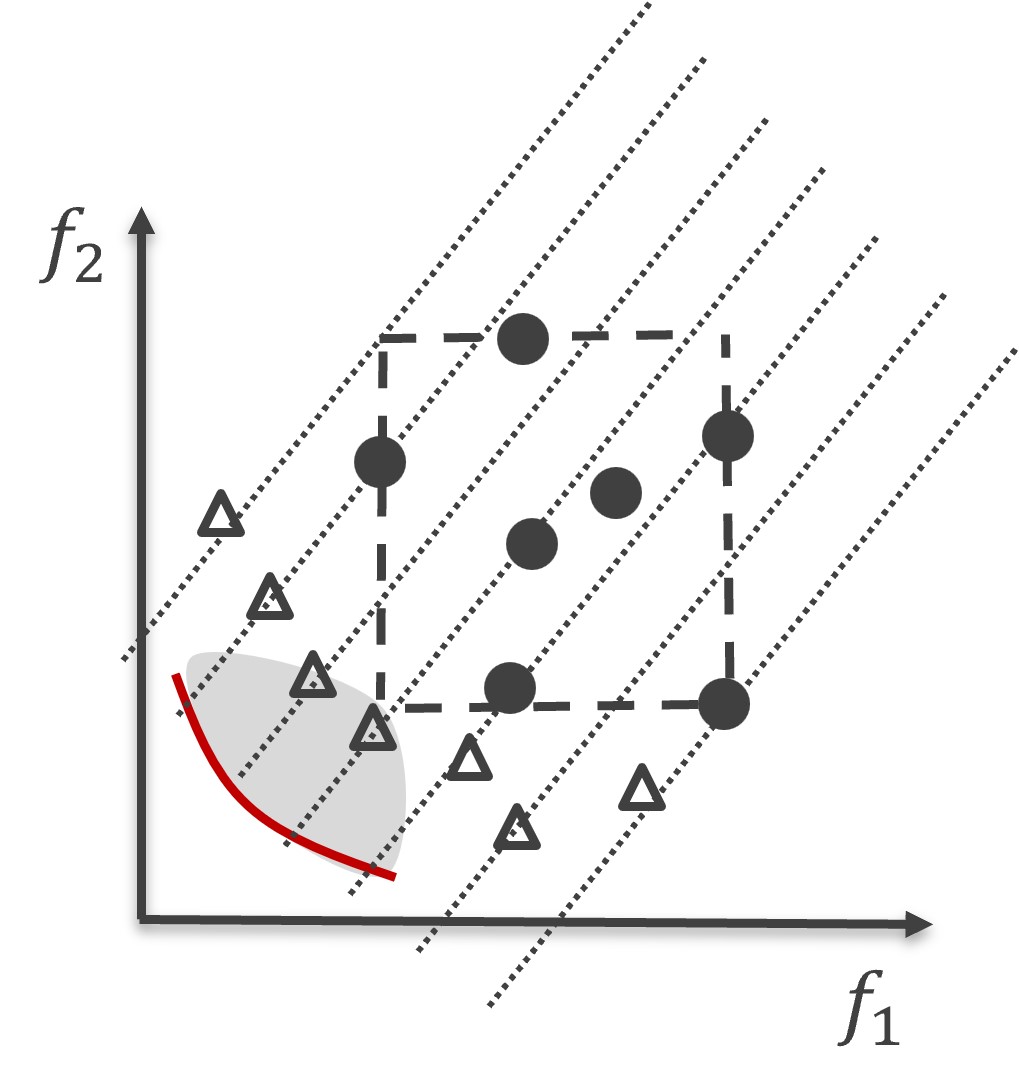}}
\subfigure[]{\label{fig:bound2}\includegraphics[width=0.15\textwidth]{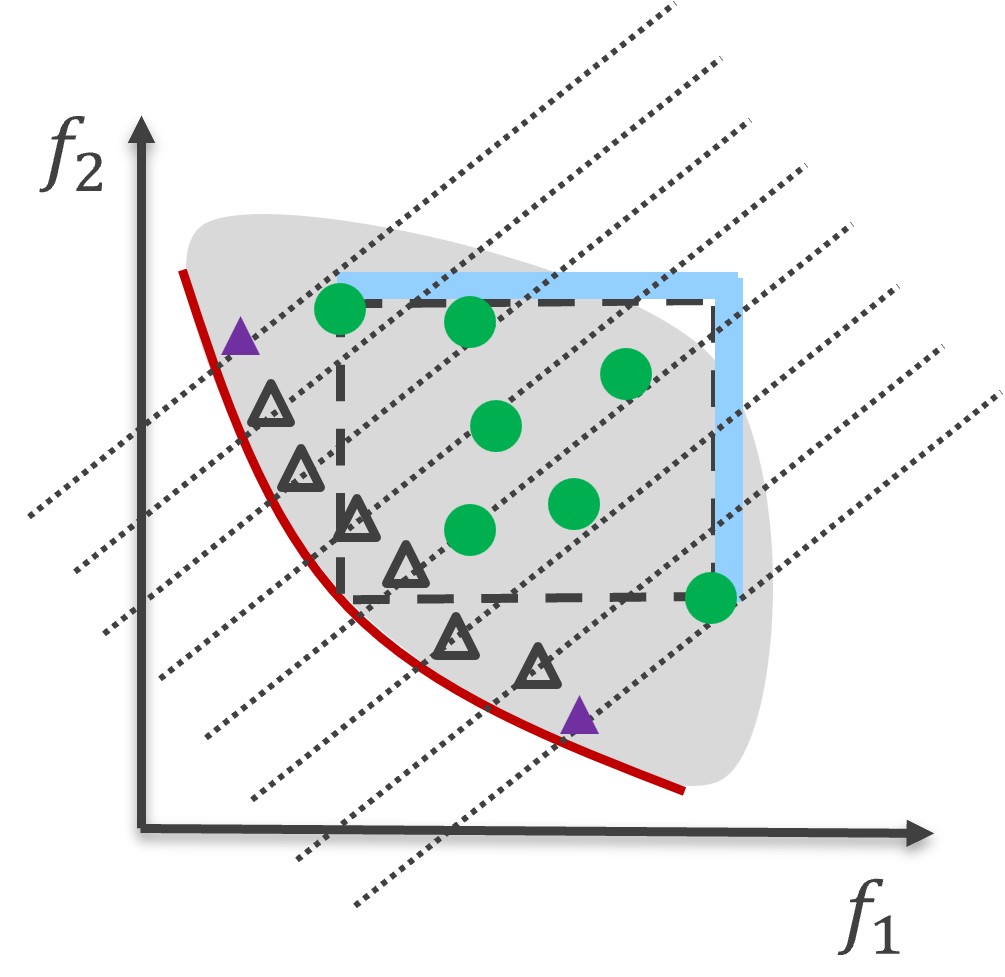}} 
\subfigure[]{\label{fig:bound3}\includegraphics[width=0.15\textwidth]{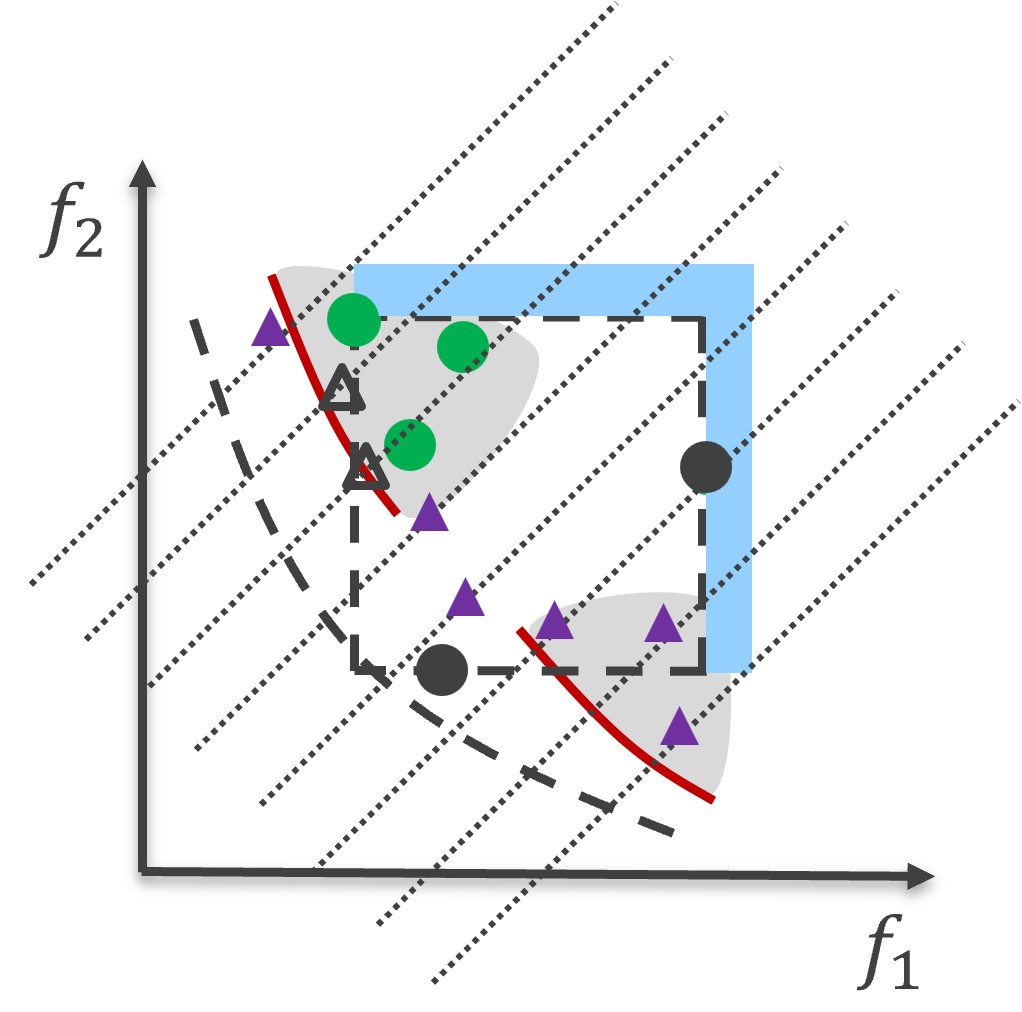}}
\caption{Normalization bound when archive is (a) fully infeasible, (b) fully feasible and (c) mix of feasible and infeasible. The gray shaded region denotes feasible region. The bluish shaded region in (b) and (c) denotes the extended bounds. The green and black circular dots represents true feasible and infeasible solutions respectively. The triangular points represent the converged predicted candidate set along the RVs.}
\label{fig:bounds}
\end{figure}

Normalization plays a crucial role in MOEAs by ensuring a fair basis for comparison when dealing with objectives that cover significantly different numerical ranges~\cite{he2021survey}. Normalization is particularly important in decomposition-based MOEAs, so that the uniformly distributed reference vectors~(RVs) could map to uniform objectives irrespective of their ranges. However, setting suitable bounds for normalization is challenging the when true ideal~($\mathbf{Z^I}$) and Nadir~($\mathbf{Z^N}$) points are unknown. On the other hand, setting bounds by only considering the feasible ND solutions in the archive~$\mathbf{A}$ could make the search overly conservative, diminishing the likelihood of identifying competitive solutions beyond the current range~\cite{rahi2022steady,wang2021investigating}. Another open question is what bounds to use when there is no feasible solution in the archive? In this context, a new and adaptive way of setting the normalization bounds is introduced based on the feasibility status of $\mathbf{A}$ at the beginning of each generation. The bounds will be utilized during search on the surrogate space~(will be detailed in the next subsection). The process of identifying the normalization bounds is outlined in Algo.~\ref{algo:norm} and further illustrated through Fig.~\ref{fig:bounds}. 

\begin{algorithm}[ht]\footnotesize
\caption{Setting normalization bound}
\begin{flushleft}
\algorithmicrequire \hspace{1mm}Archive $\mathbf{A}$. \\
\textbf{Output:} Normalization bounds, $\mathbf{Z^I}$ and $\mathbf{Z^N}$.
\end{flushleft}
\begin{algorithmic}[1]
\STATE Identify feasible~($\mathbf{A}_{\text{feas}}$) and infeasible~($\mathbf{A}_{\text{infeas}}$) solutions in $\mathbf{A}$ and take their corresponding objective values $\mathbf{F}_{\text{feas}}$ and $\mathbf{F}_{\text{infeas}}$ respectively.
\IF{$\mathbf{A}_{\text{feas}} == \emptyset$}
\STATE $\mathbf{Z^I}$ = $\text{min}{\{{\mathbf{F}}^i_{\text{infeas}}\}}^{M}$; $\mathbf{Z^N}$ = $\text{max}{\{{\mathbf{F}}^i_{\text{infeas}}\}}^{M}$. \COMMENT{i = 1,\ldots,M} \label{line:norm1}
\ELSIF{$\mathbf{A}_{\text{infeas}} == \emptyset$} \label{line:norm2}
\STATE \textbf{Identify} ${\mathbf{Fnd}}_{\text{feas}}$ from $\mathbf{A}_{\text{feas}}$. \COMMENT{Feasible ND set in $\mathbf{A}_{\text{feas}}$}
\STATE $\mathbf{Z^I}$ = $\text{min}{\{{\mathbf{Fnd}}^i_{\text{feas}}\}}^{M}$; $\mathbf{Z^N}$ = $\text{max}{\{{\mathbf{Fnd}}^i_{\text{feas}}\}}^{M}$. \COMMENT{i = 1,\ldots,M}
\STATE \textbf{Extend} $\mathbf{Z^N}$ following it: $\mathbf{Z^N}$ = $\mathbf{Z^N}$ + $\left(\left(\mathbf{Z^N} - \mathbf{Z^I}\right)\times 0.1\right)$. \label{line:norm3}
\ELSE
\STATE \textbf{Identify} ${\mathbf{Fnd}}_{\text{feas}}$ from $\mathbf{A}_{\text{feas}}$. \label{line:norm4}
\STATE \textbf{Identify} and \textbf{list} the infeasible solution(s) which are ND w.r.t any ${\mathbf{Fnd}}_{\text{feas}}$, presented as ${\mathbf{F}}^L_{\text{infeas}}$.
\STATE \textbf{Combine} $\mathbf{F}_{\text{comb}} \longleftarrow \{{\mathbf{Fnd}}_{\text{feas}}, {\mathbf{F}}^L_{\text{infeas}}\}$.  
\STATE $\mathbf{Z^I}$ = $\text{min}{\{{\mathbf{F}}^i_{\text{comb}}\}}^{M}$; $\mathbf{Z^N}$ = $\text{max}{\{{\mathbf{F}}^i_{\text{comb}}\}}^{M}$. \COMMENT{i = 1,\ldots,M}
\STATE \textbf{Extend} $\mathbf{Z^N}$ as: $\mathbf{Z^N}$ = $\mathbf{Z^N}$ + $\left(\left(\mathbf{Z^N} - \mathbf{Z^I}\right)\times 0.1\right)$. \label{line:norm5}
\ENDIF
\end{algorithmic}
\label{algo:norm}
\end{algorithm}

From Section~\ref{subsec:motive}, it could be inferred that search bounds should adapt in different stages of the search. For example, in the beginning when no feasible solution has been identified, priority has to be given to the exploration when searching for a new candidate solution in the predicted fitness landscape~(normalized in terms of current $\mathbf{Z^I}$ and $\mathbf{Z^N}$). To ensure this, the complete available original objective space is used for searching by calculating the minimum~($\mathbf{Z^I}$) and maximum~($\mathbf{Z^N}$) numerical values of each objective in $\mathbf{A}$~(Algo.~\ref{algo:norm}, line~\ref{line:norm1}). As shown in Fig.~\ref{fig:bound1}, although the initial true solutions are still outside the feasible region, the RVs generated based on the current $\mathbf{Z^I}$ and $\mathbf{Z^N}$ cover the available feasible region and intersect the CPF uniformly, leading to solutions with a good spread in predicted space~(denoted with triangles). The second type includes the situation when the full $\mathbf{A}$ is feasible~(Algo.~\ref{algo:norm}, line~\ref{line:norm2} - \ref{line:norm3}). In this case, $\mathbf{Z^I}$ and $\mathbf{Z^N}$ are calculated by taking the minimum and maximum of feasible non-dominated~(ND) objective values respectively. After that, $\mathbf{Z^N}$ is modified by adopting a boundary extension technique introduced in \cite{rahi2022steady}. It has shown efficacy to identify  well distributed solutions on the PF regardless of the shape~(regular/irregular)~\cite{rahi2022steady}. As illustrated in Fig.~\ref{fig:bound2}, the rectangle with dashed line indicates the current bound which is extended following  equation in Algo.~\ref{algo:norm}, line~\ref{line:norm3}, shown by the blue shaded region. Such a simple extension promotes diversity which increases the likelihood of covering the whole CPF. For example, the two candidates along two extreme RVs represented by magenta colored triangle can be obtained only through this extension. The third case denotes a more frequent scenario encountered by the SA-CMOEAs in which $\mathbf{A}$ contains both feasible and infeasible solutions. In this case, search has to be directed in a way to explore more feasible regions~(if present) and at the same time, to sample well distributed solutions on the CPF of the already identified feasible region. Setting the normalization bound at this stage only based on the infeasible solutions~(as case 1) would result in misdirected evaluations by searching in an unreasonably wide area. On the other hand, considering the bounds based only on feasible ND solutions might bias the search towards one of multiple disconnected regions shown in Fig.~\ref{fig:bound3}. Therefore, a novel strategy is proposed by maintaining the trade-off between convergence and diversity in which the high quality infeasible solutions in terms of the objective values are considered alongside the feasible ND solutions to construct the bounds~(Algo.~\ref{algo:norm}, line~\ref{line:norm4} - \ref{line:norm5}). The list of infeasible solutions which are ND to any feasible ND solution in terms of objectives are added to the feasible ND set to construct the combined set. After that, the minimum and maximum values of each objective in this combined set are considered as the $\mathbf{Z^I}$ and $\mathbf{Z^N}$ respectively, followed by the extension of the $\mathbf{Z^N}$. The benefit of the proposed strategy can be clearly noticed from Fig.~\ref{fig:bound3}. By considering the high quality~(ND) infeasible solutions, the search can cover both the feasible regions. The magenta colored solutions are only obtained~(in the predicted space) using the proposed normalization bound. Using these bounds, the predicted value~($\boldsymbol\mu$) and the associated uncertainty~($\boldsymbol\sigma$) of the solution are scaled using Eq.~\ref{eqn:norm} for the subsequent steps.

\begin{equation}\footnotesize
\begin{array}{lr}
\mu^{'}_i(\mathbf{x}) = \frac{\mu_i(\mathbf{x}) - Z^I_i}{Z^{N}_i - Z^I_i}; \sigma^{'}_i(\mathbf{x}) = \frac{\sigma_i(\mathbf{x})}{{(Z^{N}_i - Z^I_i)}^2};i = 1,\ldots,M
\end{array}
\label{eqn:norm}
\end{equation}

\subsection{SubEA to identify promising candidate solutions}
\label{subsec:SubEA}

As discussed in the overview and outlined in Algo.~\ref{algo:general}, line~\ref{line:gen5}, a SubEA is used to search the surrogate to identify a promising candidate pool for infill identification in the subsequent step. A decomposition-based evolutionary approach is employed for this purpose, with \emph{mirrored} reference vectors that have been shown to perform well for irregular shaped Pareto fronts~\cite{elarbi2019approximating,rahi2022steady}. An illustration of mirrored RVs can be seen in Fig.~\ref{fig:bounds}, represented by black dotted lines. These are formed by connecting the uniformly generated reference points on the unit simplex using Normal Boundary Intersection~(NBI) method with their corresponding mirrored points~(which will lie in third quadrant for Fig.~\ref{fig:bounds}). The details of SubEA are given below, of which the environmental selection using probabilistic constraint dominance is where this work makes the key contribution, and hence is discussed in more detail. Note that all fitness values used in SubEA undergo normalization using the bounds identified previously~(Algo.~\ref{algo:norm}). Furthermore, as SubEA operates on the surrogate landscape, it doesn't involve any expensive evaluations.

\subsubsection{Initialization and evaluation}
\label{subsubsec:initial}
In this step, a predefined number of samples~(population size, $N_S$) are 
 required as the initial population for surrogate assisted search. For enhanced seeding~\cite{gong2023effects}, part of the truly evaluated solutions in $\mathbf{A}$ are inherited directly into this population. For this, the solutions in $|\mathbf{A}|$ are ordered using infeasibility-driven ranking strategy~\cite{ray2009infeasibility} by maintaining recommended $20\%$ marginally infeasible solutions on top of the feasible ones. The top $N_S$ solutions are kept if $|\mathbf{A}|> N_S$. Otherwise, all solutions of $\mathbf{A}$ are kept followed by the remaining ones generated using LHS sampling. The performance of any newly generated solution(s) is estimated using the surrogate models. 

\subsubsection{Offspring generation}
\label{subsubsec:evolution}
Offspring solutions are evolved from the parent population using simulated binary crossover~(SBX) and polynomial mutation~(PM)~\cite{deb2002fast}. If any of these offspring solutions turn out identical to either the parent population and/or any truly evaluated solution in the archive, the offspring is substituted with a new randomly generated solution. Two solutions are deemed identical if the Euclidean distance computed between them in the normalized variable space is less than a tolerance of $1e^{-4}$.

\subsubsection{Environmental selection}
\label{subsubsec:selection}

The SubEA follows a population based approach, wherein $N_S$ new samples are evolved and evaluated~(using surrogates) in each generation. Out of the $2N_S$~(parent+offspring) solutions, denoted as $\mathbf{Q}$, $N_S$ are retained for the next generation through environmental selection. The selection process discussed below is developed through careful considerations of the surrogate predictions and associated uncertainties of the objectives and constraints, and forms one of the key contributions of this study. The pseudo-code of the selection scheme is outlined in Algo.~\ref{algo:ranking}. 
 
The mean and uncertainty of the response functions for the $2N_S$ solutions are denoted as $\boldsymbol\mu$ and $\boldsymbol\sigma$, respectively, estimated from Kriging models. Given the set $\mathbf{W}$ of RVs, each of these $2N_S$ solutions are assigned to its closest RV based on minimum angle~(denoted as $\sphericalangle$) assignment~\cite{cheng2016reference} in the predicted objective space. For this, the objective distributions of $2N_S$ solutions are scaled using Eq.~\ref{eqn:norm}. This assignment may lead to one or more solutions being associated with a single RV~(referred to as active RV), while some RVs may not have any solution associated with them~(inactive RV).

\begin{algorithm}[ht]\footnotesize
\caption{Environmental selection method for SubEA}
\begin{flushleft}
\algorithmicrequire \hspace{1mm}Population to order~$\mathbf{Q}$~(of size $\approx2N_S$), $\mathbf{Z^I}$, $\mathbf{Z^N}$, SearchFlag. \\
\textbf{Output:} A set of solutions $N_W$, one assigned to each reference vector. 
\end{flushleft}
\begin{algorithmic}[1]
\STATE Normalize the objective vectors in $\mathbf{Q}$ using the $\mathbf{Z^I}$ and $\mathbf{Z^N}$.
\STATE Generate reference vectors~(RVs)~$\mathbf{W}$ and set $n_W = N_W = |W|$.
\WHILE{$n_W \neq \emptyset$}
\STATE Assign each solution to its closest RV using angle as the proximity metric: $\sphericalangle_i(\boldsymbol\mu_i,\mathbf{W})$; $i = 1,\ldots,\mathbf{Q}$.  
\STATE Project the assigned solutions along the corresponding RV.
\IF{SearchFlag == `Constrained'}
\STATE Along each RV, compare each solution pairwise with all others according to Eq.~\ref{eqn:P_AB}. The average of the pairwise scores yields the final score of the corresponding solution.
\ELSIF{SearchFlag == `Unconstrained'}
\STATE Along each RV, compare each solution pairwise with all others according to Eq.~\ref{eqn:PD}. The average of the pairwise scores yields the final score of the corresponding solution. 
\ENDIF
\STATE Select the solution with the highest score along each RV. 
\STATE Remove the selected solutions and the RVs~(which had at least one solution assigned) from $\mathbf{Q}$ and ~$\mathbf{W}$ respectively.
\STATE $n_W = |W|$
\ENDWHILE
\end{algorithmic}
\label{algo:ranking}
\end{algorithm}

For comparing solutions assigned to each~(active) RV~($\hat{\mathbf{W}_j}$; $\{j = 1,\ldots,N_W (= \text{No. of RVs})\}$) an integrated performance criterion is proposed, denoted \emph{probability of constrained domination~($PCD$)} shown in Eq.~\ref{eqn:P_AB}, which quantitatively estimates the probability of a solution $\mathbf{x}$ being better than solution $\mathbf{y}$. PCD comprises of three terms, capturing three scenarios respectively: (a) $\mathbf{x}$ is feasible and $\mathbf{y}$ is not; (b) both are feasible and $\mathbf{x}$ dominates $\mathbf{y}$ in terms of objective values; and (c) both are infeasible and $\mathbf{x}$ has a lower $CV$ than $\mathbf{y}$. 

\begin{equation}\footnotesize
\begin{array}{lr}
\begin{split}
PCD=\mathbf{P(x \prec y)} & =\underbrace{{\text{PoF}}_{\mathbf{x}}(1-{\text{PoF}}_{\mathbf{y}})}_{\mathbf{x}~ \text{feasible,}~\mathbf{y}~\text{infeasible}}+\underbrace{{\text{PoF}}_{\mathbf{x}}{\text{PoF}}_{\mathbf{y}}\mathbf{P}(\mathbf{F(x)} \prec \mathbf{F(y)})}_{\mathbf{x}~ \text{feasible,}~\mathbf{y}~\text{feasible}} \\
& ~+\underbrace{(1-{\text{PoF}}_{\mathbf{x}})(1-{\text{PoF}}_{\mathbf{y}})\mathbf{P}(CV(\mathbf{x}) < CV(\mathbf{y}))}_{\mathbf{x}~ \text{infeasible,}~\mathbf{y}~\text{infeasible}}
\end{split}
\end{array}
\label{eqn:P_AB}
\end{equation}

In the above equations, the probability of feasibility~(PoF) estimates the likelihood that a solution~(e.g. $\mathbf{x}$) falls into the feasible region, computed by integrating the joint prior distributions of all constraints as shown in Eq.~\ref{eqn:pof}.

\begin{equation}\footnotesize
    \text{PoF}_\mathbf{x} = \prod_{i=1}^{p} \Phi(g_i(\mathbf{x}) \leq 0) = \prod_{i=1}^{p} \Phi\left(-\frac{{\mu_g}_i(\mathbf{x})}{{\sigma_g}_i(\mathbf{x})}\right)
    \label{eqn:pof}
\end{equation}

To estimate the dominance as per the second component of Eq.~\ref{eqn:P_AB}, the distributions of the candidates along each objective~(${\mu_f}_i(\mathbf{x})$, ${\sigma_f}_i(\mathbf{x})$) are projected along the corresponding $\hat{\mathbf{W}_j}$. The projection of the distributions on the RV, using Eq.~\ref{eqn:projection}, yields single $\mu_f(\mathbf{x})$ and $\sigma_f(\mathbf{x})$ for a solution, $\mathbf{x}$. 

\begin{equation}\footnotesize
\begin{array}{lr}
\mu_f(\mathbf{x}) = \hat{\mathbf{W}_j}\boldsymbol\mu^\intercal;\hspace{2mm} \sigma_f(\mathbf{x}) = \hat{\mathbf{W}_j}\Psi{\hat{\mathbf{W}_j}}^\intercal 
\end{array}
\label{eqn:projection}
\end{equation}

Here, $\Psi$ represents the covariance matrix $\begin{pmatrix}\sigma_1 & 0\\0 & \sigma_2\end{pmatrix}$ and $\mu_f(\mathbf{x})$, $\sigma_f(\mathbf{x})$ are the mean and standard deviation, respectively, projected along the RV. The probability that $\mathbf{x}$ dominates $\mathbf{y}$ in terms of the $M$ original objectives is estimated by applying the concept of probabilistic dominance~($\text{PD}_{\text{F}}$)~\cite{hughes2001evolutionary}, as shown in Eq.~\ref{eqn:PD}.

\begin{equation}\footnotesize
\mathbf{P\left(F(x) \prec F(y)\right)} = \prod_{i=1}^{M}\left(\frac{1}{2} + \frac{1}{2} erf \left(\dfrac{{\mu_f}_i(\mathbf{y})-{\mu_f}_i(\mathbf{x})}{\sqrt{2\left({{\sigma_f}_i(\mathbf{x})}^2 + {{\sigma_f}_i(\mathbf{y})}^2\right)}}\right)\right)
\label{eqn:PD}
\end{equation}
where, error function 'erf' is defined as: 
\begin{equation}\footnotesize
erf(\mathbf{x}) = \frac{2}{\sqrt \pi}\int_{0}^\mathbf{x} e^{t^{2}}dt
\end{equation}

For the third component, the probability that $CV(\mathbf{x})<CV(\mathbf{y})$ is needed to be estimated. However, from the individual Kriging models, we only get the individual constraint distributions of each solution. Computing $P(CV(\mathbf{x})<CV(\mathbf{y}))$ directly based on these distributions is not appropriate since a highly feasible distribution in the predicted landscape can bias the estimation, therefore resulting in incorrect classification. To overcome this, The concept of rectified Gaussian distribution has been adopted to determine the corresponding CV distributions. In probability theory, it represents a modification of the Gaussian distribution wherein its negative elements are reset to 0, analogous to an electronic rectifier.  Essentially, it is a combination of a discrete distribution (constant 0) and a continuous distribution (a truncated Gaussian distribution with the interval $(0, \infty)$ due to censoring~\cite{sun2023modified}). 

For a normal distribution $\mathcal{N}(\mu,\sigma^2)$, the mean of the rectified distribution~($\mu_R$) yields a higher value than $\mu$ since some probability density is shifted to a higher value~(since negative values are transformed to 0). Conversely, the rectified standard deviation~($\sigma_R$) is smaller than $\sigma$. Following the method proposed in~\cite{palmer2017methods}, the interval between $a = 0$ and $b = \mu + 6\sigma$~(to allow $6\sigma$ limit) is set first to be acting on a standard normal distribution: $c = \frac{a-\mu}{\sigma}$, $d = \frac{b-\mu}{\sigma}$. Using the transformed constraints, $\mu_R$ and $\sigma^2_R$ can be expressed as shown in Eq.~\ref{eqn:RecGauss}.

\begin{equation}\footnotesize
\begin{split}
    \mu_t & = \frac{1}{\sqrt{2\pi}}\left(e^{-\frac{c^2}{2}}-e^{-\frac{d^2}{2}}\right)+\frac{c}{2}\left(1+erf\left(\frac{c}{\sqrt{2}}\right)\right) \\ 
    & +\frac{d}{2}\left(1-erf\left(\frac{d}{\sqrt{2}}\right)\right), \\
    \sigma^2_t & = \frac{\mu^2_t+1}{2}\left(erf\left(\frac{d}{\sqrt{2}}\right)-erf\left(\frac{c}{\sqrt{2}}\right)\right) \\
    & - \frac{1}{\sqrt{2\pi}}\left((d-2\mu_t)e^{-\frac{d^2}{2}}-(c-2\mu_t)e^{-\frac{c^2}{2}}\right) \\
    & + \frac{(c-\mu_t)^2}{2}\left(1+erf\left(\frac{c}{\sqrt{2}}\right)\right)+\frac{(d-\mu_t)^2}{2}\left(1-erf\left(\frac{d}{\sqrt{2}}\right)\right), \\
    \mu_R & = \mu + \sigma\mu_t, \sigma^2_R = \sigma^2\sigma^2_t\\
\end{split}
\label{eqn:RecGauss}
\end{equation}

Once the CV distribution of each individual constraint is available, and assuming that the individual CV distributions are mutually independent, the estimated total~(sum of) CV of a solution, $\mathbf{x}$ can be mathematically expressed as: $\mathcal{N}\left(\sum_{i=1}^{p}{\mu_g}_i, \sum_{i=1}^{p}{\sigma_g}_i^2\right)$. 

\begin{figure}[!ht]
\centering    
\includegraphics[width=0.35\textwidth]{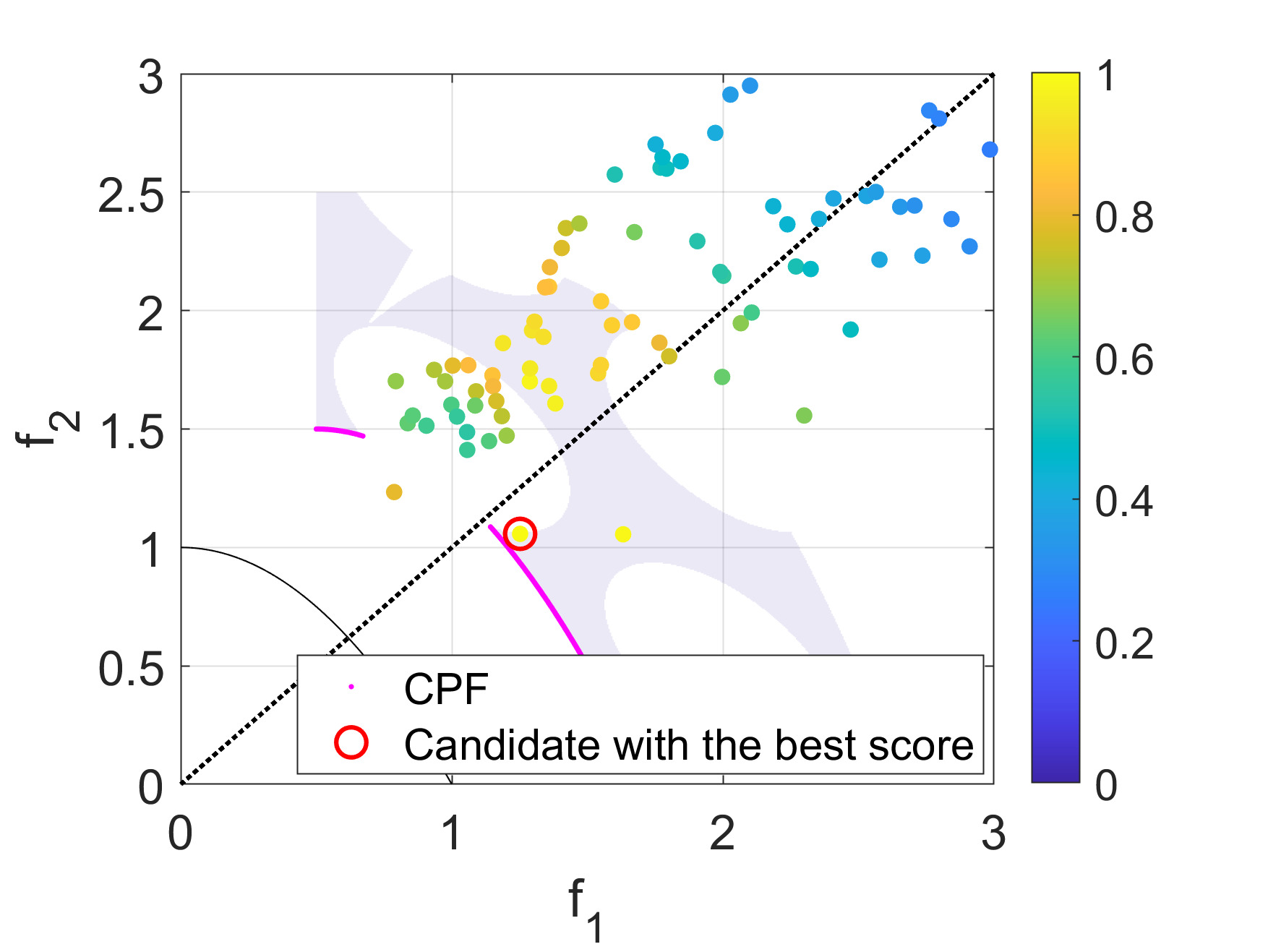}
\caption{Best candidate selection along the RV according to Eq.~\ref{eqn:P_AB}.}
\label{fig:ranking}
\end{figure}

For a given solution $\mathbf{x}$, the above pairwise comparison is done with all solutions associated with the same RV, and the corresponding $PCD$ values are averaged to yield its overall fitness score. This is repeated for each solution along the given RV, and solution with the highest score is selected as the best performing solution along the RV. A proof-of-concept of ranking along a given RV is shown in Fig.~\ref{fig:ranking} for DASCMOP1 problem~\cite{fan2020difficulty} based on 100 samples. The samples span both feasible and infeasible regions of the search space. The colormap of the solutions denotes the score~(higher the better) in a normalized scale. It can be seen that the selected candidate is competitive, as it is feasible and dominates all other feasible solutions. The scores of all other candidates also reflect their quality well in terms of convergence.

The above procedure is repeated for all active RVs. Subsequently, all the active RVs and their corresponding best solutions are selected from $\mathbf{W}$ and $\mathbf{Q}$, respectively. Using the remaining set of RVs and solutions, the process (involving assignment, projection, and selection) is repeated, until no empty RV remains. Thus, in the end, each RV is paired with an associated best solution~(totaling $N_W$ solutions). In the scenario where $N_S > N_W$, the remaining $N_S - N_W$ solutions are randomly chosen to form the surviving set of $N_S$ solutions for the subsequent generation.

\subsection{Infill sample identification}
\label{subsec:infill}

In this stage one solution out of the population~($\mathbf{C}$) delivered by the SubEA needs to be identified for expensive evaluation. To maintain dynamic balance among feasibility, convergence and diversity, the sampling criteria is set based on the status of archive $\mathbf{A}$: (a) $\mathbf{A}$ is fully infeasible -- in this case feasibility and diversity are targeted; (b) $\mathbf{A}$ is fully infeasible -- convergence and diversity are targeted; and (c) $\mathbf{A}$ is a mix of feasible and infeasible solutions -- feasibility is targeted, followed by convergence and diversity. The proposed sampling process is outlined in Algo.~\ref{algo:infill}, noting that any duplicate solutions in $\mathbf{C}$ are removed a priori. Each of the above three cases are discussed in more detail below.

\begin{algorithm}[!ht] \footnotesize
\caption{Infill identification for expensive evaluation}
\begin{flushleft}
\algorithmicrequire \hspace{1mm}Archive~$\mathbf{A}$, candidates~$\mathbf{C}$, $\mathbf{Z^I}$, $\mathbf{Z^N}$, SearchFlag, RVTagFlag. \\
\textbf{Output:} Infill sample, $\mathbf{C}^{In}$ for expensive evaluation.
\end{flushleft}
\begin{algorithmic}[1]
\STATE Identify feasible~($\mathbf{A}_{\text{feas}}$) and infeasible~($\mathbf{A}_{\text{infeas}}$) solutions in $\mathbf{A}$.
\IF{$\mathbf{A}_{\text{feas}} == \emptyset$} \label{line:infill1}
\FOR{$i = 1:|\mathbf{C}|$}
\FOR{$j = 1:|\mathbf{C}|$}
\STATE Compute $\mathbf{P}(\mathbf{C}_i < \mathbf{C}_j)$. \COMMENT{According to PCD~(Eq.~\ref{eqn:P_AB}) if SearchFlag is `Constrained'; otherwise based on ${\text{PD}}_{\text{F}}$~(Eq.~\ref{eqn:PD})}
\ENDFOR
\STATE $\text{Score}_i \longleftarrow \text{mean}\{\mathbf{P}(\mathbf{C}_i < \mathbf{C}_{j=1:|\mathbf{C}|})\}$. 
\ENDFOR
\STATE SC $\longleftarrow \{\text{Score}_1, \ldots, \text{Score}_{\mathbf{C}}\}$. \COMMENT{Final score list of all $\mathbf{C}$}
\STATE Arrange $\mathbf{C}$ in descending order of SC and their corresponding RVs are tagged accordingly. \label{line:infill2}
\IF{RVTagFLag == `ON'}\label{line:infill3}
\STATE Select the top scored candidate along an unique RV as the new infill, $\mathbf{C}^{In}$.
\STATE Store the selected RV for subsequent stage.
\ENDIF
\STATE Select the top scored candidate as the new infill, $\mathbf{C}^{In}$ and store the corresponding RV for subsequent stage.
\STATE Return RVTagFLag = `ON'. \label{line:infill4}
\ELSIF{$\mathbf{A}_{\text{infeas}} == \emptyset$} \label{line:infill5}
\STATE Identify the reference set of solutions, $\mathbf{A}_{\text{ref}}$ where $\mathbf{A}_{\text{ref}}$ = $\mathbf{A}_{\text{ND}}$.
\STATE $\mathbf{C}^{In} \longleftarrow \text{Infill}(\mathbf{A}, \mathbf{C}, \mathbf{Z^I}, \mathbf{Z^N}, \mathbf{A}_{\text{ref}})$. \cite{rahi2022steady} \label{line:infill6}
\ELSE
\STATE \textbf{Identify} ${\mathbf{Fnd}}_{\text{feas}}$ from $\mathbf{A}_{\text{feas}}$. \label{line:infill7}
\STATE \textbf{Identify} and \textbf{list} the infeasible solution(s) which are ND w.r.t any ${\mathbf{Fnd}}_{\text{feas}}$, presented as ${\mathbf{F}}^L_{\text{infeas}}$.
\STATE \textbf{Combine} $\mathbf{F}_{\text{comb}} \longleftarrow \{{\mathbf{Fnd}}_{\text{feas}}, {\mathbf{F}}^L_{\text{infeas}}\}$ and treat them as $\mathbf{A}_{\text{ref}}$. \label{line:infill10}
\STATE $\mathbf{C}^{In} \longleftarrow \text{Infill}(\mathbf{A}, \mathbf{C}, \mathbf{Z^I}, \mathbf{Z^N}, \mathbf{A}_{\text{ref}})$. \cite{rahi2022steady} \label{line:infill9}
\ENDIF \label{line:infill8}
\end{algorithmic}
\label{algo:infill}
\end{algorithm}

\begin{figure}[!ht]
\centering
\subfigure[Iteration ($t$)]{\label{fig:infills1}\includegraphics[width=0.23\textwidth]{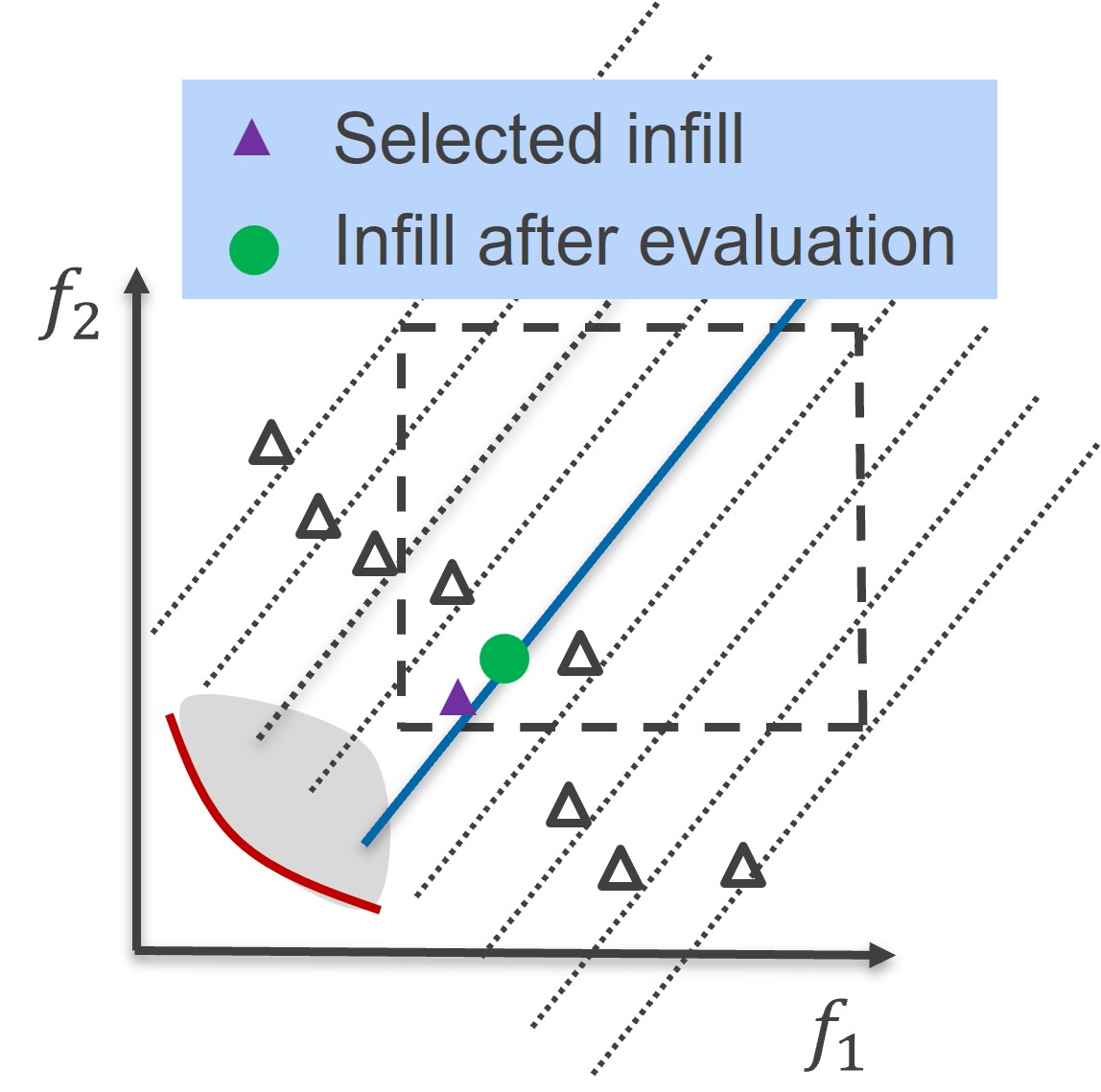}} \quad
\subfigure[Iteration ($t+1$)]{\label{fig:infills2}\includegraphics[width=0.23\textwidth]{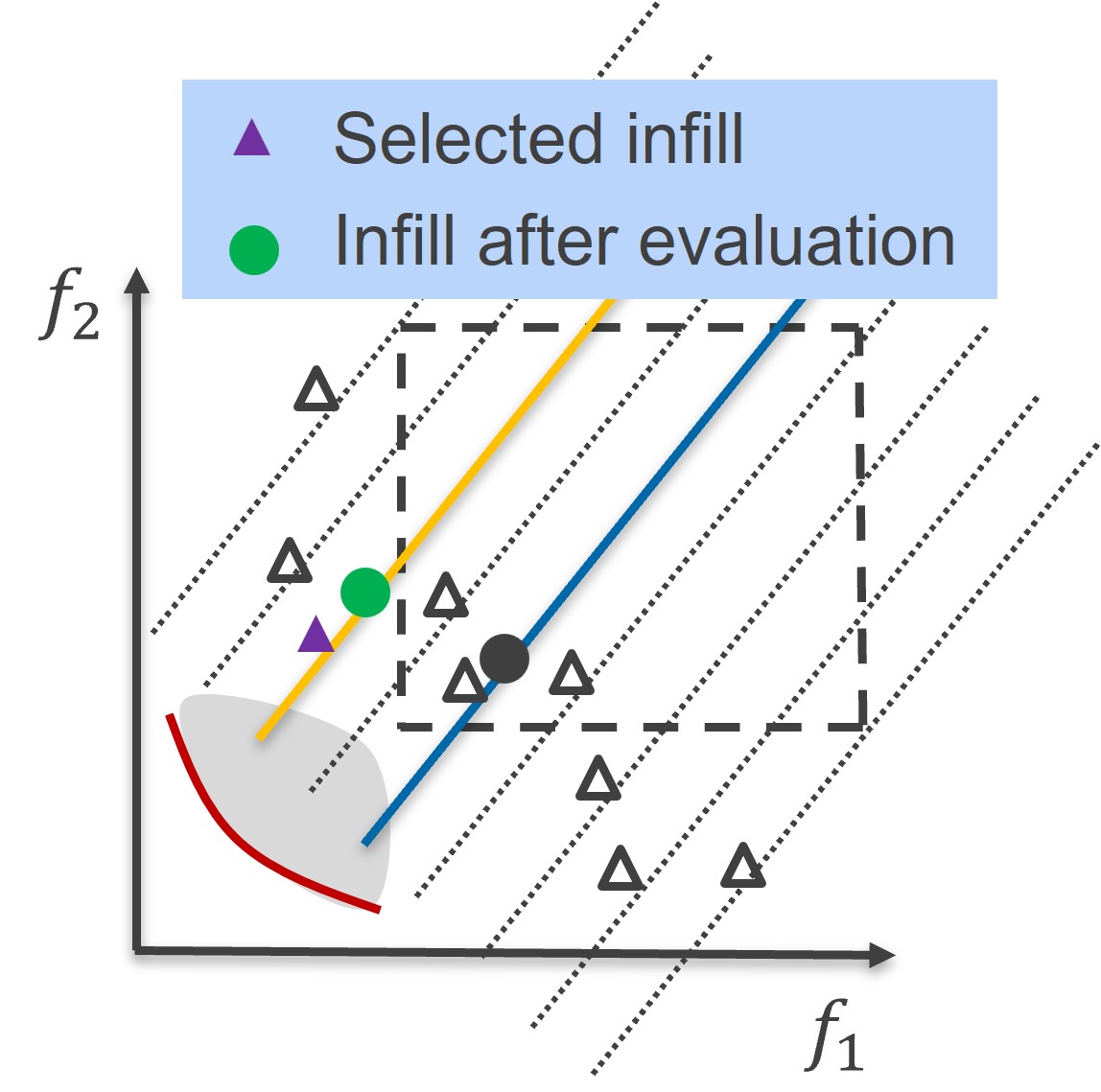}} 
\caption{Infill selection when the full evaluated archive~$\mathbf{A}$ is infeasible.}
\label{fig:infills}
\end{figure}

Until no feasible solution has been identified in $\mathbf{A}$, the preferred infill solution is the one that is most likely of being feasible along a new RV, to enhance feasibility and diversity. For this, the average score of each solution in $\mathbf{C}$ is computed by comparing it with all other solutions in the set using Eq.~\ref{eqn:P_AB}, and the one with highest score is chosen for evaluation. The selection process is illustrated in Fig.~\ref{fig:infills}. The triangular markers represent candidate set $\mathbf{C}$, along the RVs in the predicted landscape. The one with highest score is highlighted in magenta and selected for true evaluation at iteration $t$ as shown in Fig.~\ref{fig:infills1}. The green circular marker shows the solution in the original objective space after evaluation.  The corresponding RV~(blue solid line) is also stored in $\mathbf{A}$, to promote diversity in the subsequent selection process. Fig.~\ref{fig:infills2} represents the selection stage of iteration~($t+1$), where the search bound is still unchanged due to not achieving improved performance from iteration~($t$). The selection of blue RV again~(same RV as the previous iteration) indicates that the corresponding solution based on the prediction still represents the best performing one. In such a scenario, the selection strategy avoids the blue RV sample and instead selects the next best performing candidate~(along yellow RV). The same process is repeated as long as the search bound in the original objective space stays unchanged and no feasible solution is identified in $\mathbf{A}$. This is maintained by the `RVTagFlag' as mentioned in Algo.~\ref{algo:infill}, line~\ref{line:infill3} - \ref{line:infill4}. This simple diversity enhancing approach encourages search for feasible regions from different directions resulting in enhanced performance where the problem has multiple disconnected feasible regions.  

On the other hand, when some or all solutions in $\mathbf{A}$ are feasible, the intent is to promote convergence and diversity during infill sample selection aiming for well distributed solutions on the CPF, outlined in Algo.~\ref{algo:infill}, line~\ref{line:infill5} - \ref{line:infill8}. This is done through an adapted version of two-state infill identification, which as demonstrated in~\cite{rahi2022steady} for unconstrained problems. In the first stage, the subset of $\mathbf{C}$ that is ND w.r.t. a reference set $\mathbf{A}_{\text{ref}}$ is identified. In case no such solution exists in $\mathbf{C}$, the ND set of $\mathbf{C}$ is chosen instead. In the second stage, a distance-based selection~(DSS) approach~\cite{singh2018distance} is applied on the subset to select the most diverse candidate as the final infill solution. The key element here is the use of Mahalanobis distance~(MD)~\cite{rahi2022steady} for computing distance instead of commonly used Euclidean distance~(ED). This is because MD considers both the mean and uncertainties in the computation of distances~(in this instance, characterized by predicted mean and uncertainty of the Kriging model). 

The notable difference in this study from \cite{rahi2022steady} is the construction of $\mathbf{A}_{\text{ref}}$. In \cite{rahi2022steady}, $\mathbf{A}_{\text{ref}}$ is simply the ND set of archived solutions~($\mathbf{A}_{\text{ND}}$), since there are no infeasible solutions to deal with in unconstrained problems. However, here, when the archive includes a mix of feasible and infeasible solutions, $\mathbf{A}_{\text{ref}}$ is constructed by assembling feasible ND solution(s)~($\mathbf{Fnd}_{\text{feas}}$) with the set of infeasible solution(s) which are ND w.r.t any $\mathbf{Fnd}_{\text{feas}}$ in terms of the original objectives~(Algo.~\ref{algo:infill}, line~\ref{line:infill7} - \ref{line:infill10}). This is consistent with the setting of normalization bound as discussed in Section~\ref{subsec:normalbound}. The general idea behind preserving the infeasible solution(s) with good objective performance in $\mathbf{A}_{\text{ref}}$ is to explore diverse areas for sampling if any feasible region still remains undiscovered. This will particularly benefit the search in problems with multiple disconnected feasible regions. 

Finally, the concept of `shadow ND archive'~($\mathbf{A^S}$) is also adopted from~\cite{rahi2022steady} with slight modification. It contains the set of infill solutions that ended up being dominated after evaluation, or whose closest solution in $\mathbf{A}$ is dominated. This set is included into $\mathbf{A}_{\text{ref}}$ while applying DSS to reduce the possibility of selecting an infill in a dominated region. Unlike \cite{rahi2022steady}, in the presence of constraints, apart from the newly evaluated solution being dominated, there is a chance of it being infeasible as well. When such a solution is encountered, it is also assigned to $\mathbf{A^S}$. Note that $\mathbf{A^S}$ is only activated and updated progressively from the point of identifying a feasible solution in $\mathbf{A}$. If feasible solution(s) exists in the initial archive, $\mathbf{A^S}$ is constructed at the beginning as outlined in Algo.~\ref{algo:general}, line~\ref{line:gen1} - \ref{line:gen2}. Initially, the infeasible solutions which are ND to any feasible solution, are added to the $\mathbf{A^S}$. 

After the MD computations based on the above sets, the candidate solution in the subset of $\mathbf{C}$ with the highest merit is selected as the final infill solution, employing the same principle as in DSS. The solution then undergoes true~(expensive) evaluation.

\subsection{Search switching mechanism}
\label{subsec:switch}

To expedite feasible region identification and convergence, an adaptive search switching mechanism is additionally embedded within PSCMOEA framework which is only triggered based on certain conditions. The `SearchFlag' essentially controls the environmental selection step in SubEA~(as discussed in Section~\ref{subsubsec:selection}) and infill selection~(when $\mathbf{A}$ is fully infeasible) by setting `Unconstrained' or `Constrained' based probabilistic ranking criteria. Note that `Constrained' flag is the default whereas the `Unconstrained' flag is only activated based on two specific conditions, which are (a) no feasible solution is identified so far in the $\mathbf{A}$ and (b) the Kendall-tau rank correlation~\cite{kendall1948rank} between objective optimization and CV minimization is at least moderately positive~(defined by a value $\ge 0.27$ in this study, based on approximate ranges suggested in statistics literature). This general idea is to bias the search towards the UPF for identifying potential feasible location(s), when the UPF and CPF are along the same search direction. This offers the strategy to adapt based on the problem characteristics and is effective in expediting the convergence. For determining the rank correlation, referred to as $\tau$, two ranks are derived for every solution. The first denotes the rank of the solution based on its CV~(in ascending order) and the second rank is based on their corresponding front by front rank calculated by non-dominated sorting approach. The correlation between these two is denoted by $\tau$. Once/if the `Unconstrained' search is triggered, it is continued to the subsequent iteration only if the latest evaluated solution yields the minimum CV value, otherwise it returns back to regular `Constrained' search. The switching mechanism is also terminated once the first feasible solution is identified. 

\section{Numerical Experiments}
\label{sec:exp}

The proposed PSCMOEA is assessed on multiple well-known benchmark problem suites to cover a wide range of problem characteristics. The performance is compared with five state-of-the-art SA-CMOEAs proposed in recent years. Additionally, an ablation study is conducted to highlight the functionality and significance of some the algorithm's key components.

\subsection{Test problems and experimental settings}
\label{subsec:problems}

As detailed in Section~\ref{subsec:motive}, two test cases, namely Test1 and Test2, are formulated in this study to highlight some challenging scenarios, the detailed formulations of which are in the supplementary online material~(SOM Section I). Besides this, numerical experiments are conducted on three distinct set of popular benchmark test problems including MW series~\cite{ma2019evolutionary}~(14 instances), LIRCMOP series~\cite{fan2019improved}~(14 instances) and DASCMOP series~\cite{fan2020difficulty}~(9 instances). While most of the problems involves bi-objective formulations, few of them such as MW4, MW8, MW14, LIRCMOP13 - LIRCMOP14 and DASCMOP7 - DASCMOP9 contain 3 objectives. For this study, variable dimension~($D$) is consistently set to as 10.  
These test suites collectively cover a wide range of challenges, such as low feasibility, disconnected CPFs, local fronts, and irregular CPF shapes. 

For PSCMOEA, the crossover and mutation probability is set to as 0.9 and 0.1, respectively with distribution index as 10 and 20, respectively. Regarding SubEA, a population of 100 solutions undergoes evolution over 100 generations. The spacing parameter ($H$) for generating RVs is set to 99 for 2-objective instances and 12 for 3-objective instances. The five representative state-of-the-art algorithms considered are KMGSAEA~\cite{zhang2023multigranularity}, ASA-MOEA/D~\cite{yang2023surrogate}, KTS~\cite{song2023balancing}, MultiObjectiveEGO~\cite{hussein2016generative}, and SILE~\cite{yin2022fast}. Four of them have been published in last two years~(2022-23) and evaluate more than one solution per generation, while one of them, MultiObjectiveEGO, is a steady-state method that evaluates one solution per generation. The global parameter settings remain consistent across all algorithms. For KMGSAEA, MultiObjectiveEGO and SILE including PSCMOEA, the initial sample size is determined as 11$D$ - 1, where $D$ represents the number of variables. Note that KTS is an exception, with a fixed sample size of 100 for all cases due to algorithmic requirements. Similarly, ASA-MOEA/D uses 50 and 91 initial samples for 2- and 3-objective instances respectively. The initial samples for each independent run are generated using the same seed for all algorithms, ensuring a fair comparison without initialization bias. The experiments are conducted with maximum evaluations $FE_{max}=500$ for all algorithms and all problems. Other algorithm specific parameters are set to their defaults recommended in the respective publications. Results of 31 runs were used for statistical performance comparison.

\vspace{-1mm}
\subsection{Performance measurement}
\label{subsec:perfmetrics}

To measure the quality of the obtained CPF approximations in terms of convergence and diversity, three widely recognized metrics are utilized: IGD~\cite{coello2007evolutionary}, IGD\textsuperscript{+}~\cite{ishibuchi2015modified} and HV~\cite{bader2011hype}. The reference point for HV calculation is set to the commonly used $(1.1,\ldots,1.1)$. For IGD and IGD\textsuperscript{+}, reference CPF sets with 1000 points were generated using PlatEMO for MW, LIRCMOP, and DASCMOP series. For Test1 and Test2 problems, same reference sets are adopted as FCP1 and FCP3 test problems~\cite{yuan2021indicator} as they share identical CPFs. All metrics are computed in the normalized objective space using true ideal and nadir points. The computation is performed on the feasible points obtained by an algorithm in each run for a given problem. If an algorithm in unable to identify any feasible solution in a given run, the following process if followed. The worst metric value in terms of IGD, IGD\textsuperscript{+} or HV is determined across all the considered algorithms for a particular problem, and is deteriorated slightly further by adding for~(IGD, IGD\textsuperscript{+}) or subtracting~(HV) an additional small quantity $0.1$. 
    
To measure how efficiently the algorithm finds feasible solutions, the number of evaluations required for the algorithm to identify the first feasible solution~(FFE) is tracked. In case no feasible solution is identified in a particular run, FFE is assigned as the $FE_{max}=500$. Additionally, to establish the consistency in performance of an algorithm, the number of successful trials~(ST) is counted, where the algorithm successfully identified at least one feasible solution. Notably, among the test problems, some are ``feasibility-hard'' for which it is difficult to locate a feasible solution within a limited budget, such as MW series, LIRCMOP1 - LIRCMOP4 and DASCMOP5 - DASCMOP9. On the other hand, some problems such as LIRCMOP5 - LIRCMOP14 and DASCMOP1 - DASCMOP3, are feasibility-easy, where feasible solutions may often exist in the initial randomly generated samples itself. 

To establish statistical significance, Wilcoxon ranksum test with 95\% confidence interval is utilized. In the pairwise comparisons, a statistically significantly better, worse and equivalent performance in terms of a given metric is classified as `win'~($\uparrow$), `loss'~($\approx$) and `tie'~($\downarrow$) respectively.  

Lastly, to visually summarize the cumulative performance of the algorithms across multiple problems, \emph{Performance profiles}~\cite{dolan2002} based on mean IGD and mean FFE are utilized. In these plots, the x-axis indicates the ratio~($\lambda$) of the performance of a given algorithm compared to the best performing algorithm for the instance~(lower the better). The y-axis~($\rho_s(\lambda)$) denotes the proportion of total instances solved by the algorithm for a given $\lambda$. A higher area under the curve signifies a better cumulative performance.  

\subsection{Performance comparison in terms of FFE and ST}

\begin{figure}[!ht]
\centering
\subfigure[MW11]{\label{fig:perftest0}\includegraphics[width=0.23\textwidth]{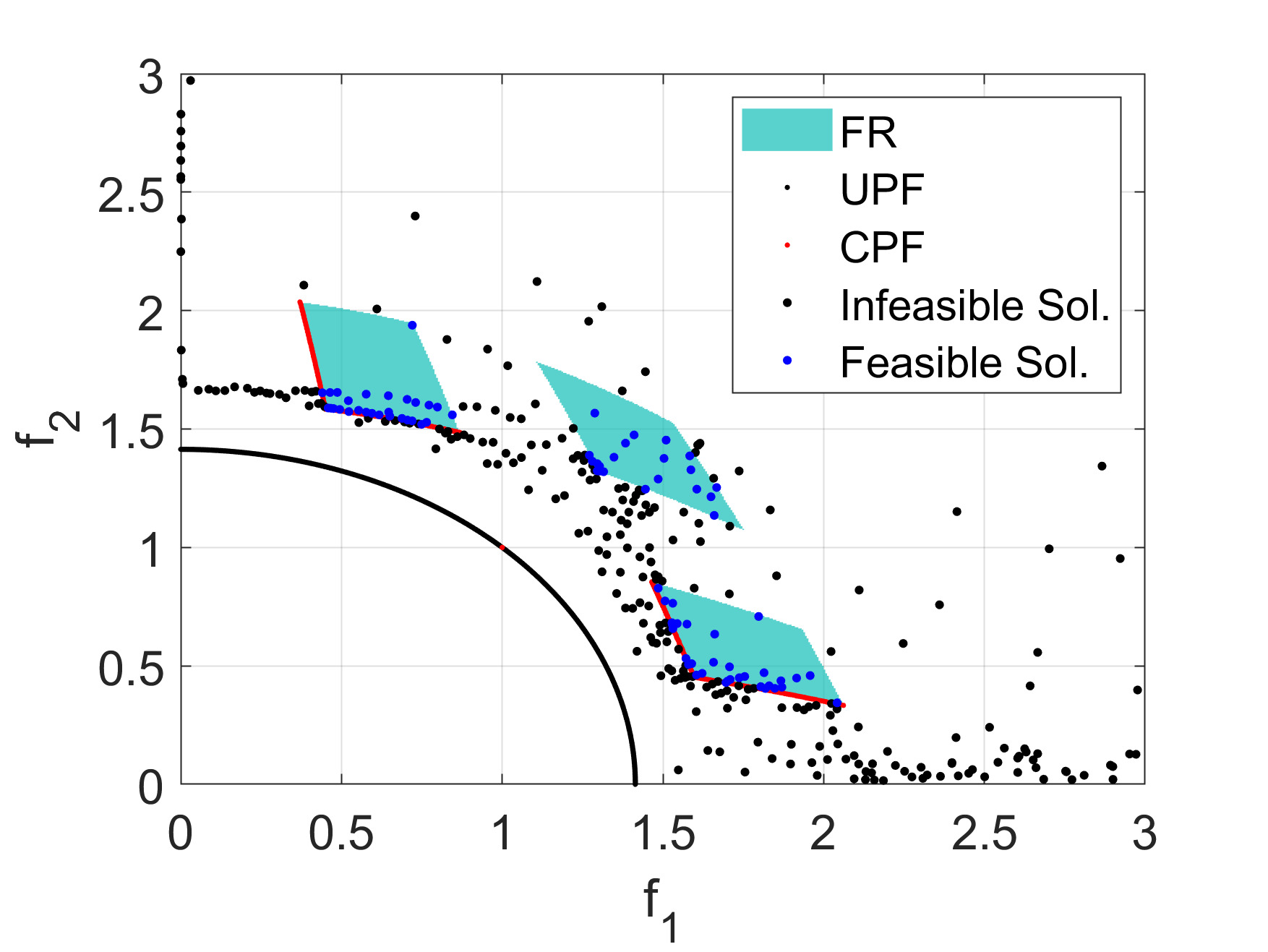}} \quad
\subfigure[Test1]{\label{fig:perftest1}\includegraphics[width=0.23\textwidth]{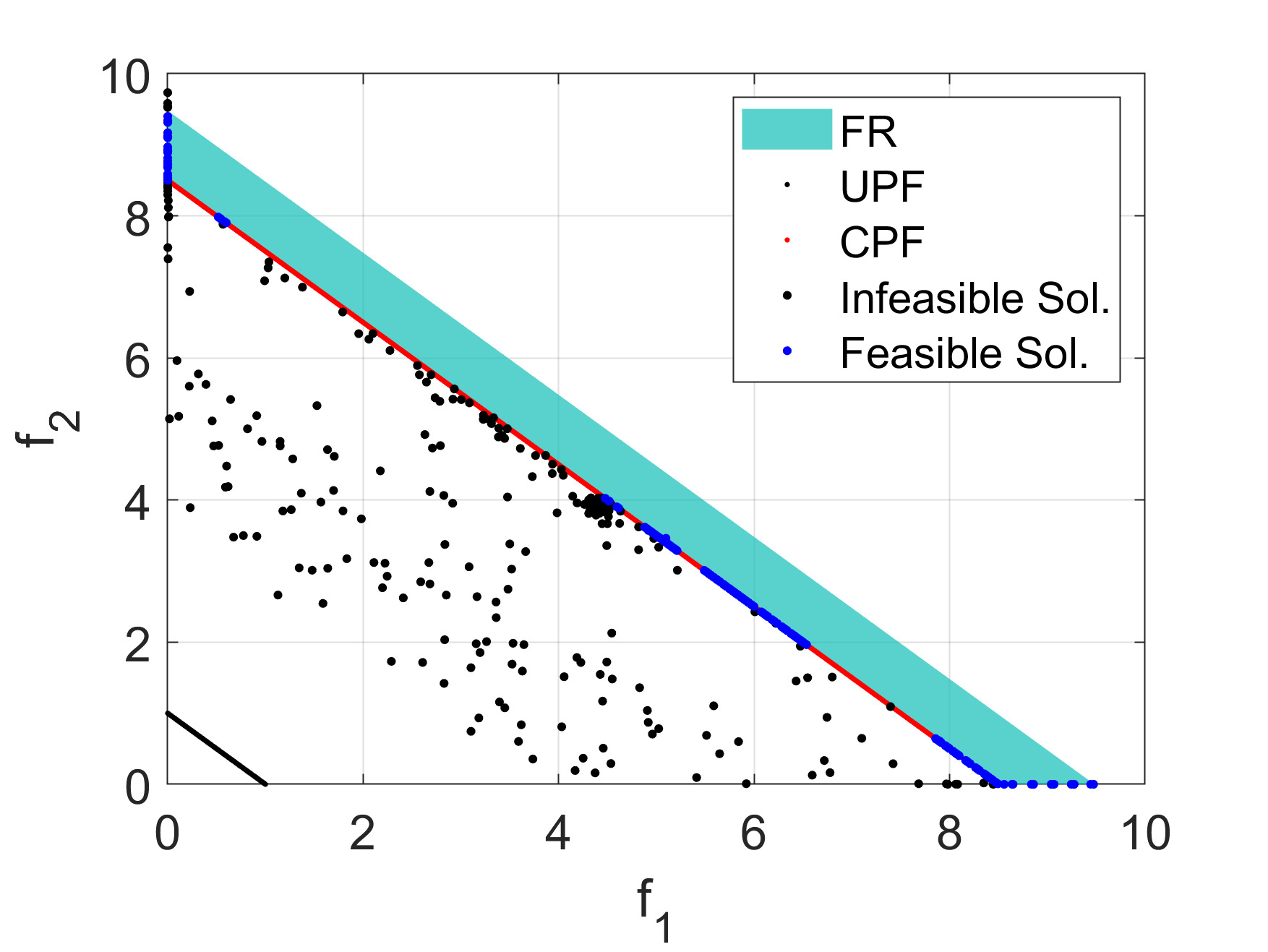}} \\
\subfigure[Test2]{\label{fig:perftest2}\includegraphics[width=0.23\textwidth]{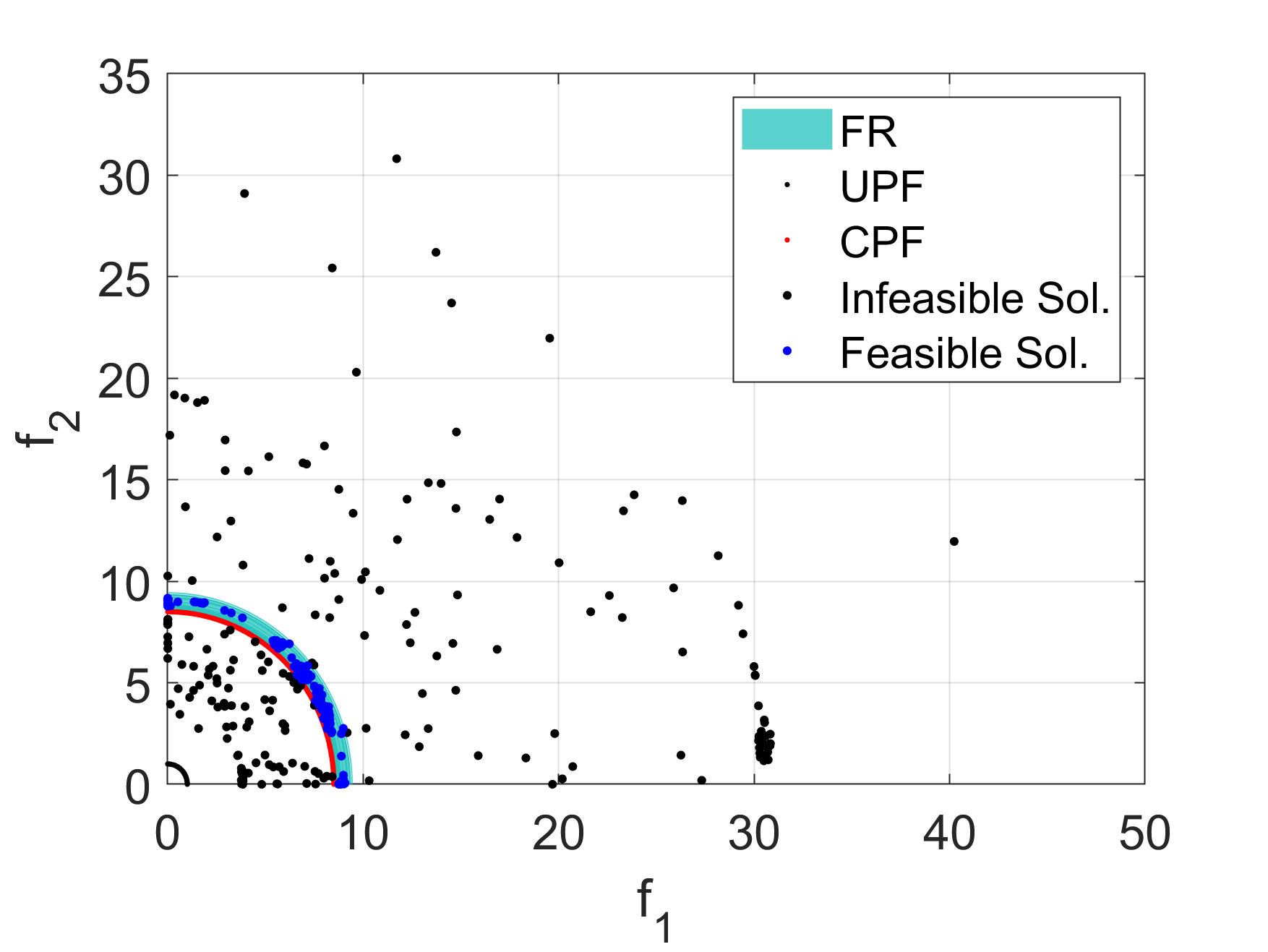}} 
\caption{Distribution of the evaluated solutions for the median run~(out of 31) in terms of IGD by PSCMOEA.}
\label{fig:perftest}
\end{figure}

\begin{figure}[!ht]
\centering
\subfigure[]{\label{fig:perfffe1}\includegraphics[width=0.12\textwidth]{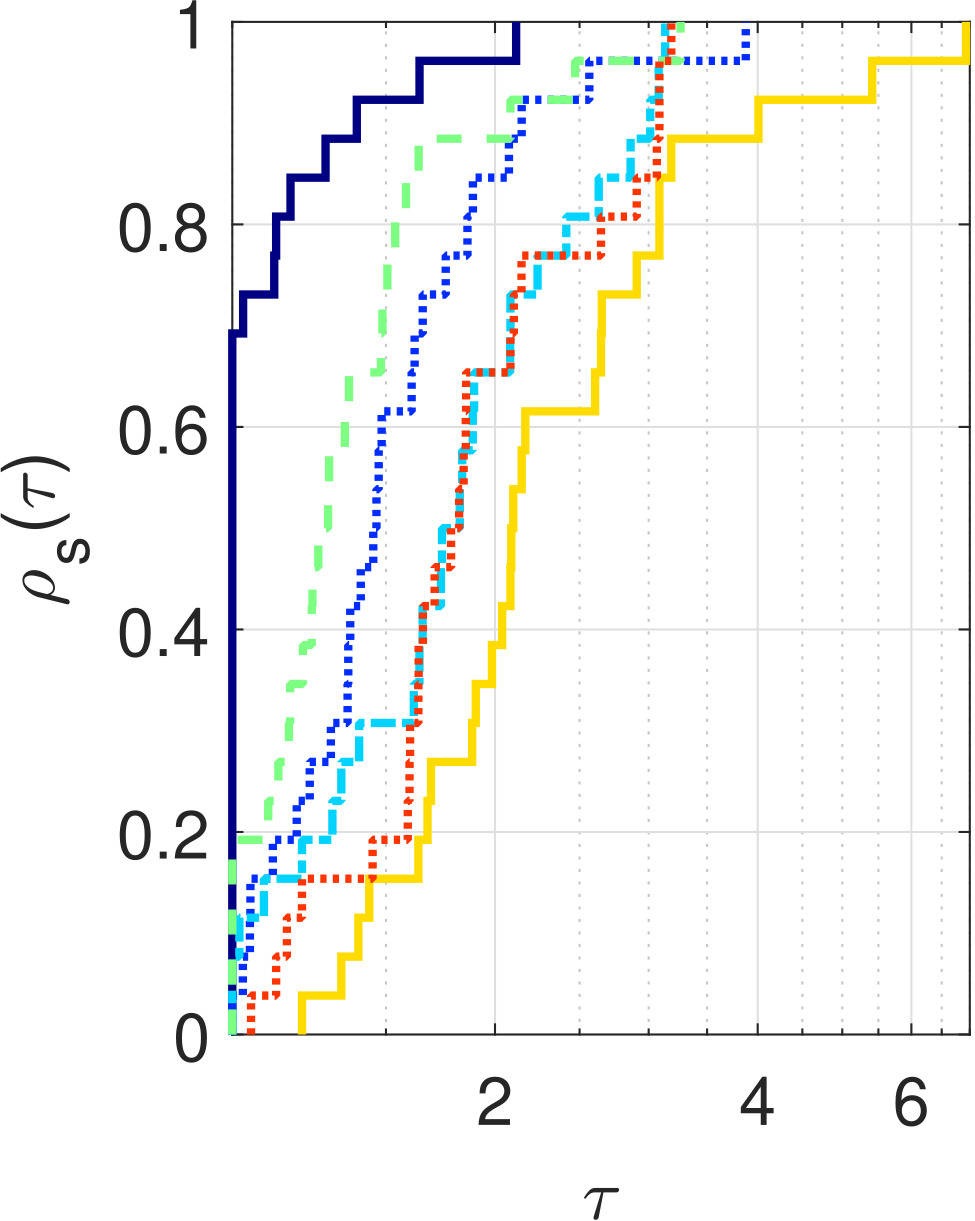}}
\subfigure[]{\label{fig:perfprof1}\includegraphics[width=0.12\textwidth]{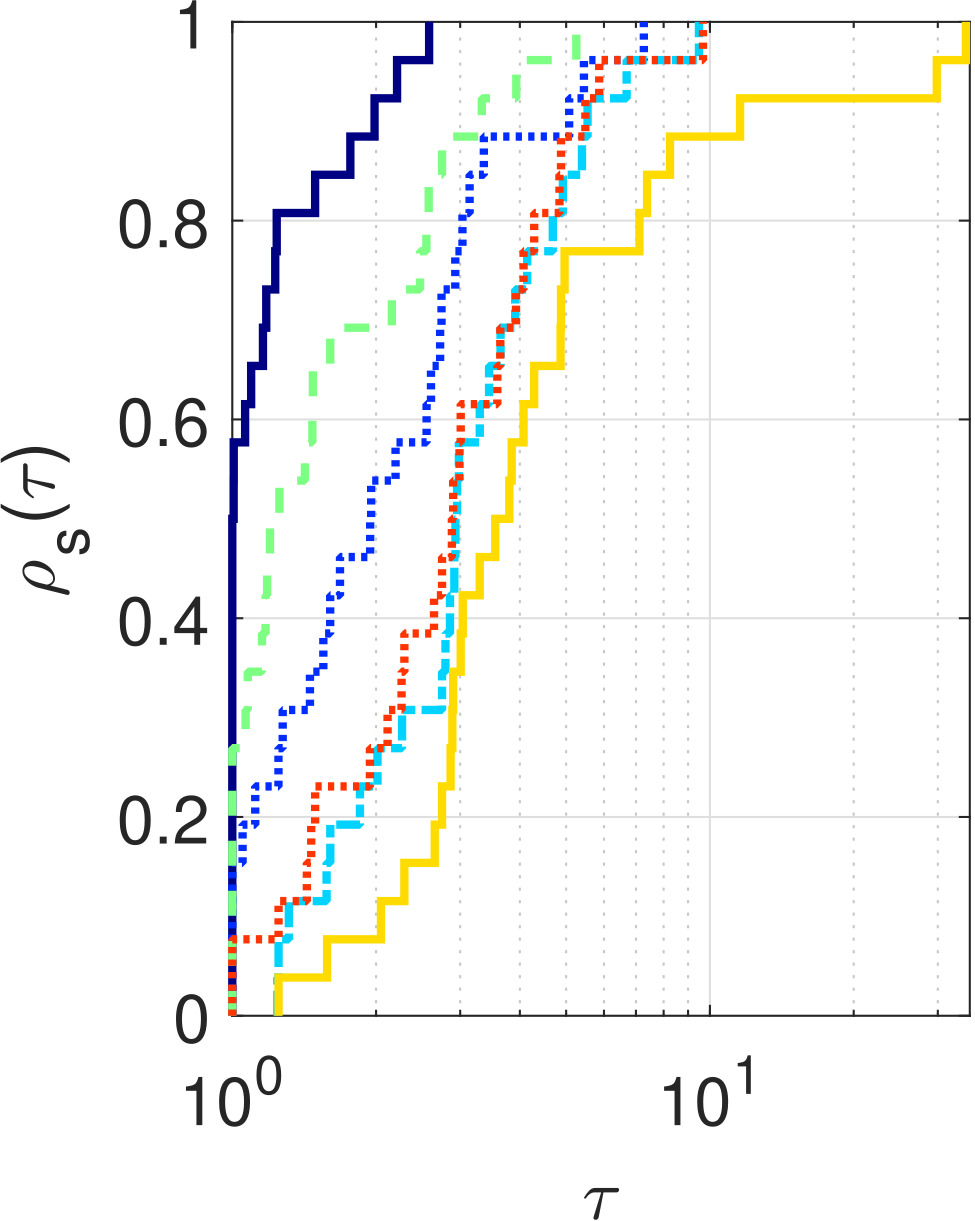}} 
\subfigure[]{\label{fig:perfprof2}\includegraphics[width=0.20\textwidth]{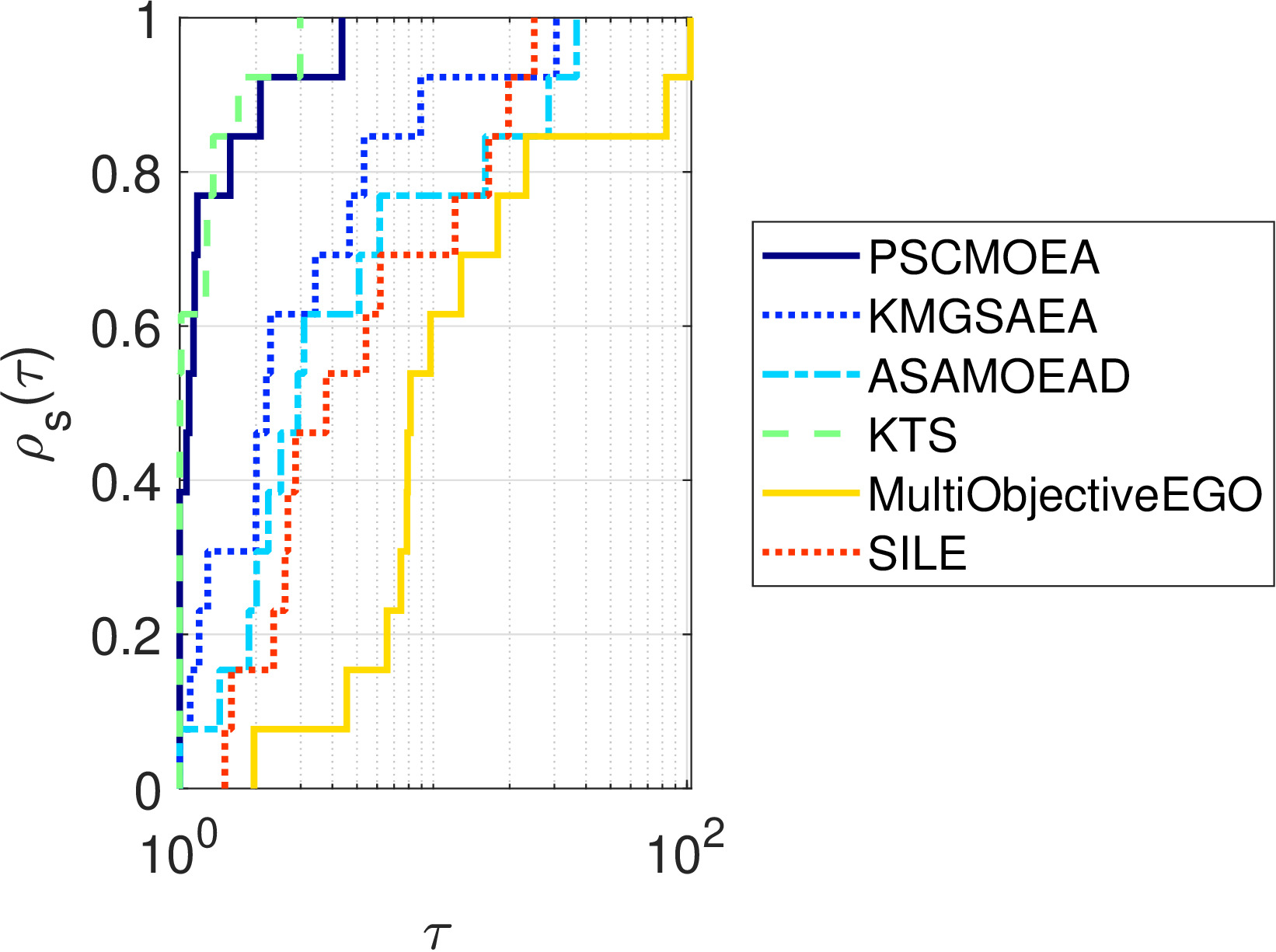}} 
\caption{Performance profiles based on (a) mean FFE and (b) mean IGD for feasibility-hard problems~(Test1-2, MW1-14, LIRCMOP1-4, DASCMOP4-9), (c) mean IGD for feasibility-easy problems~(LIRCMOP5-14 and DASCMOP1-3).}
\label{fig:perfprof}
\end{figure}

The statistics obtained FFE and ST across 31 runs are shown in Table~\ref{tab:FFEST}. From the ST values, it is clearly noticeable that PSCMOEA is overall the most successful and consistent method in identifying feasible regions for the feasibility hard problems such as Test1, Test2, MW1, MW10, LIRCMOP1, LIRCMOP2, DASCMOP9 etc. For DASCMOP9, PSCMOEA identifies feasible solution in all runs, no other peer algorithm could identify any feasible solution in any run. Similar trends can be noticed from Test1, Test2, MW10 problems as well. There are only a few exceptions where PSCMOEA is not the best but is competitive, such as MW8, MW11, DASCMOP5, DASCMOP6~(inferior to KMGSAEA) and LIRCMOP3~(inferior to KTS). The overall performance establishes the credibility of the proposed probabilistic selection strategy to drive the search in the predicted landscape. The validation of correct search behavior can be confirmed from the illustration in Fig.~\ref{fig:perftest} where its clear that PSCMOEA could guide the search towards the CPF for Test1/Test2 problems even though UPF is not located in same search direction as CPF. 
As for the FFE statistics, PSCMOEA once again achieves superior performance in most of the feasibility-hard problems compared to its counterparts. This means that PSCMOEA is able to identify feasible solution swiftly. Thus, fewer function evaluations are spent on the infeasible space, leaving more of the budget towards identifying a well distributed set of CPF solutions. Additionally, the cumulative results of the pairwise significance test ($\uparrow$/$\approx$/$\downarrow$ of compared method vs PSCMOEA) at the bottom of Table~\ref{tab:FFEST} confirms the clear advantage of PSCMOEA over it's peers in terms of FFE. This is also reflected strongly in the performance profile in terms of mean FFE presented in Fig.~\ref{fig:perfffe1} for the feasibility-hard problems. PSCMOEA obtains the best results in close to 70\% of the instances, and competitive performance in remaining problems with low $\tau$ values relative to the peers.

\begin{table*}[!ht]\scriptsize
\centering
\caption{Mean FFE~(std.) and ST across 31 trials obtained by PSCMOEA and peer-algorithms. $\uparrow$/$\approx$/$\downarrow$ symbols denote whether the corresponding algorithm is significantly better/ equivalent/ worse than PSCMOEA. The gray shade highlights the best mean performance.}
\label{tab:FFEST}
\tabcolsep 1.5mm
\renewcommand{\arraystretch}{1.0}
\begin{center}
\vspace{-3mm}
\begin{tabular}{|c|c|c|c|c|c|c|c|c|c|c|c|c|c|}
\hline
\multirow{2}{*}{\textbf{Problems}} & \multirow{2}{*}{\textbf{M}} & \multicolumn{2}{c|}{\textbf{PSCMOEA}} & \multicolumn{2}{c|}{\textbf{KMGSAEA}} & \multicolumn{2}{c|}{\textbf{ASA-MOEA/D}} & \multicolumn{2}{c|}{\textbf{KTS}} & \multicolumn{2}{c|}{\textbf{MultiObjectiveEGO}} & \multicolumn{2}{c|}{\textbf{SILE}} \\ \cline{3-14} 
&    & \multicolumn{1}{c|}{\textbf{FFE}} & \textbf{ST} & \multicolumn{1}{c|}{\textbf{FFE}} & \textbf{ST} & \multicolumn{1}{c|}{\textbf{FFE}} & \textbf{ST} & \multicolumn{1}{c|}{\textbf{FFE}} & \textbf{ST} & \multicolumn{1}{c|}{\textbf{FFE}} & \textbf{ST} & \multicolumn{1}{c|}{\textbf{FFE}} & \textbf{ST}  \\ \hline

\multirow{1}{*}{\textbf{Test1}} & \multirow{1}{*}{2} & \cellcolor{gray!25}129 (50) & \cellcolor{gray!25}31 & 500 (0) $\downarrow$ & 0 & 168 (183) $\uparrow$ & 24 & 421 (149) $\downarrow$ & 7 & 336 (147) $\downarrow$ & 25 & 220 (146) $\downarrow$ & 25 \\
\hline

\multirow{1}{*}{\textbf{Test2}} & \multirow{1}{*}{2} & \cellcolor{gray!25}138 (116) & \cellcolor{gray!25}29 & 354 (216) $\downarrow$ & 10 & 150 (148) $\approx$ & 28 & 341 (218) $\downarrow$ & 11 & 299 (210) $\downarrow$ & 17 & 296 (209) $\downarrow$ & 16 \\
\hline

\multirow{1}{*}{\textbf{MW1}} & \multirow{1}{*}{2}  & \cellcolor{gray!25}296 (123) & \cellcolor{gray!25}27 & 433 (74) $\downarrow$ & 20 & 485 (43) $\downarrow$ & 5 & 484 (47) $\downarrow$ & 6 & 500 (0) $\downarrow$ & 0 & 484 (43) $\downarrow$ & 5 \\
\hline

\multirow{1}{*}{\textbf{MW2}} & \multirow{1}{*}{2}  & \cellcolor{gray!25}147 (28) & \cellcolor{gray!25}31 & 218 (68) $\downarrow$ & 30 & 196 (86) $\downarrow$ & \cellcolor{gray!25}31 & 166 (69) $\approx$ & \cellcolor{gray!25}31 & 211 (50) $\downarrow$ & \cellcolor{gray!25}31 & 235 (60) $\downarrow$ & \cellcolor{gray!25}31 \\
\hline

\multirow{1}{*}{\textbf{MW3}} & \multirow{1}{*}{2}  & 118 (8) & \cellcolor{gray!25}31 & 119 (4) $\approx$ & \cellcolor{gray!25}31 & \cellcolor{gray!25}72 (22) $\uparrow$ & \cellcolor{gray!25}31 & 107 (11) $\uparrow$ & \cellcolor{gray!25}31 & 499 (3) $\downarrow$ & 1 & 131 (17) $\downarrow$ & \cellcolor{gray!25}31 \\
\hline

\multirow{1}{*}{\textbf{MW4}} & \multirow{1}{*}{3}  & \cellcolor{gray!25}263 (81) & \cellcolor{gray!25}30 & 387 (101) $\downarrow$ & 18 & 496 (12) $\downarrow$ & 4 & 396 (88) $\downarrow$ & 22 & 500 (0) $\downarrow$ & 0 & 487 (43) $\downarrow$ & 3 \\
\hline

\multirow{1}{*}{\textbf{MW5}} & \multirow{1}{*}{2}  & \cellcolor{gray!25}252 (68) & \cellcolor{gray!25}31 & 344 (66) $\downarrow$ & 29 & 437 (90) $\downarrow$ & 12 & 373 (118) $\downarrow$ & 20 & 500 (0) $\downarrow$ & 0 & 467 (76) $\downarrow$ & 6 \\
\hline

\multirow{1}{*}{\textbf{MW6}} & \multirow{1}{*}{2}  & \cellcolor{gray!25}157 (29) & \cellcolor{gray!25}31 & 292 (118) $\downarrow$ & 28 & 379 (125) $\downarrow$ & 19 & 194 (60) $\downarrow$ & \cellcolor{gray!25}31 & 320 (96) $\downarrow$ & 28 & 330 (100) $\downarrow$ & 27 \\
\hline

\multirow{1}{*}{\textbf{MW7}} & \multirow{1}{*}{2}  & 125 (9) & \cellcolor{gray!25}31 & 122 (11) $\approx$ & \cellcolor{gray!25}31 & \cellcolor{gray!25}90 (16) $\uparrow$ & \cellcolor{gray!25}31 & 111 (8) $\uparrow$ & \cellcolor{gray!25}31 & 487 (73) $\downarrow$ & 1 & 143 (18) $\downarrow$ & \cellcolor{gray!25}31 \\
\hline

\multirow{1}{*}{\textbf{MW8}} & \multirow{1}{*}{3}  & 229 (106) & 30 & 243 (48) $\approx$ & \cellcolor{gray!25}31 & 376 (115) $\downarrow$ & 22 & \cellcolor{gray!25}205 (92) $\approx$ & 29 & 428 (115) $\downarrow$ & 12 & 339 (118) $\downarrow$ & 23 \\
\hline

\multirow{1}{*}{\textbf{MW9}} & \multirow{1}{*}{2}  & \cellcolor{gray!25}306 (76) & \cellcolor{gray!25}31 & 321 (73) $\approx$ & 30 & 494 (30) $\downarrow$ & 2 & 394 (92) $\downarrow$ & 23 & 500 (0) $\downarrow$ & 0 & 500 (0) $\downarrow$ & 0 \\
\hline

\multirow{1}{*}{\textbf{MW10}} & \multirow{1}{*}{2}  & \cellcolor{gray!25}416 (118) & \cellcolor{gray!25}14 & 463 (70) $\approx$ & 11 & 500 (0) $\downarrow$ & 0 & 483 (48) $\downarrow$ & 4 & 500 (0) $\downarrow$ & 0 & 500 (0) $\downarrow$ & 0 \\
\hline

\multirow{1}{*}{\textbf{MW11}} & \multirow{1}{*}{2}  & 258 (91) & 30 & \cellcolor{gray!25}122 (15) $\uparrow$ & \cellcolor{gray!25}31 & 273 (85) $\approx$ & \cellcolor{gray!25}31 & 147 (56) $\uparrow$ & \cellcolor{gray!25}31 & 489 (57) $\downarrow$ & 2 & 374 (149) $\downarrow$ & 17 \\
\hline

\multirow{1}{*}{\textbf{MW12}} & \multirow{1}{*}{2}  & \cellcolor{gray!25}233 (71) & \cellcolor{gray!25}31 & 374 (52) $\downarrow$ & 29 & 441 (105) $\downarrow$ & 10 & 317 (98) $\downarrow$ & 29 & 500 (0) $\downarrow$ & 0 & 429 (84) $\downarrow$ & 19 \\
\hline

\multirow{1}{*}{\textbf{MW13}} & \multirow{1}{*}{2}  & 119 (50) & \cellcolor{gray!25}31 & 109 (37) $\approx$ & \cellcolor{gray!25}31 & 108 (55) $\approx$ & \cellcolor{gray!25}31 & \cellcolor{gray!25}106 (38) $\approx$ & \cellcolor{gray!25}31 & 281 (194) $\downarrow$ & 24 & 119 (45) $\approx$ & \cellcolor{gray!25}31 \\
\hline

\multirow{1}{*}{\textbf{MW14}} & \multirow{1}{*}{3}  & 142 (28) & 31 & 179 (61) $\downarrow$ & 31 & 240 (95) $\downarrow$ & 31 & \cellcolor{gray!25}138 (50) $\approx$ & 31 & 184 (54) $\downarrow$ & 31 & 145 (53) $\approx$ & 31 \\
\hline

\multirow{1}{*}{\textbf{LIRCMOP1}} & \multirow{1}{*}{2}  & \cellcolor{gray!25}342 (90) & \cellcolor{gray!25}30 & 480 (39) $\downarrow$ & 10 & 478 (79) $\downarrow$ & 3 & 376 (99) $\approx$ & 23 & 477 (94) $\downarrow$ & 2 & 395 (100) $\downarrow$ & 25 \\
\hline

\multirow{1}{*}{\textbf{LIRCMOP2}} & \multirow{1}{*}{2}  & \cellcolor{gray!25}279 (60) & \cellcolor{gray!25}31 & 490 (26) $\downarrow$ & 4 & 461 (126) $\downarrow$ & 3 & 325 (94) $\approx$ & 27 & 467 (93) $\downarrow$ & 4 & 404 (93) $\downarrow$ & 24 \\
\hline

\multirow{1}{*}{\textbf{LIRCMOP3}} & \multirow{1}{*}{2}  & 309 (98) & 28 & 500 (2) $\downarrow$ & 1 & 483 (85) $\downarrow$ & 2 & \cellcolor{gray!25}265 (72) $\uparrow$ & \cellcolor{gray!25}31 & 499 (4) $\downarrow$ & 1 & 423 (96) $\downarrow$ & 18 \\
\hline

\multirow{1}{*}{\textbf{LIRCMOP4}} & \multirow{1}{*}{2}  & 298 (82) & \cellcolor{gray!25}31 & 500 (0) $\downarrow$ & 0 & 485 (84) $\downarrow$ & 1 & \cellcolor{gray!25}233 (63) $\uparrow$ & \cellcolor{gray!25}31 & 489 (61) $\downarrow$ & 1 & 415 (101) $\downarrow$ & 16 \\
\hline

\multirow{1}{*}{\textbf{LIRCMOP5-12}} & \multirow{1}{*}{2}  & 1 (0) & 31 & 1 (0) $\approx$ & 31 & 1 (0) $\approx$ & 31 & 1 (0) $\approx$ & 31 & 1 (0) $\approx$ & 31 & 1 (0) $\approx$ & 31 \\
\hline

\multirow{1}{*}{\textbf{LIRCMOP13-14}} & \multirow{1}{*}{3}  & 1 (0) & 31 & 1 (0) $\approx$ & 31 & 1 (0) $\approx$ & 31 & 1 (0) $\approx$ & 31 & 1 (0) $\approx$ & 31 & 1 (0) $\approx$ & 31 \\
\hline

\multirow{1}{*}{\textbf{DASCMOP1}} & \multirow{1}{*}{2}  & 7 (5) & 31 & 7 (5) $\approx$ & 31 & 6 (4) $\approx$ & 31 & 4 (5) $\approx$ & 31 & 5 (4) $\approx$ & 31 & 7 (5) $\approx$ & 31 \\
\hline

\multirow{1}{*}{\textbf{DASCMOP2}} & \multirow{1}{*}{2}  & 7 (5) & 31 & 7 (5) $\approx$ & 31 & 6 (4) $\approx$ & 31 & 7 (7) $\approx$ & 31 & 6 (5) $\approx$ & 31 & 7 (5) $\approx$ & 31 \\
\hline

\multirow{1}{*}{\textbf{DASCMOP3}} & \multirow{1}{*}{2}  & 7 (5) & 31 & 7 (5) $\approx$ & 31 & 6 (4) $\approx$ & 31 & 6 (4) $\approx$ & 31 & 6 (5) $\approx$ & 31 & 7 (5) $\approx$ & 31 \\
\hline

\multirow{1}{*}{\textbf{DASCMOP4}} & \multirow{1}{*}{2}  & \cellcolor{gray!25}162 (38) & \cellcolor{gray!25}31 & 220 (74) $\downarrow$ & \cellcolor{gray!25}31 & 499 (6) $\downarrow$ & 2 & 249 (128) $\downarrow$ & 25 & 500 (0) $\downarrow$ & 0 & 500 (0) $\downarrow$ & 0 \\
\hline

\multirow{1}{*}{\textbf{DASCMOP5}} & \multirow{1}{*}{2}  & \cellcolor{gray!25}172 (68) & 30 & 211 (66) $\downarrow$ & \cellcolor{gray!25}31 & 492 (29) $\downarrow$ & 5 & 272 (141) $\downarrow$ & 23 & 500 (0) $\downarrow$ & 0 & 500 (0) $\downarrow$ & 0 \\
\hline

\multirow{1}{*}{\textbf{DASCMOP6}} & \multirow{1}{*}{2}  & \cellcolor{gray!25}189 (104) & 29 & 198 (67) $\downarrow$ & \cellcolor{gray!25}31 & 497 (11) $\downarrow$ & 2 & 244 (131) $\downarrow$ & 25 & 500 (0) $\downarrow$ & 0 & 500 (0) $\downarrow$ & 0 \\
\hline

\multirow{1}{*}{\textbf{DASCMOP7}} & \multirow{1}{*}{3}  & \cellcolor{gray!25}157 (39) & \cellcolor{gray!25}31 & 254 (112) $\downarrow$ & 30 & 492 (27) $\downarrow$ & 4 & 197 (84) $\downarrow$ & 29 & 500 (0) $\downarrow$ & 0 & 500 (0) $\downarrow$ & 0 \\
\hline

\multirow{1}{*}{\textbf{DASCMOP8}} & \multirow{1}{*}{3}  & \cellcolor{gray!25}162 (71) & \cellcolor{gray!25}30 & 235 (105) $\downarrow$ & \cellcolor{gray!25}30 & 488 (49) $\downarrow$ & 4 & 218 (113) $\downarrow$ & 27 & 500 (0) $\downarrow$ & 0 & 500 (0) $\downarrow$ & 0 \\
\hline

\multirow{1}{*}{\textbf{DASCMOP9}} & \multirow{1}{*}{3}  & \cellcolor{gray!25}240 (66) & \cellcolor{gray!25}31 & 498 (9) $\downarrow$ & 1 & 500 (0) $\downarrow$ & 0 & 500 (0) $\downarrow$ & 0 & 500 (0) $\downarrow$ & 0 & 500 (0) $\downarrow$ & 0 \\
\hline

\multirow{1}{*}{\textbf{FFE ($\uparrow$/$\approx$/$\downarrow$)}} & & & & 1/19/19 & & 3/16/20 & & 5/22/15 & & 0/13/26 & & 0/15/22 & \\\hline

\end{tabular}
\end{center}
\end{table*}

\subsection{Performance comparison in terms of IGD, IGD\textsuperscript{+} and HV}

The IGD statistics based on 31 runs are presented in Table~\ref{tab:igd}, where PSCMOEA shows the best performance in 18~(out of 39) instances followed by KTS in 13 instances. However, KTS performs the best on LIRCMOP2 - LIRCMOP4 where the feasible region essentially is a very narrow strip. From our observation, both PSCMOEA and KTS could identify only a few feasible solutions~(maximum 2 or 3). The larger values in the metric is only due to one additional feasible solution. PSCMOEA performs the best for most of the feasibility-hard problems since it could identify the feasible solutions faster and more consistently, leaving remaining computational budget for obtaining high quality CPF approximation. The pairwise significance tests based on IGD~($\uparrow$/$\approx$/$\downarrow$ of the compared algorithms against PSCMOEA) are presented at the bottom of Table~\ref{tab:igd}. It can be seen that peer algorithms obtain significantly worse performance metrics than PSCMOEA. KTS performs best among them but shows worse performance than PSCMOEA in 18 instances, while being better in 7 instances including LIRCMOP2 - LIRCMOP4 as discussed above. For reference, the total $\uparrow$/$\approx$/$\downarrow$ statistics in terms of IGD\textsuperscript{+} and HV is also added at the bottom which also signifies the similar performance trend. More detailed tables of IGD\textsuperscript{+} and HV for MW, LIRCMOP and DASCMOP series are included in the SOM II, III and IV respectively. The plots of the obtained solutions by all compared algorithms on MW, LIRCMOP and DASCMOP series problems are also included as supplementary data for the interested readers in SOM Sections II, III and IV respectively, due to space constraints.

The performance profiles, based on mean IGD, are presented in Fig.~\ref{fig:perfprof} to summarize cumulative performance of the algorithms. Two profiles are drawn, corresponding to feasibility-hard and feasibility-easy problems(Fig.~\ref{fig:perfprof1}) and easy~(Fig.~\ref{fig:perfprof2}, respectively). For the former category, PSCMOEA clearly outperforms the other algorithm by obtaining the best results in almost 60\% of the total instances, and being competitive on the remaining instances. For the feasibility-easy problems, KTS shows marginally better performance than PSCMOEA, since it could use total evaluation budget to search for a good approximation of the CPF without wasting evaluations searching for UPF in such cases. PSCMOEA displays very similar profile to (and better than others), highlighting its ability to identify well distributed set of solutions on CPF.

\begin{table*}[!ht]\scriptsize
\centering
\caption{Mean IGD~(std.) across 31 runs obtained by PSCMOEA and peer-algorithms. $\uparrow$/$\approx$/$\downarrow$ symbols denote if the compared algorithm is significantly better/ equivalent/ worse than PSCMOEA. The gray shade highlights the best mean performance. The cumulative $\uparrow$/$\approx$/$\downarrow$ statistics of each algorithm against PSCMOEA in terms of IGD, IGD\textsuperscript{+} and HV are provided at the bottom.}
\label{tab:igd}
\tabcolsep 1mm
\renewcommand{\arraystretch}{1.0}
\begin{center}
\vspace{-3mm}
\begin{tabular}{|c|c|c|c|c|c|c|c|}
\hline
\multirow{2}{*}{\textbf{Prob.}} & \multirow{2}{*}{\textbf{M}} & \multicolumn{6}{c|}{\textbf{Algorithms}}\\\cline{3-8} 
 & & \multicolumn{1}{c|}{\textbf{PSCMOEA}} & \multicolumn{1}{c|}{\textbf{KMGSAEA}} & \multicolumn{1}{c|}{\textbf{ASA-MOEA/D}} & \multicolumn{1}{c|}{\textbf{KTS}} & \multicolumn{1}{c|}{\textbf{MultiObjectiveEGO}} & \multicolumn{1}{c|}{\textbf{SILE}}\\\hline

\multirow{1}{*}{\textbf{Test1}} & \multirow{1}{*}{2} & 2.08e-01 (1.10e-01)& 8.51e-01 (0.00e+00) $\downarrow$ & 2.08e-01 (3.53e-01) $\uparrow$ & 6.59e-01 (3.60e-01) $\downarrow$ & 5.96e-01 (2.15e-01) $\downarrow$ & \cellcolor{gray!25}1.68e-01 (3.40e-01) $\uparrow$ \\
\hline

\multirow{1}{*}{\textbf{Test2}} & \multirow{1}{*}{2} & \cellcolor{gray!25}1.21e-01 (1.93e-01) & 6.58e-01 (2.48e-01) $\downarrow$ & 1.59e-01 (2.35e-01) $\approx$ & 6.34e-01 (2.42e-01) $\downarrow$ & 5.88e-01 (2.24e-01) $\downarrow$ & 4.74e-01 (3.57e-01) $\downarrow$ \\
\hline

\multirow{1}{*}{\textbf{MW1}} & \multirow{1}{*}{2} & \cellcolor{gray!25}2.50e-01 (3.80e-01) & 6.80e-01 (2.89e-01) $\downarrow$ & 9.11e-01 (1.27e-01) $\downarrow$ & 8.32e-01 (2.98e-01) $\downarrow$ & 9.61e-01 (4.51e-16) $\downarrow$ & 9.09e-01 (1.37e-01) $\downarrow$ \\
\hline

\multirow{1}{*}{\textbf{MW2}} & \multirow{1}{*}{2} & \cellcolor{gray!25}1.58e-01 (1.30e-01) & 4.96e-01 (3.74e-01) $\downarrow$ & 5.45e-01 (2.29e-01) $\downarrow$ & 1.82e-01 (2.16e-01) $\approx$ & 4.20e-01 (1.74e-01) $\downarrow$ & 2.31e-01 (1.49e-01) $\downarrow$ \\
\hline

\multirow{1}{*}{\textbf{MW3}} & \multirow{1}{*}{2} & 2.05e-02 (7.83e-03)& 4.53e-02 (1.02e-02) $\downarrow$ & 9.64e-02 (3.41e-02) $\downarrow$ & \cellcolor{gray!25}1.74e-02 (2.50e-03) $\approx$ & 5.97e-01 (1.80e-02) $\downarrow$ & 8.42e-02 (8.39e-02) $\downarrow$ \\
\hline

\multirow{1}{*}{\textbf{MW4}} & \multirow{1}{*}{3} & \cellcolor{gray!25}4.99e-01 (3.98e-01) & 7.74e-01 (6.46e-01) $\approx$ & 1.46e+00 (1.57e-01) $\downarrow$ & 7.08e-01 (5.53e-01) $\downarrow$ & 1.52e+00 (6.77e-16) $\downarrow$ & 1.44e+00 (2.44e-01) $\downarrow$ \\
\hline

\multirow{1}{*}{\textbf{MW5}} & \multirow{1}{*}{2} & \cellcolor{gray!25}2.67e-01 (1.48e-01) & 4.50e-01 (2.51e-01) $\downarrow$ & 7.97e-01 (3.05e-01) $\downarrow$ & 6.82e-01 (3.30e-01) $\downarrow$ & 1.02e+00 (4.51e-16) $\downarrow$ & 9.60e-01 (1.28e-01) $\downarrow$ \\
\hline

\multirow{1}{*}{\textbf{MW6}} & \multirow{1}{*}{2} & \cellcolor{gray!25}5.44e-01 (2.17e-01) & 7.92e-01 (3.30e-01) $\downarrow$ & 1.01e+00 (3.58e-01) $\downarrow$ & 6.83e-01 (3.07e-01) $\downarrow$ & 8.60e-01 (3.03e-01) $\downarrow$ & 8.12e-01 (3.05e-01) $\downarrow$ \\
\hline

\multirow{1}{*}{\textbf{MW7}} & \multirow{1}{*}{2} & 2.87e-02 (1.15e-02)& 4.83e-02 (1.24e-02) $\downarrow$ & 3.97e-02 (1.31e-02) $\downarrow$ & \cellcolor{gray!25}2.47e-02 (5.88e-03) $\approx$ & 7.38e-01 (1.80e-02) $\downarrow$ & 5.23e-02 (7.04e-02) $\downarrow$ \\
\hline

\multirow{1}{*}{\textbf{MW8}} & \multirow{1}{*}{3} & \cellcolor{gray!25}3.01e-01 (4.06e-01) & 4.82e-01 (2.34e-01) $\downarrow$ & 1.62e+00 (2.74e-01) $\downarrow$ & 4.43e-01 (4.21e-01) $\downarrow$ & 1.49e+00 (6.10e-01) $\downarrow$ & 9.00e-01 (6.71e-01) $\downarrow$ \\
\hline

\multirow{1}{*}{\textbf{MW9}} & \multirow{1}{*}{2} & \cellcolor{gray!25}2.87e-01 (2.36e-01) & 6.30e-01 (3.32e-01) $\downarrow$ & 1.19e+00 (1.63e-01) $\downarrow$ & 6.19e-01 (4.34e-01) $\downarrow$ & 1.23e+00 (2.26e-16) $\downarrow$ & 1.23e+00 (2.26e-16) $\downarrow$ \\
\hline

\multirow{1}{*}{\textbf{MW10}} & \multirow{1}{*}{2} & \cellcolor{gray!25}6.46e-01 (2.11e-01) & 6.79e-01 (2.13e-01) $\approx$ & 8.07e-01 (1.13e-16) $\downarrow$ & 7.58e-01 (1.46e-01) $\downarrow$ & 8.07e-01 (1.13e-16) $\downarrow$ & 8.07e-01 (1.13e-16) $\downarrow$ \\
\hline

\multirow{1}{*}{\textbf{MW11}} & \multirow{1}{*}{2} & 2.86e-01 (2.02e-01)& 1.38e-01 (1.30e-01) $\uparrow$ & 1.74e-01 (1.38e-01) $\approx$ & \cellcolor{gray!25}1.11e-01 (1.52e-01) $\uparrow$ & 7.88e-01 (5.56e-02) $\downarrow$ & 6.07e-01 (2.14e-01) $\downarrow$ \\
\hline

\multirow{1}{*}{\textbf{MW12}} & \multirow{1}{*}{2} & \cellcolor{gray!25}1.80e-01 (1.99e-01) & 4.60e-01 (4.00e-01) $\downarrow$ & 1.21e+00 (4.38e-01) $\downarrow$ & 4.65e-01 (4.10e-01) $\downarrow$ & 1.49e+00 (2.26e-16) $\downarrow$ & 1.06e+00 (3.66e-01) $\downarrow$ \\
\hline

\multirow{1}{*}{\textbf{MW13}} & \multirow{1}{*}{2} & 4.88e-01 (2.75e-01)& \cellcolor{gray!25}4.85e-01 (5.21e-01) $\approx$ & 2.39e+00 (1.15e+00) $\downarrow$ & 7.78e-01 (7.34e-01) $\downarrow$ & 3.58e+00 (1.33e+00) $\downarrow$ & 1.28e+00 (1.23e+00) $\downarrow$ \\
\hline

\multirow{1}{*}{\textbf{MW14}} & \multirow{1}{*}{3} & \cellcolor{gray!25}7.74e-02 (2.62e-02) & 5.63e-01 (1.81e-01) $\downarrow$ & 7.34e-01 (2.19e-01) $\downarrow$ & 1.91e-01 (1.47e-01) $\downarrow$ & 8.95e-01 (1.93e-01) $\downarrow$ & 7.48e-01 (1.72e-01) $\downarrow$ \\
\hline

\multirow{1}{*}{\textbf{LIRCMOP1}} & \multirow{1}{*}{2} & 5.41e-01 (1.59e-01)& 8.55e-01 (1.50e-01) $\downarrow$ & 8.86e-01 (1.55e-01) $\downarrow$ & 4.69e-01 (3.25e-01) $\approx$ & 9.01e-01 (1.35e-01) $\downarrow$ & \cellcolor{gray!25}4.40e-01 (2.77e-01) $\uparrow$ \\
\hline

\multirow{1}{*}{\textbf{LIRCMOP2}} & \multirow{1}{*}{2} & 5.96e-01 (1.64e-01)& 9.10e-01 (7.19e-02) $\downarrow$ & 8.75e-01 (1.83e-01) $\downarrow$ & \cellcolor{gray!25}3.00e-01 (2.79e-01) $\uparrow$ & 8.59e-01 (1.97e-01) $\downarrow$ & 4.30e-01 (2.82e-01) $\uparrow$ \\
\hline

\multirow{1}{*}{\textbf{LIRCMOP3}} & \multirow{1}{*}{2} & 5.52e-01 (1.75e-01)& 9.15e-01 (1.94e-02) $\downarrow$ & 8.92e-01 (1.07e-01) $\downarrow$ & \cellcolor{gray!25}3.12e-01 (1.52e-01) $\uparrow$ & 9.03e-01 (9.08e-02) $\downarrow$ & 6.07e-01 (2.73e-01) $\approx$ \\
\hline

\multirow{1}{*}{\textbf{LIRCMOP4}} & \multirow{1}{*}{2} & 5.98e-01 (1.61e-01)& 9.09e-01 (3.39e-16) $\downarrow$ & 8.91e-01 (9.89e-02) $\downarrow$ & \cellcolor{gray!25}2.70e-01 (1.28e-01) $\uparrow$ & 8.91e-01 (9.94e-02) $\downarrow$ & 6.11e-01 (2.95e-01) $\approx$ \\
\hline

\multirow{1}{*}{\textbf{LIRCMOP5}} & \multirow{1}{*}{2} & 1.21e-01 (2.23e-01)& 3.05e-02 (6.83e-03) $\uparrow$ & 1.02e+00 (4.65e-01) $\downarrow$ & \cellcolor{gray!25}2.77e-02 (9.07e-03) $\uparrow$ & 2.86e+00 (4.09e-01) $\downarrow$ & 6.94e-01 (2.21e-01) $\downarrow$ \\
\hline

\multirow{1}{*}{\textbf{LIRCMOP6}} & \multirow{1}{*}{2} & 3.94e-02 (3.93e-02)& \cellcolor{gray!25}3.47e-02 (4.57e-03) $\approx$ & 9.89e-01 (5.04e-01) $\downarrow$ & 4.40e-02 (1.93e-02) $\downarrow$ & 2.88e+00 (5.51e-01) $\downarrow$ & 4.23e-01 (8.17e-02) $\downarrow$ \\
\hline

\multirow{1}{*}{\textbf{LIRCMOP7}} & \multirow{1}{*}{2} & 2.26e-01 (1.09e-01)& 2.47e-01 (8.30e-02) $\approx$ & 4.17e-01 (3.81e-01) $\downarrow$ & \cellcolor{gray!25}2.07e-01 (4.56e-02) $\approx$ & 1.68e+00 (4.68e-01) $\downarrow$ & 3.12e-01 (1.65e-01) $\downarrow$ \\
\hline

\multirow{1}{*}{\textbf{LIRCMOP8}} & \multirow{1}{*}{2} & \cellcolor{gray!25}1.86e-01 (7.25e-02) & 2.41e-01 (1.79e-01) $\approx$ & 4.17e-01 (3.81e-01) $\downarrow$ & 1.89e-01 (5.56e-02) $\approx$ & 1.82e+00 (4.43e-01) $\downarrow$ & 4.37e-01 (1.87e-01) $\downarrow$ \\
\hline

\multirow{1}{*}{\textbf{LIRCMOP9}} & \multirow{1}{*}{2} & \cellcolor{gray!25}1.36e-01 (5.49e-02) & 2.98e-01 (6.36e-02) $\downarrow$ & 3.40e-01 (1.01e-01) $\downarrow$ & 1.74e-01 (6.09e-02) $\downarrow$ & 1.07e+00 (2.24e-01) $\downarrow$ & 3.63e-01 (5.14e-02) $\downarrow$ \\
\hline

\multirow{1}{*}{\textbf{LIRCMOP10}} & \multirow{1}{*}{2} & 1.07e-01 (7.34e-02)& 3.17e-01 (8.69e-02) $\downarrow$ & 3.46e-01 (1.04e-01) $\downarrow$ & \cellcolor{gray!25}6.78e-02 (2.59e-02) $\uparrow$ & 8.72e-01 (1.43e-01) $\downarrow$ & 3.68e-01 (4.54e-02) $\downarrow$ \\
\hline

\multirow{1}{*}{\textbf{LIRCMOP11}} & \multirow{1}{*}{2} & \cellcolor{gray!25}1.09e-01 (5.63e-02) & 2.19e-01 (6.39e-02) $\downarrow$ & 3.20e-01 (9.23e-02) $\downarrow$ & 1.10e-01 (6.02e-02) $\approx$ & 8.14e-01 (1.05e-01) $\downarrow$ & 3.13e-01 (4.81e-02) $\downarrow$ \\
\hline

\multirow{1}{*}{\textbf{LIRCMOP12}} & \multirow{1}{*}{2} & \cellcolor{gray!25}1.12e-01 (3.02e-02) & 2.25e-01 (7.52e-02) $\downarrow$ & 2.11e-01 (7.62e-02) $\downarrow$ & 1.52e-01 (4.18e-02) $\downarrow$ & 8.90e-01 (1.64e-01) $\downarrow$ & 2.93e-01 (4.94e-02) $\downarrow$ \\
\hline

\multirow{1}{*}{\textbf{LIRCMOP13}} & \multirow{1}{*}{3} & 1.41e-01 (1.96e-01)& 6.05e-01 (1.93e-01) $\downarrow$ & 1.09e+00 (2.02e-01) $\downarrow$ & \cellcolor{gray!25}6.79e-02 (1.16e-02) $\approx$ & 1.22e+00 (1.15e-01) $\downarrow$ & 1.12e+00 (1.17e-01) $\downarrow$ \\
\hline

\multirow{1}{*}{\textbf{LIRCMOP14}} & \multirow{1}{*}{3} & 2.05e-01 (2.43e-01)& 5.98e-01 (1.50e-01) $\downarrow$ & 1.07e+00 (1.50e-01) $\downarrow$ & \cellcolor{gray!25}1.75e-01 (3.75e-02) $\downarrow$ & 1.15e+00 (1.18e-01) $\downarrow$ & 1.08e+00 (1.33e-01) $\downarrow$ \\
\hline

\multirow{1}{*}{\textbf{DASCMOP1}} & \multirow{1}{*}{2} & \cellcolor{gray!25}2.06e-02 (1.07e-02) & 6.28e-01 (1.40e-01) $\downarrow$ & 6.36e-02 (2.14e-02) $\downarrow$ & 6.14e-02 (7.43e-02) $\approx$ & 4.77e-01 (1.18e-01) $\downarrow$ & 4.07e-01 (1.59e-01) $\downarrow$ \\
\hline

\multirow{1}{*}{\textbf{DASCMOP2}} & \multirow{1}{*}{2} & 1.06e-01 (2.09e-02)& 4.96e-01 (6.06e-02) $\downarrow$ & 1.34e-01 (2.12e-02) $\downarrow$ & \cellcolor{gray!25}9.30e-02 (1.03e-02) $\uparrow$ & 4.24e-01 (5.79e-02) $\downarrow$ & 3.51e-01 (6.48e-02) $\downarrow$ \\
\hline

\multirow{1}{*}{\textbf{DASCMOP3}} & \multirow{1}{*}{2} & 2.77e-01 (6.89e-02)& 5.95e-01 (8.33e-02) $\downarrow$ & \cellcolor{gray!25}2.61e-01 (5.83e-02) $\approx$ & 4.44e-01 (7.89e-02) $\downarrow$ & 5.12e-01 (5.91e-02) $\downarrow$ & 4.18e-01 (8.24e-02) $\downarrow$ \\
\hline

\multirow{1}{*}{\textbf{DASCMOP4}} & \multirow{1}{*}{2} & 4.43e-01 (3.39e-01)& \cellcolor{gray!25}4.05e-01 (2.60e-01) $\approx$ & 1.20e+00 (8.16e-02) $\downarrow$ & 4.79e-01 (3.97e-01) $\approx$ & 1.22e+00 (0.00e+00) $\downarrow$ & 1.22e+00 (0.00e+00) $\downarrow$ \\
\hline

\multirow{1}{*}{\textbf{DASCMOP5}} & \multirow{1}{*}{2} & 4.78e-01 (2.93e-01)& \cellcolor{gray!25}4.49e-01 (1.81e-01) $\approx$ & 1.26e+00 (1.31e-01) $\downarrow$ & 6.63e-01 (4.23e-01) $\approx$ & 1.30e+00 (6.77e-16) $\downarrow$ & 1.30e+00 (6.77e-16) $\downarrow$ \\
\hline

\multirow{1}{*}{\textbf{DASCMOP6}} & \multirow{1}{*}{2} & 5.88e-01 (2.81e-01)& \cellcolor{gray!25}5.84e-01 (1.76e-01) $\approx$ & 1.32e+00 (6.69e-02) $\downarrow$ & 7.01e-01 (3.52e-01) $\approx$ & 1.34e+00 (9.03e-16) $\downarrow$ & 1.34e+00 (9.03e-16) $\downarrow$ \\
\hline

\multirow{1}{*}{\textbf{DASCMOP7}} & \multirow{1}{*}{3} & 3.85e-01 (3.58e-01)& 3.29e-01 (2.44e-01) $\approx$ & 1.21e+00 (1.38e-01) $\downarrow$ & \cellcolor{gray!25}2.58e-01 (3.42e-01) $\approx$ & 1.26e+00 (4.51e-16) $\downarrow$ & 1.26e+00 (4.51e-16) $\downarrow$ \\
\hline

\multirow{1}{*}{\textbf{DASCMOP8}} & \multirow{1}{*}{3} & \cellcolor{gray!25}3.67e-01 (3.79e-01) & 4.10e-01 (2.76e-01) $\approx$ & 1.44e+00 (1.78e-01) $\downarrow$ & 3.96e-01 (4.77e-01) $\approx$ & 1.50e+00 (6.77e-16) $\downarrow$ & 1.50e+00 (6.77e-16) $\downarrow$ \\
\hline

\multirow{1}{*}{\textbf{DASCMOP9}} & \multirow{1}{*}{3} & \cellcolor{gray!25}4.60e-01 (1.61e-01) & 1.26e+00 (1.80e-02) $\downarrow$ & 1.26e+00 (4.51e-16) $\downarrow$ & 1.26e+00 (4.51e-16) $\downarrow$ & 1.26e+00 (4.51e-16) $\downarrow$ & 1.26e+00 (4.51e-16) $\downarrow$ \\
\hline

\multirow{1}{*}{\textbf{IGD ($\uparrow$/$\approx$/$\downarrow$)}} & & & 2/11/26 & 1/3/35 & 7/14/18 & 0/0/39 & 3/2/34 \\\hline

\multirow{1}{*}{\textbf{IGD\textsuperscript{+} ($\uparrow$/$\approx$/$\downarrow$)}} & & & 2/11/26 & 1/2/36 & 7/13/19 & 0/0/39 & 3/2/34 \\\hline

\multirow{1}{*}{\textbf{HV ($\uparrow$/$\approx$/$\downarrow$)}} & & & 1/9/29 & 1/4/34 & 5/17/17 & 0/0/39 & 4/3/32 \\\hline

\end{tabular}
\end{center}
\end{table*}

\subsection{Component-wise performance assessment of PSCMOEA}
To systematically observe the benefits offered by the key internal components of PSCMOEA, three variants of PSCMOEA are constructed. Of these, the first two are constructed by modifying the probabilistic selection ranking strategy. The third variant is constructed to show the effectiveness of the search switching mechanism. The brief outline of the variants is presented below.

\begin{itemize}
    \item $V1$: In the first variant, instead of ranking the solutions along the RVs for `Constrained' search according to Eq.~\ref{eqn:P_AB}, a lexicographic ranking strategy is undertaken which essentially follows a feasibility-first~(FF) principle. According to it, the solution cluster along a RV is divided into feasible and infeasible set based on their corresponding PoF score as represented in Eq.~\ref{eqn:pof}. The threshold value of PoF for each individual constraint distribution is set as $\ge 0.99$ to treat a solution as feasible. Note that each constraint has to satisfy the threshold PoF value. Among the feasible block, the solutions are sorted based on their $\text{PD}_\text{F}$ score, as represented in Eq.~\ref{eqn:PD}. For the infeasible block, the solutions are first sorted based on their PoF product score~(higher the better). If there are solutions with the same PoF product score, then they are further sorted based on their number of satisfied~(NS) constraints~(higher, the better). If there remain solutions with the same NS counts, they are sorted based on their mean CV predictions~(smaller, the better). Finally the sorted feasible block of solutions are placed above the sorted infeasible ones. Note that the same ranking strategy is followed during the infill identification stage as well when the archive is fully infeasible. All other components are kept as same as PSCMOEA.     

    \item $V2$: The second variant contains a simpler ranking strategy. For a block of candidate solutions along a RV, numerical score of PoF and $\text{PD}_\text{F}$ is calculated for each individual candidate. For each solution, the product of PoF and $\text{PD}_\text{F}$ scores~($\text{PoF}\times\text{PD}_\text{F}$) is considered as the overall score. The scores for all solutions along the RV is calculated, and they are sorted in descending order. Same principle is undertaken during the infill identification stage by keeping all the remaining components as same as PSCMOEA.

    \item $V3$: This variant is constructed simply by taking off the search switching mechanism~(Sec.~\ref{subsec:switch}) from PSCMOEA while keeping all other components the same.
\end{itemize}

All variants~(including PSCMOEA), are run on each test problem 31 times. The performance is compared pairwise with each other variant in terms of IGD, IGD\textsuperscript{+}, HV and FFE, and the results are summarized in Table~\ref{tab:perfsig}.  For example, MW contains a total of 14 problems. This means total 42~(= $14\times3$) comparisons are performed for PSCMOEA against the three variants. More detailed statistics including mean IGD, IGD\textsuperscript{+}, HV, FFE, ST are included in the SOM Section V.

As mentioned above, the first two variants~($V1$ and $V2$) are constructed to observe the effectiveness of the proposed integrated ranking method using PoF and PD metrics following Eq.~\ref{eqn:P_AB}. From the given statistics in Table~\ref{tab:perfsig}, one can clearly observe that both $V1$ and $V2$ perform worse than PSCMEOA in terms of all metrics. This is because the ranking approach used in $V1$ is biased more on PoF metric than $\text{PD}_\text{F}$, where $\text{PD}_\text{F}$ is only applied for ranking the solutions with high likelihood of feasibility~(by considering $\text{PoF}\ge 0.99$ as feasible). For the feasibility-hard problems, such an approach drives the search very slowly in the original objective space as it ignores the dominance relationship among the candidates in the predicted fitness landscape. On the other hand, the second variant $V2$ equally weighs them by simply multiplying PoF and $\text{PD}_\text{F}$ to order the candidate solutions. For feasibility-hard problems, such approach can expedite the convergence rate for the instances where a good positive correlation exists between objective optimization and CV minimization direction. This is because $\text{PD}_\text{F}$ can drive the search towards optimal objective location when PoF is very low. For this case, the win statistics improves for MW test suites compared to $V1$ as indicated in Table~\ref{tab:perfsig}. However, the weakness of such approach can be observed on the problems where the correlation is negative between objective optimization and CV minimization direction, for example on Test1 where it could not identify any feasible solution across 31 runs. Moreover, there is no means to fully rank the candidates with 0 PoF. In addition, the information regarding the CV distribution with low PoF is ignored in the approach, hence resulting in performance degradation for such scenarios. From these results, it could be inferred that PSCMOEA ranking strategy can more efficiently deal with a range of scenarios.

\begin{table}[!ht]\scriptsize
\centering
\caption{The cumulative results of pairwise significance tests~($\uparrow$/$\approx$/$\downarrow$) between all pairs of variants, including PSCMOEA.}
\label{tab:perfsig}
\tabcolsep 1.3mm
\renewcommand{\arraystretch}{1.1}
\begin{center}
\vspace{-3mm}
\begin{tabular}{ccccccc}
\hline
\multirow{2}{*}{\textbf{Metric}} & \multirow{2}{*}{\textbf{Problems}} & \multirow{2}{*}{$\mathbf{M}$} & \multicolumn{4}{c}{\textbf{Algorithms}}\\\cline{4-7} & & 
 & \multicolumn{1}{c}{\textbf{PSCMOEA}} & \multicolumn{1}{c}{\textbf{$V1$}} & \multicolumn{1}{c}{\textbf{$V2$}} & \multicolumn{1}{c}{\textbf{$V3$}}\\\hline

\multirow{3}{*}{\textbf{IGD}} & \multirow{1}{*}{\textbf{Test}} & 2 &  \cellcolor{gray!25}3/3/0 & 1/3/2 & 0/1/5 &  \cellcolor{gray!25}3/3/0 \\
& \multirow{1}{*}{\textbf{MW}} & (2 \& 3) &  \cellcolor{gray!25}23/16/3 & 7/11/24 & 13/13/16 & 11/20/11 \\
& \multirow{1}{*}{\textbf{DASCMOP}} & (2 \& 3) & 16/9/2 & 2/7/18 & 2/7/18 &  \cellcolor{gray!25}18/9/0 \\
\hline

\multirow{3}{*}{\textbf{IGD\textsuperscript{+}}} & \multirow{1}{*}{\textbf{Test}} & 2 &  \cellcolor{gray!25}3/3/0 & 1/3/2 & 0/1/5 &  \cellcolor{gray!25}3/3/0 \\
& \multirow{1}{*}{\textbf{MW}} & (2 \& 3) &  \cellcolor{gray!25}24/16/2 & 6/10/26 & 13/13/16 & 13/17/12 \\
& \multirow{1}{*}{\textbf{DASCMOP}} & (2 \& 3) & 16/8/3 & 4/5/18 & 2/7/18 &  \cellcolor{gray!25}18/8/1 \\
\hline

\multirow{3}{*}{\textbf{HV}} & \multirow{1}{*}{\textbf{Test}} & 2 &  \cellcolor{gray!25}3/3/0 & 1/3/2 & 0/1/5 &  \cellcolor{gray!25}3/3/0 \\
& \multirow{1}{*}{\textbf{MW}} & (2 \& 3) &  \cellcolor{gray!25}23/17/2 & 6/11/25 & 13/13/16 & 11/21/10 \\
& \multirow{1}{*}{\textbf{DASCMOP}} & (2 \& 3) & 16/9/2 & 4/5/18 & 2/8/17 &  \cellcolor{gray!25}16/10/1 \\
\hline

\multirow{3}{*}{\textbf{FFE}} & \multirow{1}{*}{\textbf{Test}} & 2 &  \cellcolor{gray!25}1/4/1 &  \cellcolor{gray!25}1/4/1 & 3/0/3 &  \cellcolor{gray!25}1/4/1 \\
& \multirow{1}{*}{\textbf{MW}} & (2 \& 3) &  \cellcolor{gray!25}16/22/4 & 7/12/23 & 14/18/10 & 10/22/10 \\
& \multirow{1}{*}{\textbf{DASCMOP}} & (2 \& 3) &  \cellcolor{gray!25}14/13/0 & 0/14/13 & 1/14/12 & 12/13/2 \\
\hline

\end{tabular}
\end{center}
\end{table}

\begin{figure}[!ht]
\centering    
\includegraphics[width=0.40\textwidth]{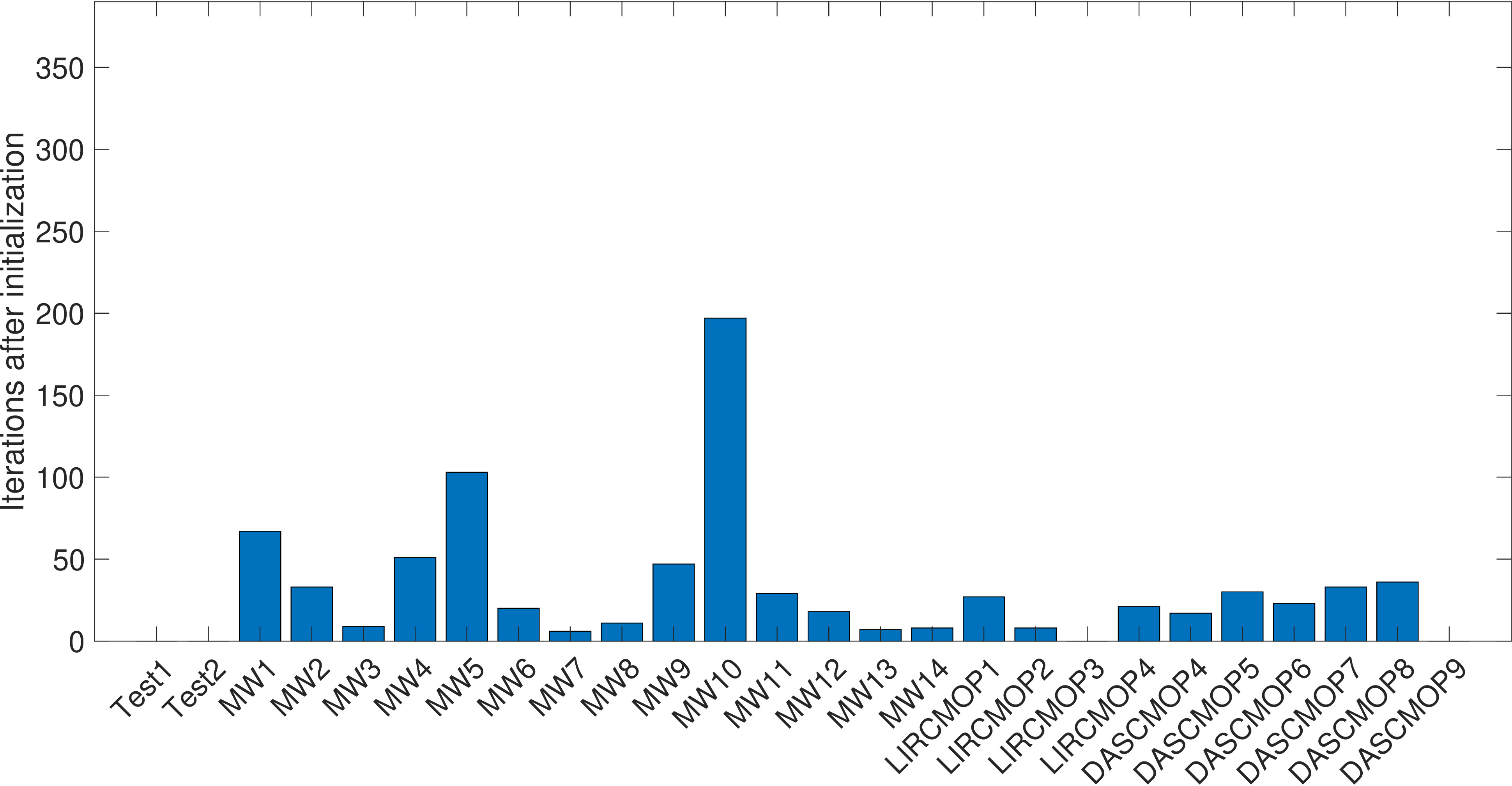}
\caption{Number of times PSCMOEA switches to unconstrained search in the median-run (out of 31 trials) in terms of IGD metric on (a) MW series, Test1 - Test2, LIRCMOP1 - LIRCMOP4 and DASCMOP4 - DASCMOP9.}
\label{fig:switch}
\end{figure}

As for $V3$, it is constructed to observe the influence of the search switching mechanism on the overall performance of PSCMOEA. From Table~\ref{tab:perfsig}, some interesting performance trends can be observed. The switching mechanism clearly improves the performance significantly in terms of all metrics on the MW test suite while a slight performance degradation can be noticed on DASCMOP series. This is because most of the problems in MW test suite have a positive correlation between UPF and CPF search direction, therefore switching to unconstrained search occasionally expedites the search. The opposite situation happens in DASCMOP series for not having good correlation. The switching behavior can be observed from the bar chart in Fig.~\ref{fig:switch} which illustrates the number of times PSCMOEA switched to unconstrained search during the median-run~(out of 31 runs in terms of IGD metric) for the feasibility-hard problems. From the figure, it is clear that there are higher number of iterations conducted in unconstrained mode for MW series than those from other test suites. In particular for MW10, the number is highest because the feasibility rate is very low and the relationship between CPF and UPF remains positive during the search. On the other hand, in Test1 and Test2, no unconstrained search was initiated due to the consistent negative correlation. LIRCMOP and DASCMOP encounter some interactions in unconstrained mode, however, it did not improve the solutions, PSCMOEA did not waste further resources on unconstrained search. Overall the number of iterations that switch to unconstrained search are typically very low, indicating that PSCMOEA is able to swiftly identify if the unconstrained search is of any benefit, and switch back to the default constrained search if not. 

\section{Concluding remarks}
\label{sec:conclusion}

In this study, some of the existing works have been reviewed that tackle ECMOPs, and highlighted how the relative position of UPF and CPF and model uncertainties can impact their performance. These observations along with the lack of steady-state methods to solve ECMOPs, formed the key motivations of the study. To address the identified gaps, the paper develops an efficient approach based on probabilistic selection, referred to as PSCMOEA, for solving ECMOPs on a low evaluation budget. The key differentiating feature of PSCMEOA is its careful consideration of model uncertainties and the status of evaluated solutions in terms of feasibility, convergence, and diversity in its components. These considerations feed into multiple components, such as environmental selection using probabilistic constrained dominance~($PCD$), adaptive setting of normalization bounds to effectively explore the predicted space and an adaptive search switching mechanism to expedite the convergence when the UPF and CPF search directions are similar. Numerical experiments and benchmarking with five state-of-the-art algorithms  were conducted on an extensive set of problems to demonstrate the efficacy of PSCMOEA. Moreover, ablation studies were conducted to highlight and verify the effectiveness of some of the key components.  

Some of the potential directions for future work include extension to many-objective problems~(i.e., those with more than 3 objectives) and development of a generational version of the proposed approach. 

\ifCLASSOPTIONcaptionsoff
  \newpage
\fi
\balance
\bibliographystyle{IEEEtran}
\bibliography{arxiv2024pscmoea.bib}

\end{document}